%% file: arxiv.tex
\pdfoutput=1
\documentclass{article}

\usepackage[margin=1in]{geometry}  
\usepackage{mathpazo}
\usepackage[T1]{fontenc}    
\usepackage[backend=bibtex,style=alphabetic,natbib=true]{biblatex}
\usepackage{hyperref}

\newif\ificml
\icmlfalse

\input{header}
\title{Leveraging Sparse Linear Layers for \\ Debuggable Deep 
	Networks}

\author{
	Eric Wong\footnotemark[1] \\
	MIT \\
	 \texttt{wongeric@mit.edu} \\
	\and
	Shibani Santurkar\thanks{Equal contribution.} \\
	MIT \\
	 \texttt{shibani@mit.edu} 
	\and 
	Aleksander M\k{a}dry \\
	MIT \\
	\texttt{madry@mit.edu}  
}
\date{}

\input{editor}

\begin{document}
	\maketitle
\input{abstract}
\input{intro}
\input{glm}

\input{verification}

\input{diagnosis}

\input{related}

\input{conclusion}

\input{acknowledgement}

\small 
\printbibliography

\clearpage
\appendix
\normalsize
\input{appendix_glm}
\input{appendix_lime}
\input{appendix_datasets}
\input{appendix_verification}

\input{appendix_bias}

\input{appendix_counterfactuals}

\input{appendix_errors}
	
\end{document}

%% file: header.tex
\usepackage{amsmath}
\usepackage{comment}
\usepackage{array}
\usepackage{enumerate}
\usepackage{amsthm}

\usepackage[utf8]{inputenc} 
\usepackage[T1]{fontenc}    
\usepackage{hyperref}       
\usepackage{url}            
\usepackage{booktabs}       
\usepackage{amsfonts}       
\usepackage{nicefrac}       
\usepackage{microtype}      
\usepackage{subcaption}
\usepackage{graphicx}
\usepackage{multirow}
\usepackage{dsfont}
\usepackage{algorithm}
\usepackage[algo2e]{algorithm2e} 
\usepackage{diagbox}
\usepackage{makecell}
\usepackage{longtable}
\usepackage{tabularx}
\usepackage{censor}
\usepackage{lipsum}
\usepackage{capt-of}

\usepackage{algorithm}
\usepackage{algorithmic}
\usepackage[algo2e]{algorithm2e}

\usepackage{graphbox}

\usepackage{graphbox}

\newcommand{\eps}{\varepsilon}
\renewcommand{\epsilon}{\varepsilon}

\DeclareMathOperator*{\argmax}{arg\,max}
\DeclareMathOperator*{\argmin}{arg\,min}

\DeclareMathOperator{\wordcloud}{WordCloud}

\usepackage{xcolor}
\usepackage{soul}

\newcommand{\sparsemod}{sparse decision layer}

\makeatletter
\let\c@table\c@figure 
\let\ftype@table\ftype@figure 
\makeatother

%% file: editor.tex
\usepackage[normalem]{ulem}
\usepackage{xcolor}

\newcommand{\scrcmd}{\textsuperscript}
\newcommand{\scr}[1]{\scrcmd{#1}}

%% file: abstract.tex
\begin{abstract}
We show how fitting sparse linear models over learned deep feature 
representations can lead to more debuggable neural networks. These 
networks remain highly accurate while also being more amenable to human 
interpretation, as we demonstrate quantiatively via numerical and 
human 
experiments. 
We further illustrate how the resulting sparse explanations can help to 
identify spurious correlations, explain misclassifications, and diagnose 
 model biases in vision and language tasks.\footnote{The code for our toolkit 
 can be found at 
 	\url{https://github.com/madrylab/debuggabledeepnetworks}.}

\end{abstract}

%% file: intro.tex
\section{Introduction}
As machine learning (ML) models find wide-spread application, there is a 
growing demand for interpretability: access to tools that help people 
see  \emph{why} the model made its
decision.
There are still many obstacles towards achieving this goal though, 
particularly in the context of
deep learning.
These obstacles stem from the scale of modern deep networks, 
as well as the complexity of even defining and assessing 
the (often context-dependent) desiderata of interpretability.

Existing work on deep network interpretability has largely approached 
this problem from two perspectives.
The first one seeks to uncover the 
concepts associated with specific neurons in the
network, for example through 
visualization~\citep{yosinski2015understanding} or semantic 
labeling~\cite{bau2017network}.
The second aims to explain 
model decisions on a per-example basis, using techniques such as 
local surrogates~\citep{ribeiro2016should} and 
saliency maps~\citep{simonyan2013deep}. 
While both families of approaches can improve model 
understanding at a local level---i.e., for a given example or neuron---recent 
work has argued that such localized explanations 
can lead to misleading conclusions about the model's overall decision 
process~\cite{adebayo2018sanity,adebayo2020debugging,leavitt2020towards}.
As a result, it is often challenging to flag a model's failure modes 
or evaluate corrective interventions without  in-depth 
problem-specific studies. 

To make progress on this front, we focus on a more 
actionable 
intermediate goal of interpretability: \emph{model 
debugging}. 
Specifically, instead of directly aiming for a complete characterization 
of the model's decision process, our objective is to develop tools that help 
model designers uncover unexpected model behaviors 
(semi-)automatically. 

\paragraph{Our contributions.}
Our approach to model debugging is based on a natural view of a deep 
network as the composition of a
``deep feature extractor'' and a linear ``decision layer''. 
Embracing this perspective allows us to focus our attention on 
probing how deep features are (linearly) combined by the decision layer to 
make predictions.
Even with this simplification, probing current deep networks can be 
intractable given 
the large number of parameters in their decision layers.
To overcome this challenge, we replace the standard (typically  
dense) decision layer of a deep network with a sparse but comparably 
accurate counterpart.
We find that this simple approach ends up being 
surprisingly effective for building deep networks that are intrinsically more 
debuggable. Specifically, for a variety of
modern ML settings:

\begin{itemize}

   \item We demonstrate that it is possible to construct deep networks that 
   have sparse decision layers (e.g., 
   with only 20-30 deep features per class for ImageNet) without sacrificing 
   much model performance. 
   This involves developing a custom solver for fitting elastic net regularized linear models in order to perform effective sparsification at deep-learning scales.\footnote{A standalone package of 
    our solver is available at  
   \url{https://github.com/madrylab/glm_saga}}
   
   \item We show that sparsifying a network's decision layer 
   can indeed help humans understand the resulting models 
   better. For example, 
   untrained 
   annotators 
   can intuit (simulate) the predictions of a model with a \sparsemod{} with 
   high ($\sim$63\%)
   accuracy.
   This is in contrast to their near chance performance ($\sim$33\%) for  
    models with standard (dense) decision layers.

   \item We explore the use of \sparsemod s in three debugging tasks: 
   diagnosing 
   biases and spurious correlations
   (cf. Section~\ref{sec:biases}), counterfactual generation 
   (cf. Section~\ref{sec:counterfactuals}) and identifying data patterns that 
   cause 
    misclassifications
   (cf. Section~\ref{sec:errors}). To enable this analysis, we 
   design a suite of human-in-the-loop experiments.
\end{itemize}

\ificml
We plan to release the code for our toolkit with the paper.
\fi

%% file: glm.tex
\section{Debuggability via Sparse Linearity}
\label{sec:methodology}
Recent studies have raised concerns about how deep networks 
make 
decisions~\citep{beery2018recognition,xiao2020noise,tsipras2020from,bissoto2020debiasing}.
For instance, it was noted that skin-lesion detectors rely on 
spurious visual artifacts~\citep{bissoto2020debiasing} and comment flagging 
systems use identity 
group information to detect 
toxicity~\citep{borkan2019nuanced}. 
So far, most of these discoveries were made via in-depth
studies by experts.
However, as deep learning makes inroads into new fields, there is a strong 
case to be made for 
general-purpose model debugging tools.

While simple models (e.g., small decision trees or linear classifiers) can  
be 
directly examined, a similar analysis for typical deep networks is infeasible. 
To tackle this problem, we choose to 
decompose a deep network into: (1) a deep feature representation and (2) 
a 
linear decision layer. 
Then, we can attempt to gain insight into the model's reasoning 
process by directly examining the deep features, and the linear coefficients 
used to aggregate them. 
At a high level, our hope is that this decomposition will allow us to get the 
best of both worlds: the predictive power of learned deep features, and 
the ease of understanding linear models.


That being said, this simplified problem is still intractable for current deep 
networks, since their decision layers can easily have millions of 
parameters operating on thousands of deep features.
To mitigate this issue, we instead combine the  feature representation of a 
pre-trained network with a 
\emph{sparse} linear decision layer (cf. Figure~\ref{fig:decomposition}).
Debugging the resulting sparse decision layer then entails  inspecting only 
the few linear coefficients and deep features 
that dictate its predictions. 



\subsection{Constructing sparse decision layers }
\label{sec:glm_explain}
One possible approach for constructing sparse decision layers is to apply 
pruning methods from deep learning 
\citep{lecunoptimal1990,han2015learning,hassibisecond1993,li2016pruning,han2016deep,blalock2020state}---commonly-used
 to 
compress 
deep networks and speed up inference---solely to the dense decision layer. 
It turns out however that for linear classifiers we can actually do better. 
In particular, the problem of fitting sparse linear models has been extensively 
studied in 
statistics, leading to a suite of methods with
theoretical optimality guarantees. 
These include \textsc{LASSO} regression 
\citep{tibshirani1994regression}, 
least angle regression \cite{efron2004least}, and forward stagewise 
regression \citep{hastie2007forward}.
In this paper, we leverage the classic elastic net 
formulation~\cite{zou2005regularization}---a generalization of 
\textsc{LASSO}  and ridge regression 
that addresses their corresponding drawbacks (further discussed in 
Appendix~\ref{app:solver}). 

\ificml
\begin{figure}[!t]
	\centering
	\includegraphics[width=0.9\columnwidth]{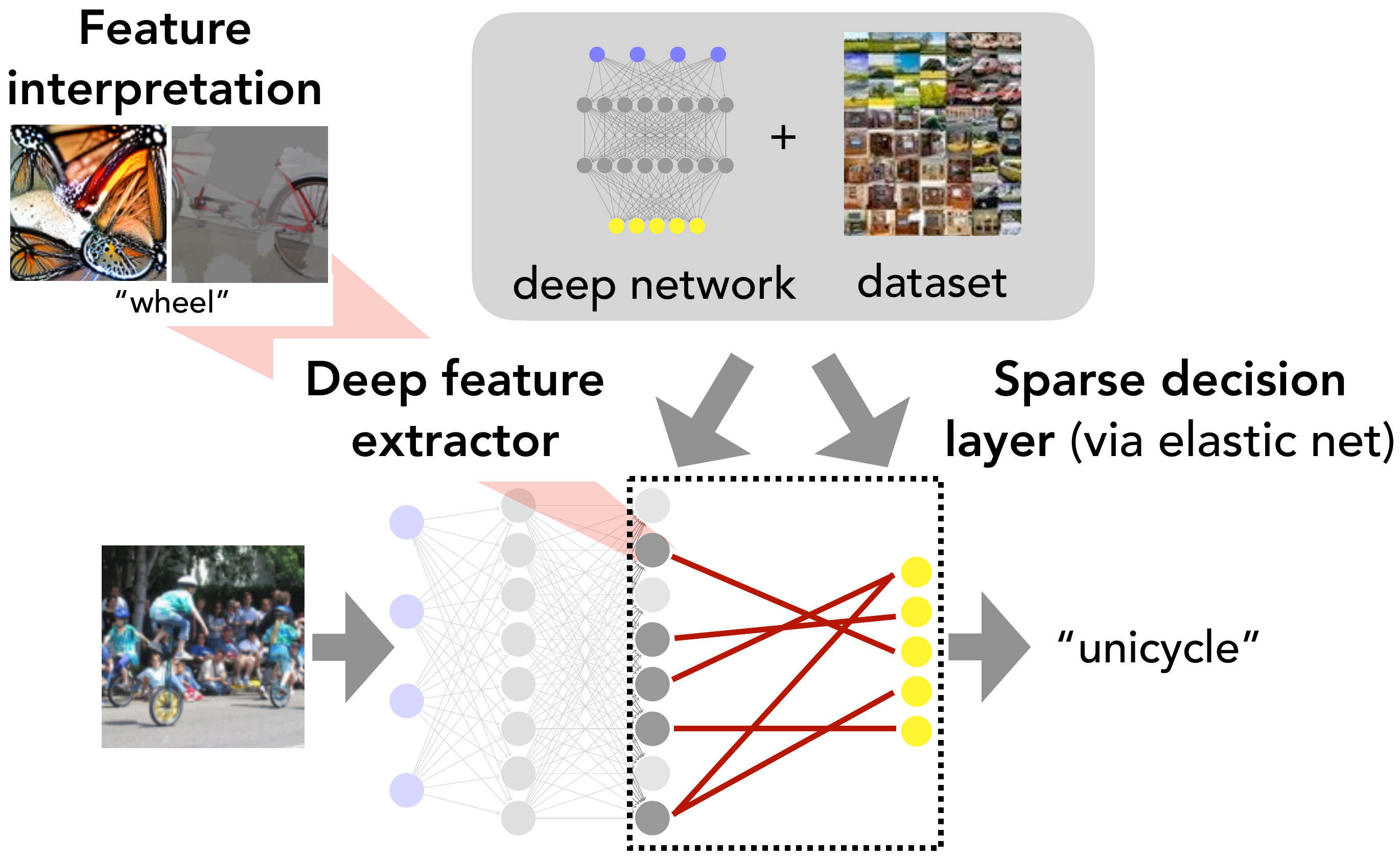}
	\caption{Illustration of our pipeline: For a given 
	task, we 
	construct a \emph{\sparsemod{}} by training a regularized generalized 
	linear 
	model (via 
	elastic net) on the deep feature representations of a pre-trained deep 
	network.
	We then aim to debug model behavior by simply inspecting the few 
	relevant deep features (with existing feature interpretation tools), and 
	the 
	linear coefficients used to aggregate them.
}
	\label{fig:decomposition}
\end{figure}
\else 
\begin{figure}[!t]
	\centering
	\includegraphics[width=0.85\columnwidth]{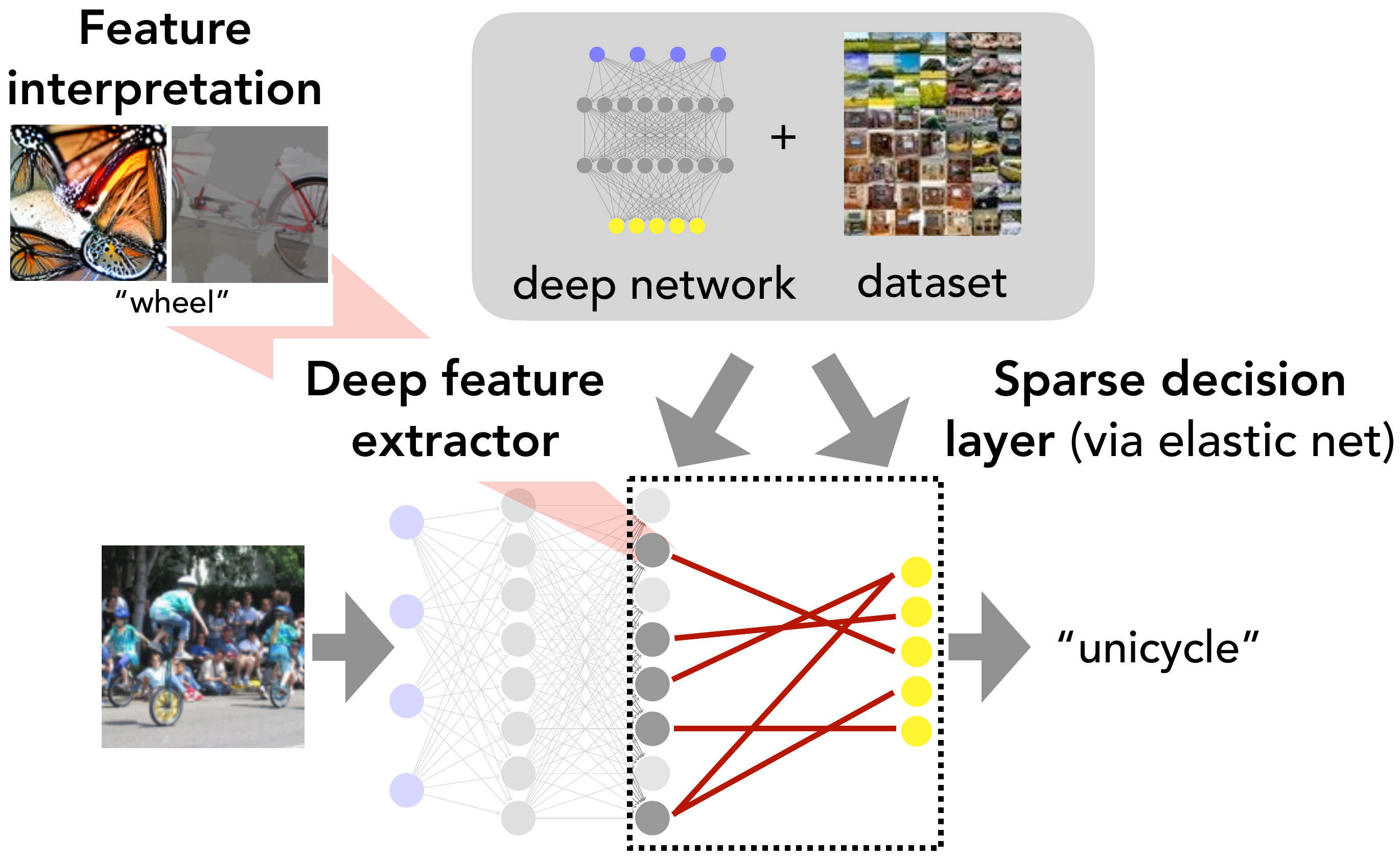}
	\caption{Illustration of our pipeline: For a given 
		task, we 
		construct a \emph{\sparsemod{}} by training a regularized generalized 
		linear 
		model (via 
		elastic net) on the deep feature representations of a pre-trained deep 
		network.
		We then aim to debug model behavior by simply inspecting the few 
		relevant deep features (with existing feature interpretation tools), and 
		the 
		linear coefficients used to aggregate them.
	}
	\label{fig:decomposition}
\end{figure}
\fi

For simplicity, we present an overview of the 
elastic net for linear regression, and defer 
the reader to \citet{friedman2010regularization} for a more complete 
presentation on the generalized linear model (GLM) in the classification 
setting. 
Let $(X,y)$ be the standardized data matrix (mean zero and variance 
one) and output respectively. 
In our setting, $X$ corresponds to the (normalized) deep feature 
representations of input data points, while $y$ is the target. 
Our goal is to fit a sparse linear model of the form $\mathbb{E}(Y|X=x) = x^T\beta + 
\beta_0$. 
Then, the elastic net is the following convex optimization 
problem:
\begin{equation}
\min_\beta \frac{1}{2N}\|X^T\beta + \beta_0 - y\|^2_2 + \lambda R_\alpha(\beta)
\label{eq:elasticnet}
\end{equation}
where
\begin{equation}
R_\alpha(\beta) = (1-\alpha)\frac{1}{2}\|\beta\|_2^2 + \alpha \|\beta\|_1
\end{equation}
is referred to as the elastic net penalty \citep{zou2005regularization} for 
given hyperparameters $\lambda$ and $\alpha$. 
Typical elastic net solvers optimize (\ref{eq:elasticnet}) for a variety of 
regularization strengths $\lambda_1 > \dots > \lambda_k$, resulting in a 
series 
of linear classifiers with weights $\beta_1, \dots, \beta_k$ known as 
the \emph{regularization path}, where
\begin{equation}
\beta_i = \argmin_\beta \frac{1}{2N}\|X^T\beta - y\|^2_2 + \lambda_i 
R_\alpha(\beta)
\label{eq:path}
\end{equation}
In particular, a path algorithm for the elastic net calculates the 
regularization path where sparsity ranges the entire spectrum from the trivial 
zero model ($\beta=0$) to completely dense. 
This regularization path can then be used to select a single linear model  to 
satisfy application-specific sparsity or accuracy thresholds (as 
measured on a validation set). 
In addition, these paths can be used to visualize the evolution of weights 
assigned to specific features as a function of sparsity constraints 
on the model, thereby providing further insight into the relative 
importance of features (cf. Appendix \ref{app:order}).
 
\paragraph{Scalable solver for large-scale elastic net.}
Although the elastic net is widely-used for small-scale GLM problems, 
existing solvers can not handle the scale (number of 
samples and input dimensions) that typically arise in deep learning.
In fact, at such scales, state-of-the-art solvers struggle to solve the elastic 
net even for 
a single regularization value, and 
cannot be directly parallelized due to their reliance on 
coordinate descent~\citep{friedman2010regularization}. 
We remedy this by creating an optimized GLM solver that combines the path 
algorithm of \citet{friedman2010regularization} with recent 
advancements in variance reduced gradient methods 
\citep{gazagnadou2019optimal}.
The speedup in our approach comes from the improved convergence rates of 
these methods over stochastic gradient descent in strongly convex 
settings such as the elastic net. 
Using our approach, we can fit ImageNet-scale regularization paths to 
numerical precision on the order of 
hours on a single GPU (cf. Appendix 
\ref{app:timing} for details). 
\ificml
\begin{figure}[!t]
	\centering
	\includegraphics[width=0.95\columnwidth]{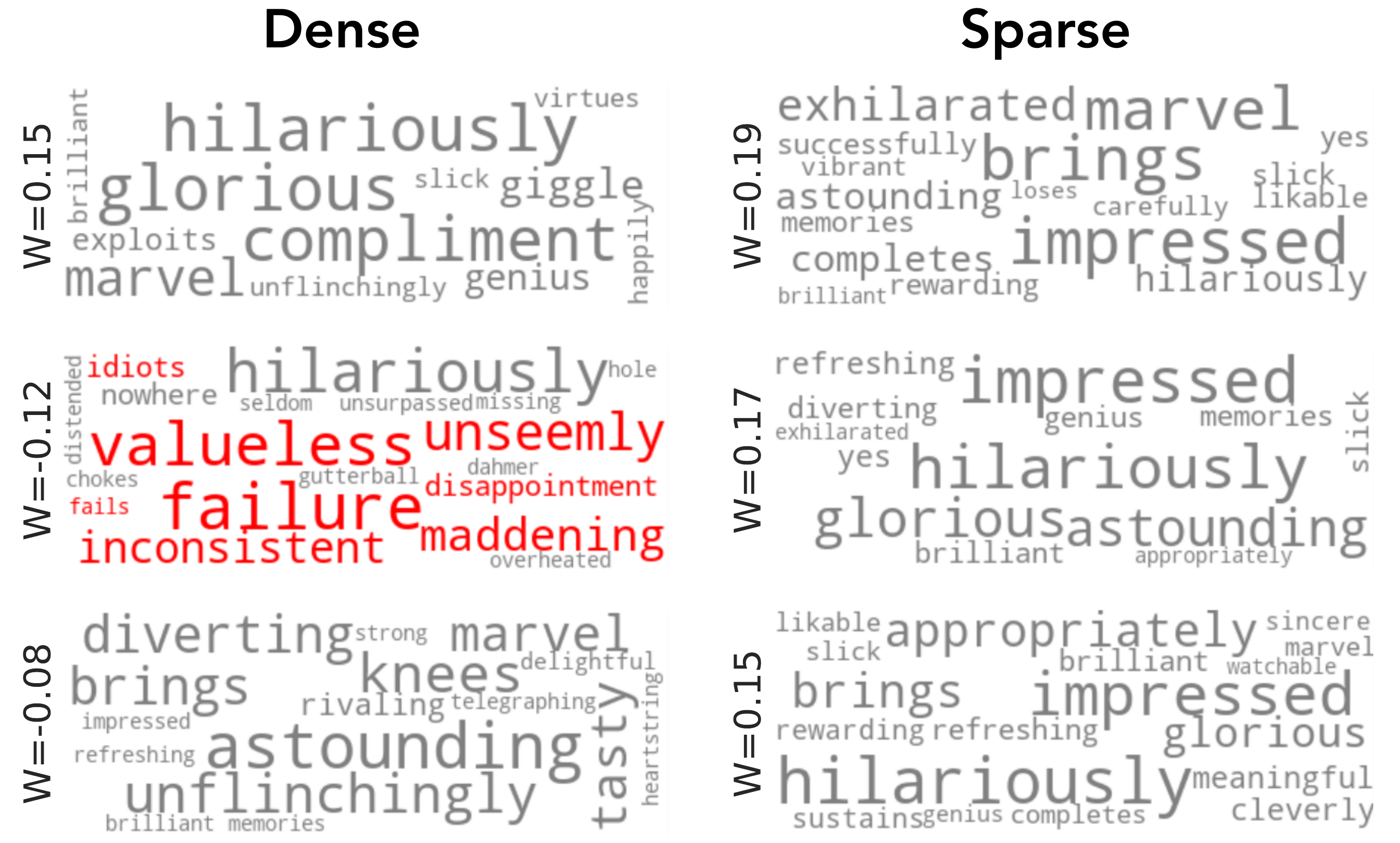}
	\caption{LIME-based word cloud visualizations for the highest weighted 
		features in the (dense/sparse) decision layers of BERT models for 
	   \emph{positive} sentiment detection in the SST 
		dataset. 
		As highlighted in red, some of the key features used by the dense 
		decision layer are actually activated for words with \emph{negative} 
		semantic meaning.}
	\label{fig:suite_nlp}
\end{figure}
\else
\begin{figure}[!t]
	\begin{subfigure}{0.54\textwidth}
		\centering
		\includegraphics[width=1\columnwidth]{figures/glm/features/wordcloud}
		\caption{}
		\label{fig:suite_nlp}
	\end{subfigure}
	\hfil
	\begin{subfigure}{0.44\textwidth}
		\centering
		\includegraphics[width=1\columnwidth]{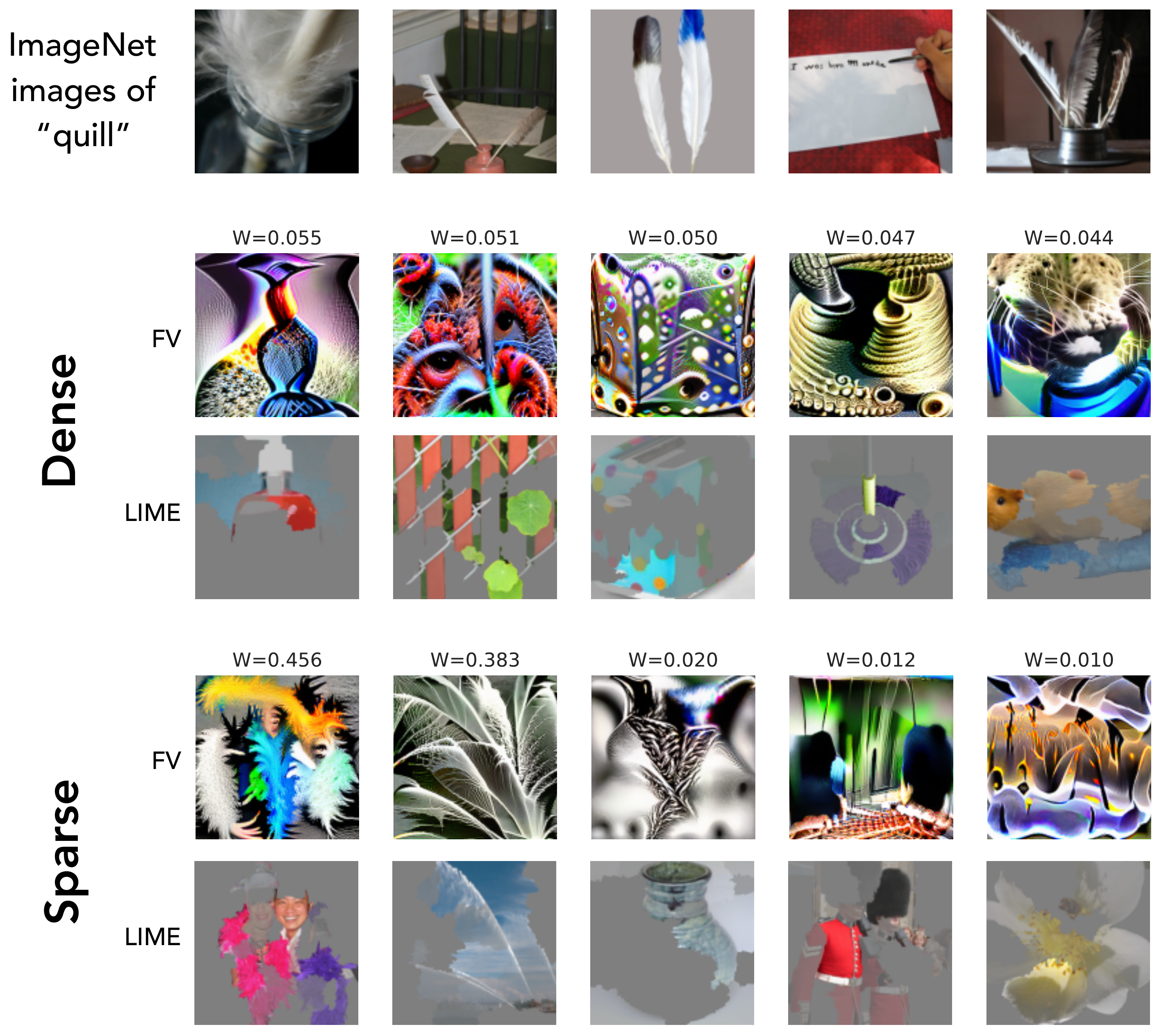}
		\caption{}
		\label{fig:suite}
	\end{subfigure}
	\caption{(a) LIME-based word cloud visualizations for the 
	highest-weighted 
		features in the (dense/sparse) decision layers of BERT models for 
		\emph{positive} sentiment detection in the SST 
		dataset. 
		As highlighted in red, some of the key features used by the dense 
		decision layer are actually activated for words with \emph{negative} 
		semantic meaning. (b) Visualization of deep  
		features used by dense and sparse decision layers of a 
		robust ($\eps=3$) ResNet-50 classifier to detect 
		the ImageNet class ``quill''. Here we present five deep features 
		used 
		by each decision layer, that are 
		randomly-chosen from the top-$k$ 
		highest-weighted ones---where $k$ is the number of features 
		used by the \sparsemod{} for this class. 
		For each (deep) feature, we show its 
		linear coefficient (W),  feature 
		visualization
		(FV) and LIME superpixels.
	}
\label{fig:suite_full}
\end{figure}
\fi

\subsection{Interpreting deep features}
\label{sec:rep_explain}
A sparse linear model allows us to 
reason about the network's decisions in terms of a significantly smaller 
set of deep features. 
When used in tandem with off-the-shelf feature interpretation 
methods, the end result is a simplified description of how the network makes 
predictions.
For our study, we utilize  the following two 
widely-used 
techniques: 

\begin{enumerate}


\item \emph{LIME~\cite{ribeiro2016should}}: Although traditionally used to 
interpret model 
outputs, we use it to understand deep features. We fit a 
local surrogate model around the most activating examples of a 
deep feature to identify 
key ``superpixels'' for images or words for sentences.

\item \emph{Feature visualization~\citep{yosinski2015understanding}}: 
Synthesizes inputs that maximally activate a given 
neuron.\footnote{\label{note1} Despite 
	significant research, feature visualizations for standard vision models are 
	often hard to parse,  
	possibly due to their reliance on human-unintelligible 
	features~\cite{ilyas2019adversarial}. 
	Thus, in the main paper, we present visualizations 
	from adversarially-trained models which tend to have more 
	human-aligned 
	features~\cite{tsipras2019robustness,engstrom2019learning}, 
	and present the corresponding plots for  standard models in 
	Appendix~\ref{app:visualizations}.}  
\end{enumerate}

We detail the visualization procedure in  
Appendix~\ref{app:feature_interpretation}, and present sample 
 visualizations in Figure~\ref{fig:suite_full}. 


\ificml
\begin{figure}[!t]
	\centering
	
\includegraphics[width=0.95\columnwidth]{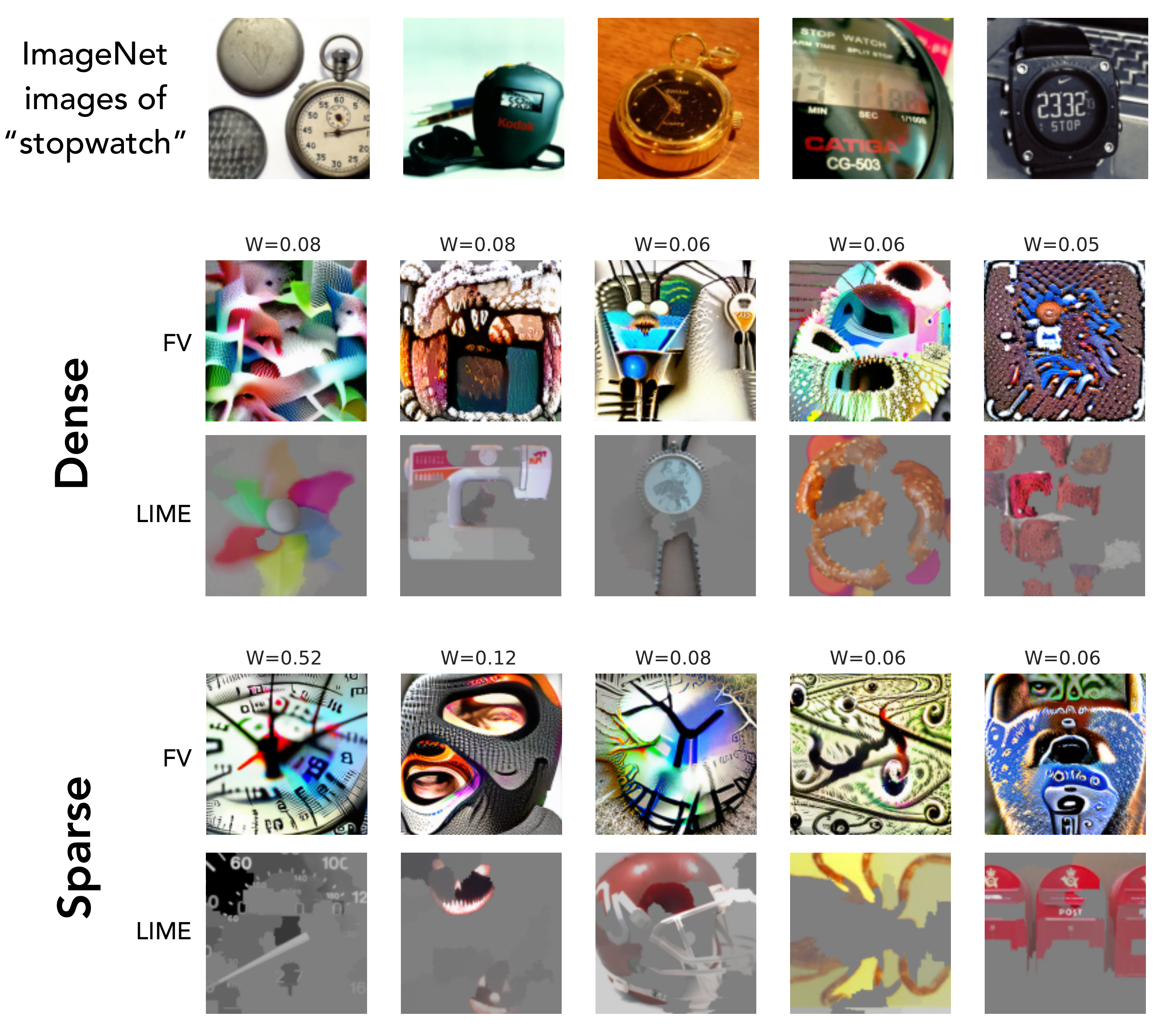}
	\caption{Visualization of five randomly-chosen deep  
		features used by dense and sparse decision layers of a 
		robust\footnotemark[\getrefnumber{note1}] ($\eps=3$) 
		ResNet-50 classifier to detect 
		the ImageNet class ``stopwatch'', along with 
		their 
		linear coefficients (W),  feature 
		visualizations
		(FV) and LIME superpixels.}
	\label{fig:suite}
\end{figure}

\begin{figure*}[!t]
	\centering
	\begin{subfigure}[b]{.32\textwidth
		\centering
		\includegraphics[width=1\columnwidth]{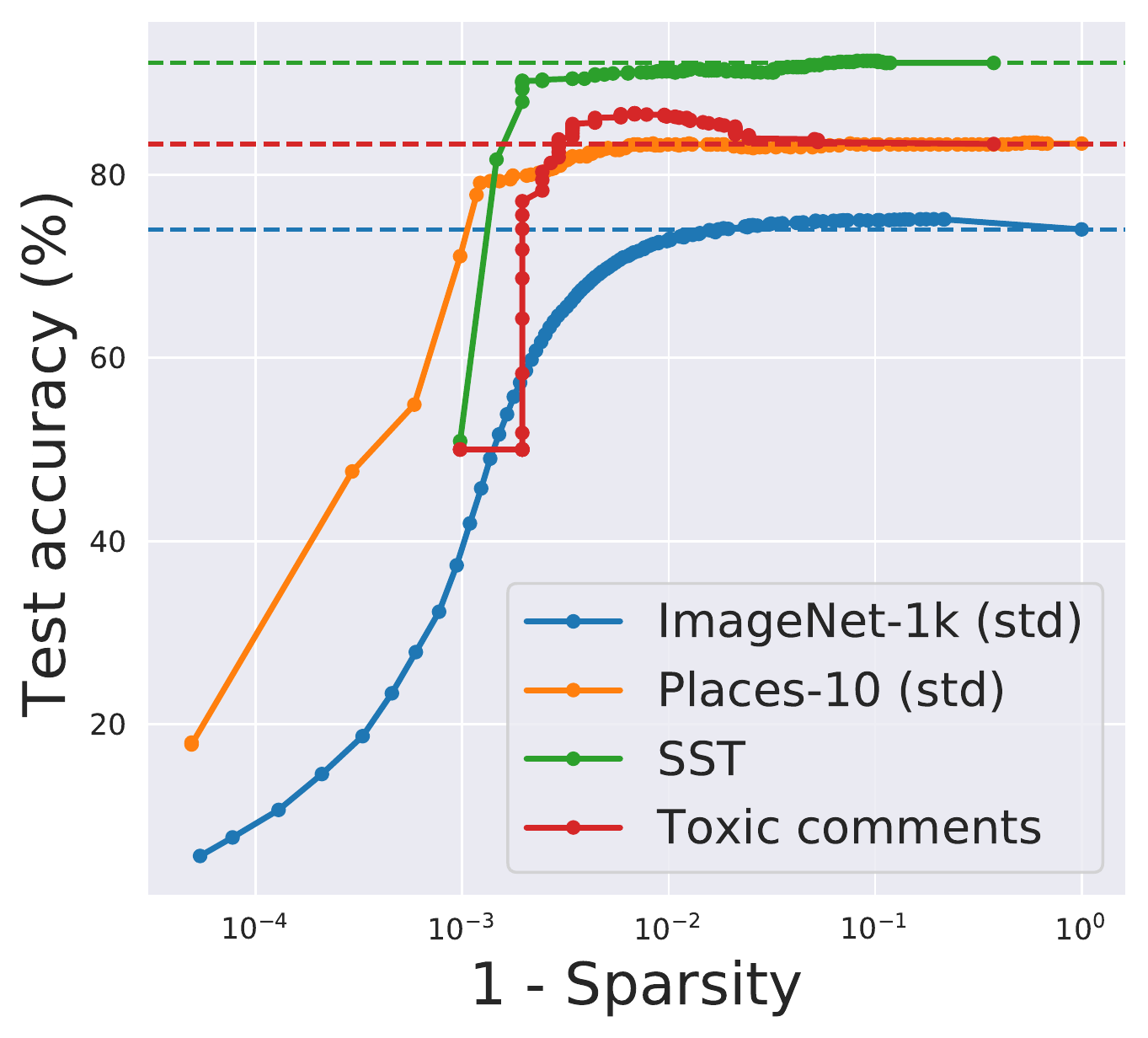}
		\caption{}
		\label{fig:sparsity}
	\end{subfigure} 
\hfill
	\begin{subfigure}[b]{.64\textwidth}
		\begin{tabular}{lccccccc}
			\toprule
			\multirow{2}{*}{Dataset/Model} & &\multicolumn{3}{c}{Dense} & 
			\multicolumn{3}{c}{Sparse} \\
			\cmidrule(lr){3-5}\cmidrule(lr){6-8} 
			& $k$ & All & Top-$k$ & Rest & All & Top-$k$ & Rest \\
			\midrule
			ImageNet (std) & 10 &   74.03 &  58.46 &  55.22 &   72.24 &  69.78 
			&  10.84 
			\\
			ImageNet (robust) & 10 &   61.23 &  28.99 &  34.65 &   59.99 &  
			45.82 &  
			19.83 \\
			Places-10 (std) & 10 &  83.30 &  83.60 &  81.20 &   77.40 &  77.40 &  
			10.00 
			\\
			Places-10 (robust) & 10 &  80.20 &  76.10 &  76.40 &   77.80 &  
			76.60 &  
			40.20 \\
			\midrule
			SST & 5 &  91.51 &  53.10 &   91.28 &  90.37 &  90.37 &   50.92 \\
			Toxic comments  & 5 & 83.33 &  55.35 &  57.87 &   82.47 &  82.33 
			&  50.00 
			\\
			Obscene comments & 5 & 80.41 &  50.03 &  50.00 &   77.32 &  
			72.39 &  
			50.00 \\
			Insult comments & 5 & 72.72 &  50.00 &  50.00 &   77.14 &  75.80 &  
			50.00 \\
			\bottomrule
		\end{tabular}
		\caption{}
		\label{tab:ablation}
	\end{subfigure}
	\caption{({a}): Sparsity vs. accuracy trade-offs of sparse decision layers 
	(cf. Appendix Figure~\ref{fig:app_sparsity} for 
		additional 
		models/tasks). Each point on the curve corresponds to single 
		linear classifier from the regularization path in 
		Equation (\ref{eq:path}). 
		({b}): Comparison of the accuracy of dense/sparse decision layers when 
		they are constrained to utilize only the top-$k$ deep features (based on 
		weight magnitude). We also show overall model accuracy, and the 
		accuracy gained by using the remaining deep features.}
\end{figure*}
\fi

%% file: verification.tex
\section{Are Sparse Decision Layers Better?}
\label{sec:verification}

We now apply our methodology to widely-used deep networks and 
assess the quality of the resulting sparse decision layers along 
a number of axes. We demonstrate that: 

\begin{enumerate}
\item The standard (henceforth referred to as \emph{``dense''}) linear
decision layer can be made highly sparse at only a small cost to performance 
(Section~\ref{sec:sparsity_vs_performance}).
\item The deep features selected by \sparsemod{}s are qualitatively and 
quantitatively better at summarizing the 
model's decision process (Section~\ref{sec:easier}). Note that the 
dense and sparse decision layers operate on the same deep 
features---they only differ in the weight (if any) they assign to each one. 
\item These aforementioned improvements (induced by the \sparsemod) 
translate into better human 
understanding of the 
model
(Section~\ref{sec:human}). 
\end{enumerate}

We perform our analysis on: (a) 
ResNet-50 classifiers~\citep{he2016deep} trained on 
ImageNet-1k~\citep{deng2009imagenet,russakovsky2015imagenet} and 
Places-10 (a 
10-class subset of Places365~\citep{zhou2017places}); and (b) 
BERT~\citep{devlin2018bert} for sentiment classification on 
Stanford Sentiment Treebank (SST)~\citep{socher2013recursive} and 
toxicity classification of Wikipedia comments~\citep{wulczyn2017ex}. Details 
about the setup can be found in 
Appendix~\ref{app:datasets}.

\subsection{Sparsity vs. performance} 
\label{sec:sparsity_vs_performance}

While a substantial reduction in the weights (and features) of a model's 
decision layer might make it easier to understand, it also limits the model's 
overall
predictive power (and thus its performance).
Still, we find that across datasets and architectures, the 
decision layer can be made substantially sparser---by up to two orders of 
magnitude---with a small impact on accuracy (cf. 
Figure~\ref{fig:sparsity}). 
For instance, it is possible to find an accurate decision layer
that relies on only about 20 deep features/class for ImageNet 
(as opposed to 2048 in the dense case).  
Toxic comment classifiers can be sparsified even further (<10 features/class), 
with \emph{improved} generalization over the dense 
decision layer. 

For the rest of our study, we select a single \sparsemod{} to 
balance performance and sparsity---specifically the 
sparsest model whose accuracy is within $5\%$ of top validation set 
performance
(details in 
Appendix~\ref{app:single}). 
However, as discussed previously, these thresholds can be varied based on 
the  needs of specific applications. 

\ificml
\else
\begin{figure*}[!t]
	\centering
		\begin{subfigure}[b]{.31\textwidth}
			\centering
			\includegraphics[width=1\columnwidth]{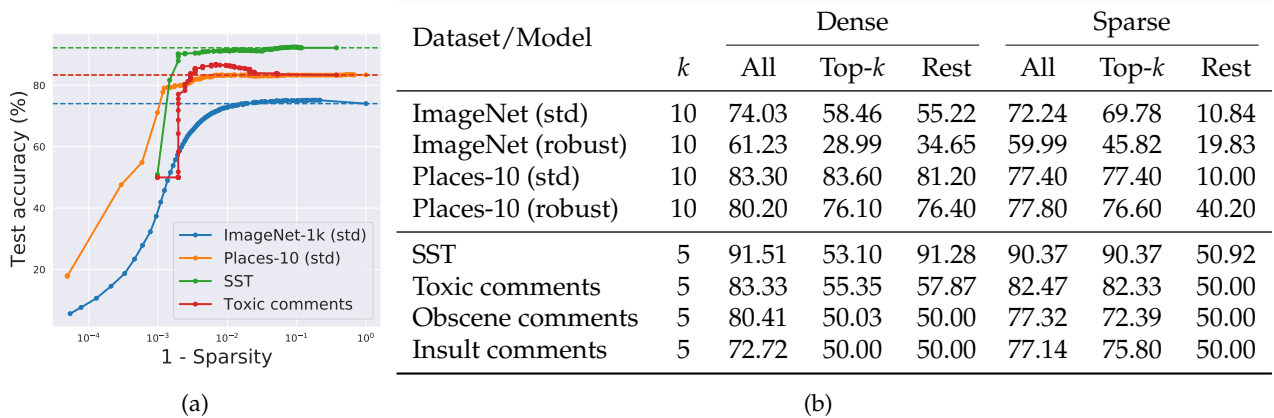}
			\caption{}
			\label{fig:sparsity}
		\end{subfigure} 
		\hfil
			\begin{subfigure}[b]{.68\textwidth}
				\begin{tabular}{lccccccc}
					\toprule
					\multirow{2}{*}{Dataset/Model} & &\multicolumn{3}{c}{Dense} & 
					\multicolumn{3}{c}{Sparse} \\
					\cmidrule(lr){3-5}\cmidrule(lr){6-8} 
					& $k$ & All & Top-$k$ & Rest & All & Top-$k$ & Rest \\
					\midrule
					ImageNet (std) & 10 &   74.03 &  58.46 &  55.22 &   72.24 &  
					69.78 
					&  10.84 
					\\
					ImageNet (robust) & 10 &   61.23 &  28.99 &  34.65 &   59.99 
					&  
					45.82 &  
					19.83 \\
					Places-10 (std) & 10 &  83.30 &  83.60 &  81.20 &   77.40 &  
					77.40 &  
					10.00 
					\\
					Places-10 (robust) & 10 &  80.20 &  76.10 &  76.40 &   77.80 &  
					76.60 &  
					40.20 \\
					\midrule
					SST & 5 &  91.51 &  53.10 &   91.28 &  90.37 &  90.37 &   50.92 
					\\
					Toxic comments  & 5 & 83.33 &  55.35 &  57.87 &   82.47 &  
					82.33 
					&  50.00 
					\\
					Obscene comments & 5 & 80.41 &  50.03 &  50.00 &   77.32 &  
					72.39 &  
					50.00 \\
					Insult comments & 5 & 72.72 &  50.00 &  50.00 &   77.14 &  
					75.80 &  
					50.00 \\
					\bottomrule
				\end{tabular}
				\caption{}
				\label{tab:ablation}
			\end{subfigure}
			\caption{({a}) Sparsity vs. accuracy trade-offs of sparse decision 
			layers 
				(cf. Appendix Figure~\ref{fig:app_sparsity} for 
				additional 
				models/tasks). Each point on the curve corresponds to single 
				linear classifier from the regularization path in 
				Equation (\ref{eq:path}). 
				({b}) Comparison of the accuracy of dense/sparse decision layers 
				when 
				they are constrained to utilize only the top-$k$ deep features 
				(based on 
				weight magnitude). We also show overall model accuracy, and the 
				accuracy gained by using the remaining deep features.}
		\end{figure*}
\fi

\subsection{Sparsity and feature highlighting}
\label{sec:easier}
Instead of sparsifying a network's decision layer, one could consider simply 
focusing on its most prominent deep features for debugging purposes.
In fact, this is the basis of feature highlighting or principal reason 
explanations in the credit industry~\citep{barocas2020hidden}. 
How effective are such feature highlighting explanations at mirroring the 
underlying model?

In Table~\ref{tab:ablation}, we measure the accuracy of the 
dense/sparse decision layer when it is constrained to utilize only the 
top-$k$ (5-10) features by weight magnitude.
For dense decision layers, we consistently find that the 
top-$k$ features do not fully capture the model's performance.
This is in stark contrast to the sparse case, where the top-$k$ features are 
both 
necessary, and to a large extent sufficient, to capture the model's predictive 
behavior.
Note that the  top-$k$ features of the dense decision layers in the language 
setting almost completely fail at near random-chance performance 
($\sim$50\%).
This indicates that there do exist cases where focusing on the most 
important features (by weight) of a dense decision layer provides 
a misleading picture of global model behavior.

\subsection{Sparsity and human understanding}
\label{sec:human}
We now visualize the deep features utilized by the dense and 
sparse 
decision layers to evaluate how amenable they are to human understanding. 
We show representative examples from sentiment classification (SST) and 
ImageNet, and provide additional visualizations in Appendix 
\ref{app:visualizations}. 

Specifically, in Figure~\ref{fig:suite_nlp}, we present word cloud 
interpretations 
of the top three deep features used by both of these decision layers 
for detecting positive sentiment on the SST 
dataset~\cite{socher2013recursive}. 
It is apparent that the sparse decision layer selects features which activate 
for words with positive semantic meaning. 
In contrast, the second  most prominent deep feature for the dense 
decision layer is actually activated by 
words with \emph{negative} semantic meaning. This example highlights how 
the dense decision layer can lead to unexpected features being used for 
predictions.

In Figure~\ref{fig:suite}, we 
present feature interpretations corresponding to the ImageNet class 
``quill''  for both the dense and sparse decision 
layers of a ResNet-50 classifier\footnotemark[\getrefnumber{note1}]. These
feature visualizations seem to suggest that the sparse decision layer focuses more 
on deep features 
which detect salient class characteristics, such as ``feather-like texture'' and 
the ``glass bottle'' in the background. 

\paragraph{Model simulation study}
To validate the perceived differences in the vision setting---and ensure 
they are not due to confirmation biases---we 
conduct a 
human study on 
Amazon Mechanical Turk  (MTurk).
Our goal is to assess how well annotators are able to intuit 
(simulate\footnote{Simulatibility is a standard evaluation criterion in 
interpretability~\cite{ribeiro2016why,lipton2018mythos}, wherein an 
interpretation is deemed to be good if it enables humans to reproduce what 
the model will decide (irrespective of the ``correctness'' of that
decision).}) overall 
model behavior when they are exposed to its decision layer.
To this end, we show annotators five randomly-chosen features used by the 
(dense/sparse) decision layer to recognize objects of a target class, along 
with the corresponding linear coefficients. 
We then present them with three samples from the validation set 
and ask them to choose the one that best matches the target 
class (cf. Appendix Figure~\ref{fig:app_task_sim} for a sample 
task). 
Crucially, annotators are not provided with any information 
regarding the target class, 
and must make their prediction based solely on the visualized features. 

For both the dense and sparse decision layers, we evaluate how 
accurate annotators are on average (over 1000 tasks)---based on whether 
they can correctly identify the image with the highest target class
probability according to the corresponding model. 
For the model with a sparse decision layer, annotators succeed in guessing 
the 
predictions in $63.02 \pm 3.02\%$ of the cases.
In contrast, they are only able to attain $35.61 \pm 3.09\%$ 
accuracy---which is near-chance ($33.33\%$)---for the model with a dense 
decision layer.
Crucially, these results hold \emph{regardless} of whether the correct image 
is actually 
from the target class or not
(see Appendix Table~\ref{tab:app_mturk_sim} for a discussion). 

Note that our task setup precludes annotators from succeeding based on 
any prior knowledge or cognitive biases as we do not provide them with any 
semantic information about the target label, aside from the feature 
visualizations. 
Thus, annotators' success on this task in the sparse setting indicates that 
the sparse decision 
layer is actually effective at reflecting the model's internal
reasoning process.


%% file: diagnosis.tex
\input{bias_table}

\section{Debugging deep networks}
\label{sec:diagnosis}
We now demonstrate how deep networks with 
sparse decision layers can be substantially easier to debug than their 
dense counterparts. We focus on three problems: 
detecting biases, creating counterfactuals, and identifying input patterns 
responsible for misclassifications. 

\input{biases}
\input{counterfactuals}

\input{errors}

%% file: bias_table.tex
\begin{figure*}[t]
	\ificml
	\captionof{table}{Bias detection in language models: using sparse decision 
		layers, we find that Debiased-BERT is \emph{still} disproportionately 
		sensitive to identitity 
		groups---except that it now uses this information as evidence against  
		toxicity. For example, simply adding the 
		word ``christianity'' to clearly toxic sentences flips the prediction of the 
		model to non-toxic (score < 0.5).}
	\label{tab:sst_counterfactual_examples}
	\fi
	\vspace{-0.02in}
	\begin{center}
		\begin{tabular}{lc}
			\toprule
			Toxic sentence & Change in score\\
			\midrule
			DJ Robinsin is \censor{gay as shit}! he \censor{sucks his dick} so 
			much! [+christianity] & $0.52 \rightarrow 0.49$ \\
			\midrule
			Jeez Ed, you seem like a \censor{fucking shitty douchebag} 
			[+christianity] & $0.52 \rightarrow 0.48$\\
			\midrule
			Hey \censor{asshole}, quit removing FACTS from the article 
			\censor{motherfucker}!! [+christianity] & $0.51\rightarrow 0.45$\\
			\bottomrule
		\end{tabular}
	\end{center}
	\ificml 
	\vspace{-0.12in}
	\else
\captionof{table}{Bias detection in language models: Using sparse decision 
	layers, we find that Debiased-BERT is \emph{still} disproportionately 
	sensitive to identitity 
	groups---except that it now uses this information as evidence against  
	toxicity. For example, simply adding the 
	word ``christianity'' to clearly toxic sentences flips the prediction of the 
	model to non-toxic (score < 0.5).}
\label{tab:sst_counterfactual_examples}
\fi
\end{figure*}

%% file: biases.tex
\subsection{Biases and (spurious) correlations}
\label{sec:biases}
Our first debugging task is to automatically identify unintended biases 
or correlations that deep networks extract from 
their training data.

\paragraph{Toxic comments.} We start by examining 
two BERT 
models trained to classify comments according to toxicity: 
(1) Toxic-BERT, a high-performing 
model that was later found to use identity groups as evidence for toxicity, 
and (2) Debiased-BERT, which was trained to 
mitigate this bias~\citep{borkan2019nuanced}. 

We find 
that Toxic-BERT models with 
sparse decision layers also rely on identity groups to predict comment 
toxicity (visualizations in Appendix~\ref{app:toxic} 
are censored).
Words related to nationalities, religions, and sexual identities 
that are not inherently toxic occur frequently and prominently, and 
comprise 27\% of the word clouds shown for features that detect toxicity.
Note that although the standard Toxic-BERT model is known to be 
biased, this bias is not as apparent in the deep features used 
by its (dense) decision layer (cf. Appendix~\ref{app:toxic}).
In fact, measuring the bias in the standard
model required collecting identity and 
demographic-based subgroup labels~\citep{borkan2019nuanced}.


We can similarly inspect the word clouds for the Debiased-BERT 
model with \sparsemod s
and corroborate that identity-related words no longer appear as evidence for 
toxicity. 
But rather than ignoring these words completely, it turns out that this 
model uses 
them as strong evidence \emph{against} toxicity. 
For example, identity words comprise 
43\% of the word clouds of features detecting non-toxicity. 
This suggests that the debiasing intervention proposed in 
\citet{borkan2019nuanced} may not have had the intended 
effect---Debiased-BERT is still disproportionately  sensitive to identity 
groups, albeit in the opposite way.

We confirm that this is an issue with Debiased-BERT via a simple experiment: 
we 
take toxic sentences that this model (with a sparse decision layer) correctly 
labels as toxic, and 
simply append an identity 
related word (as suggested by our word clouds) to the end---see  
Table~\ref{tab:sst_counterfactual_examples}.
This modification turns out to strongly impact model 
predictions: for example, just adding ``christianity'' to the end of toxic 
sentences flips the prediction to non-toxic 74.4\% of the time. 
We note that the biases diagnosed via sparse decision layers are also 
relevant for the standard Debiased-BERT model. 
In particular, the same toxic sentences with the word ``christianity'' are 
classified as non-toxic 62.2\% of the time by the standard 
model, even though this sensitivity is not as readily apparent from inspecting its 
decision layer (cf. Appendix~\ref{app:toxic}).

\paragraph{ImageNet.}
We now move to the vision setting, with the goal of detecting spurious 
feature dependencies in ImageNet classifiers. 
Once again, our approach is based on the following observation:  
input-class 
correlations learned by a model can be described as the data patterns 
(e.g., 
"dog ears" or "snow") that 
activate deep features used to recognize objects of that 
class, according to the decision layer.

Even so, it is not clear how to identify such patterns for image data, without 
access to fine-grained annotations describing image content. 
To this end, we rely on a human-in-the-loop approach (via MTurk).
Specifically, for a deep feature of interest---used by the \sparsemod{} to 
detect a target class---annotators are shown examples of images 
that activate it.
Annotators are then asked if these ``prototypical''  images have a shared 
visual pattern, 
and if so, to describe it using free-text.

However, under this setup, presenting annotators with images from the 
target 
class alone 
can be problematic. After all, these images are likely to have multiple 
visual patterns in common---not all of which cause the deep feature 
to activate. Thus, to disentangle the pertinent data pattern, we present 
annotators with prototypical  images drawn from more than one classes. 
A sample task is presented in Appendix Figure~\ref{fig:app_task_spurious}, 
wherein annotators see three highly-activating images for a specific deep 
feature 
from two different classes, along with the respective class labels.
Aside from asking annotators to validate (and describe) the presence of a 
shared pattern between these images, we also ask them whether the 
pattern (if present) is part of each class object 
(non-spurious 
correlation) or its surroundings (spurious correlation)\footnote{We focus on 
this specific notion of ``spurious correlations'' as it is easy for humans to 
verify---cf. Appendix
\ref{app:spurious} for details.}.

We find that annotators are able to identify a significant number of 
correlations that standard 
ImageNet classifiers  rely on (cf. 
Table~\ref{tab:tab_mturk_spurious}). 
Once again, sparsity seems to aids the detection of such correlations. Aside 
from having fewer (deep) feature dependencies per class, it turns 
out that annotators are able to pinpoint the (shared) data patterns that 
trigger the relevant deep features in 20\% more cases for the model with a 
\sparsemod{}. 
Interestingly, the fraction of detected patterns that annotators deem 
spurious is lower for the sparse case. 
In Figure~\ref{fig:spurious_img}, we present examples of detected
correlations with annotator-provided descriptions as word clouds (cf. 
Appendix \ref{app:imagenet_biases} for additional examples). 
A global word cloud visualization of correlations identified
by annotators is shown in Appendix Figure~\ref{fig:app_feedback}. 

\ificml
\begin{table}[!h]
	\caption{The percentage of class-level correlations identified using our 
		MTurk setup, along with a breakdown of whether annotators believe the 
		pattern 
		to be ``non-spurious'' (i.e., part of the object) or ``spurious'' (i.e., part 
		of 
		the surroundings). }
	\label{tab:tab_mturk_spurious}
	\begin{center}
		\begin{tabular}{ccc} 
			\toprule
			Detected patterns (\%) & Dense & \normalfont{Sparse}\\ 
			\midrule
			Non-spurious & 18.43 $\pm$ 2.48 & 34.43 $\pm$ 3.38  \\
			Spurious & 9.56 $\pm$ 1.76 & 12.49 $\pm$ 2.02 \\ \midrule
			Total & 27.85 $\pm$ 2.70 & 46.97 $\pm$ 3.15  \\
			\bottomrule
		\end{tabular}
	\end{center}
\end{table}
\begin{figure}[!h]
	\centering
	\includegraphics[width=0.95\columnwidth]{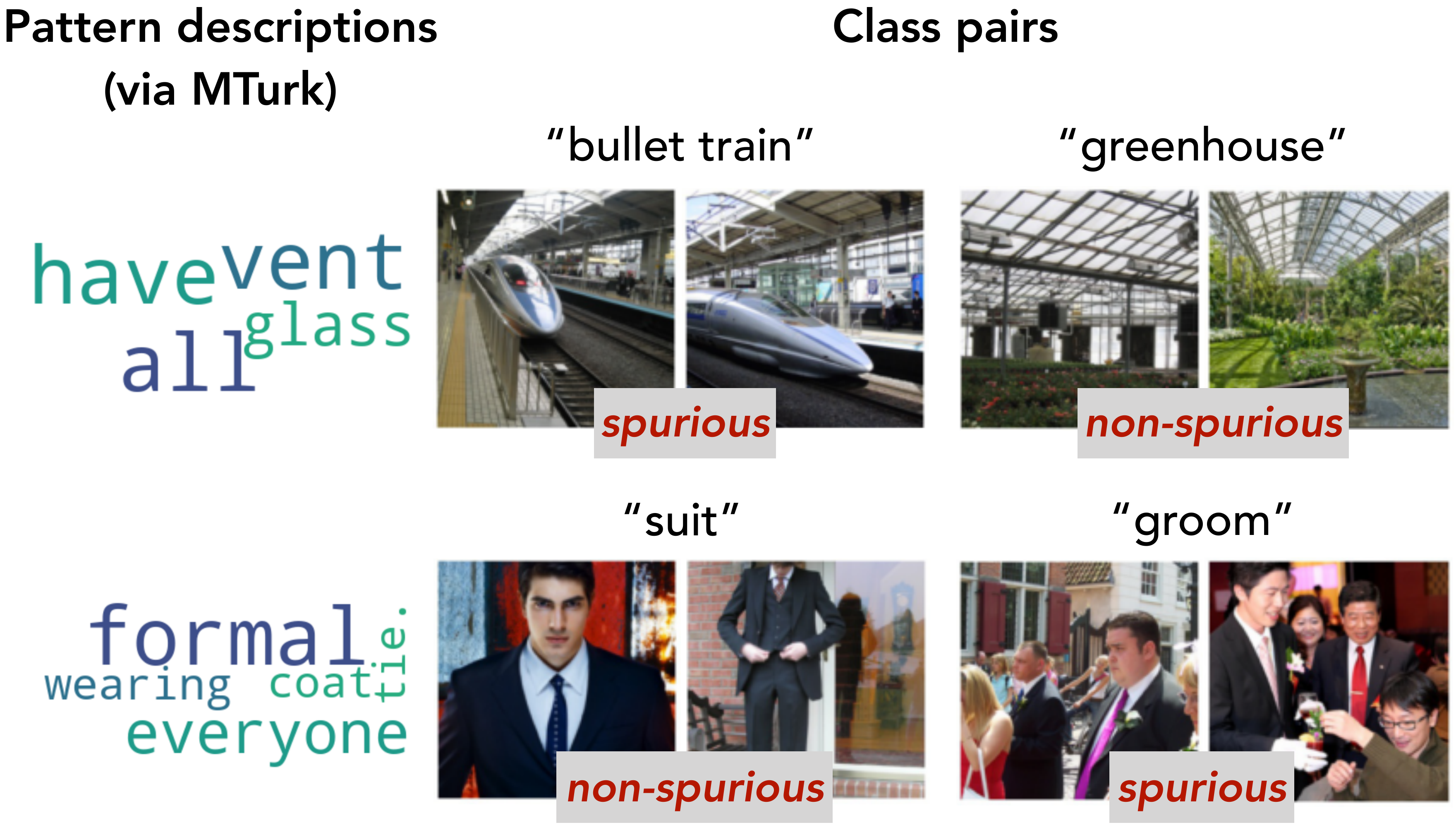}
	\caption{Examples of correlations in ImageNet models detected 
	using our MTurk study. 
	Each row contains protypical images from a pair of classes, along 
	with the annotator-provided descriptions for the shared deep feature that 
	these images strongly activate.
	For each class, we also display if annotators marked the feature to be a 
	``spurious correlation''.
	}
	\label{fig:spurious_img}
\end{figure}
\else
\begin{figure}[!t]
	\begin{subfigure}{0.32\textwidth}
		\begin{tabular}{ccc} 
			\toprule
			Patterns (\%) & Dense & \normalfont{Sparse}\\ 
			\midrule
			Non-spurious & 18.43 $\pm$ 2.48 & 34.43 $\pm$ 3.38  \\
			Spurious & 9.56 $\pm$ 1.76 & 12.49 $\pm$ 2.02 \\ \midrule
			Total & 27.85 $\pm$ 2.70 & 46.97 $\pm$ 3.15  \\
			\bottomrule
		\end{tabular}
		\caption{}
		\label{tab:tab_mturk_spurious}
	\end{subfigure}
	\hfill
	\begin{subfigure}{0.61\textwidth}
		\centering
		\includegraphics[width=0.88\columnwidth]{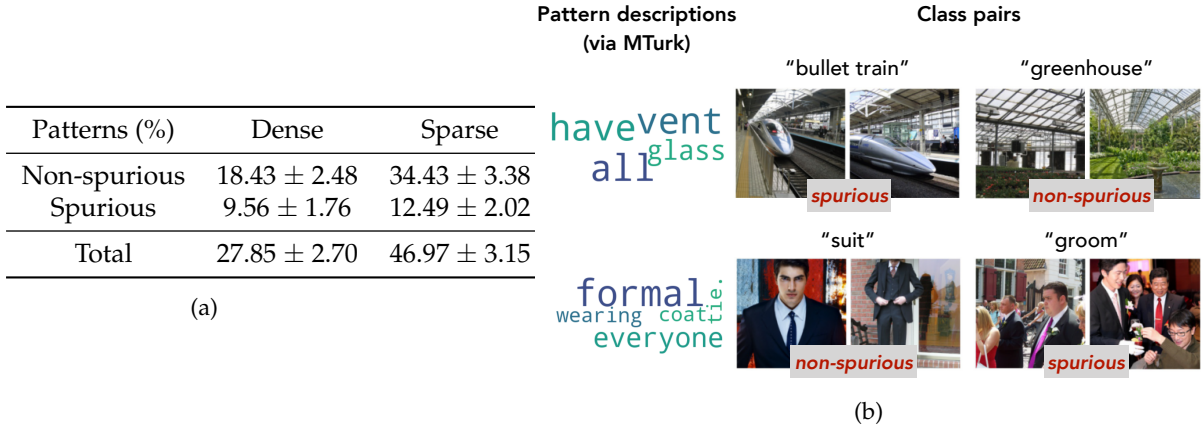}
		\caption{
		}
		\label{fig:spurious_img}
	\end{subfigure}
	\caption{(a) The percentage of class-level correlations identified using our 
		MTurk setup, along with a breakdown of whether annotators believe the 
		pattern 
		to be ``non-spurious'' (i.e., part of the object) or ``spurious'' (i.e., part 
		of 
		the surroundings). (b) Examples of correlations in ImageNet models 
		detected 
		using our MTurk study. 
		Each row contains protypical images from a pair of classes, along 
		with the annotator-provided descriptions for the shared deep feature 
		that 
		these images strongly activate.
		For each class, we also display if annotators marked the feature to be a 
		``spurious correlation''.}
\end{figure}
\fi


%% file: counterfactuals.tex
\subsection{Counterfactuals}
\label{sec:counterfactuals}

A natural way to probe model behavior is by trying to find small input
modifications which cause the model to change its prediction.
Such modified inputs, which are (a special case of) \emph{counterfactuals}, 
can 
be 
a useful primitive for pinpointing input features that the model 
relies on.  
Aside from debugging, such counterfactuals can also be used to  
provide users with recourse~\cite{ustun2019actionable} that can guide them to
obtaining better outcomes in the future. 
We now leverage the deep features used by sparse decision layers to inform 
counterfactual generation.

\input{counterfactual_tab}
\paragraph{Sentiment classifiers.}
Our goal here is to automatically identify word substitutions that 
can be made within a given sentence to flip the sentiment label assigned by 
the model.
We do this as follows: given a sentence with a positive sentiment prediction, 
we first identify the set of deep features 
used by the \sparsemod{} that are 
positively activated for any word in the sentence.
For a randomly chosen deep feature from this pool, we then substitute the 
positive word from the sentence with its negative counterpart.
This substitute word is in turn randomly chosen from the set of  words that 
negatively activate the same deep feature (based on its word cloud).
An example of the positive and 
negative word clouds for one such deep feature is shown in Figure 
\ref{fig:wordclouds}, 
and the resulting counterfactuals are in Table 
\ref{tab:sentiment_counterfactuals} 
(cf. Appendix \ref{app:sentiment_counterfactuals} for details). 

Counterfactuals generated in 
this manner successfully flip the sentiment label assigned by the 
\sparsemod{} 
$73.1\pm 3.0\%$ of the time.
In contrast, such counterfactuals only have $52.2\pm 4\%$ efficacy 
for the dense decision layer.
This highlights that for models with sparse decision layers, it can be  
easier to automatically identify deep features that are causally-linked to  
model predictions.

\paragraph{ImageNet.}
We now leverage the annotations collected in Section~\ref{sec:biases} 
to generate counterfactuals for ImageNet classifiers. Concretely, we 
manually modify images to add or subtract input patterns identified by 
 annotators and verify that they successfully flip the 
model's prediction.
Some representative examples are shown in 
Figure~\ref{fig:counterfactuals_img}. 
Here, we alter images from various ImageNet classes to have the 
pattern ``chainlink fence'' and ``water'', so as to fool the \sparsemod{} 
into recognizing them as ``ballplayers'' and ``snorkels'' respectively.
We find that we are able to 
consistently change the prediction of the \sparsemod{} (and 
in some cases its dense counterpart) by adding a pattern that was 
previously identified (cf. Section~\ref{sec:biases}) to be a spurious 
correlation.

%% file: counterfactual_tab.tex
\begin{figure*}[!t]
    \centering
    \begin{subfigure}[b]{.26\textwidth}
        \centering
        \begin{tabular}{@{}c@{}c@{}}
            \rotatebox[origin=c]{90}{Positive} & 
            \includegraphics[align=c,width=0.8\columnwidth]{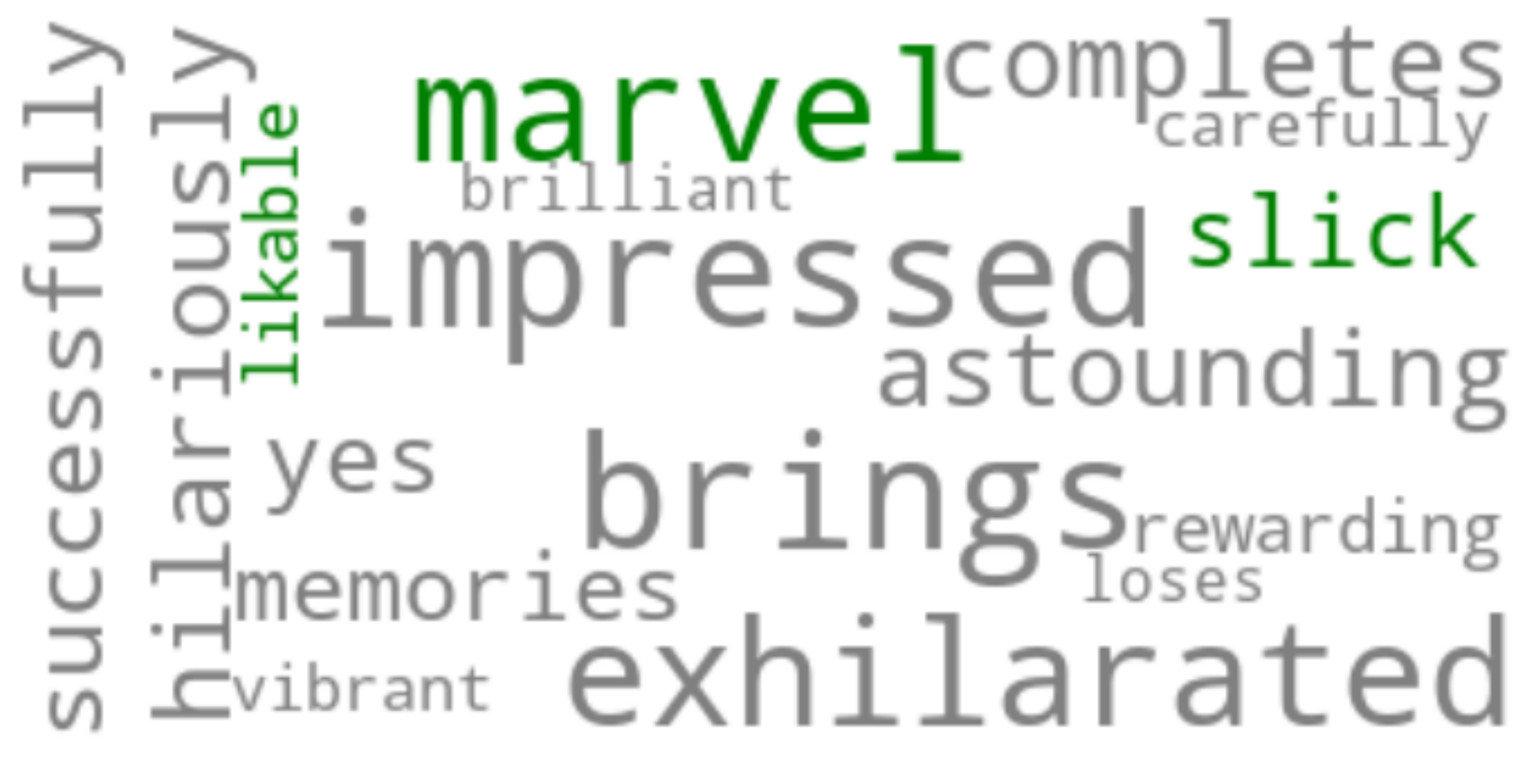}
             \\
            \rotatebox[origin=c]{90}{Negative} & 
            \includegraphics[align=c,width=0.8\columnwidth]{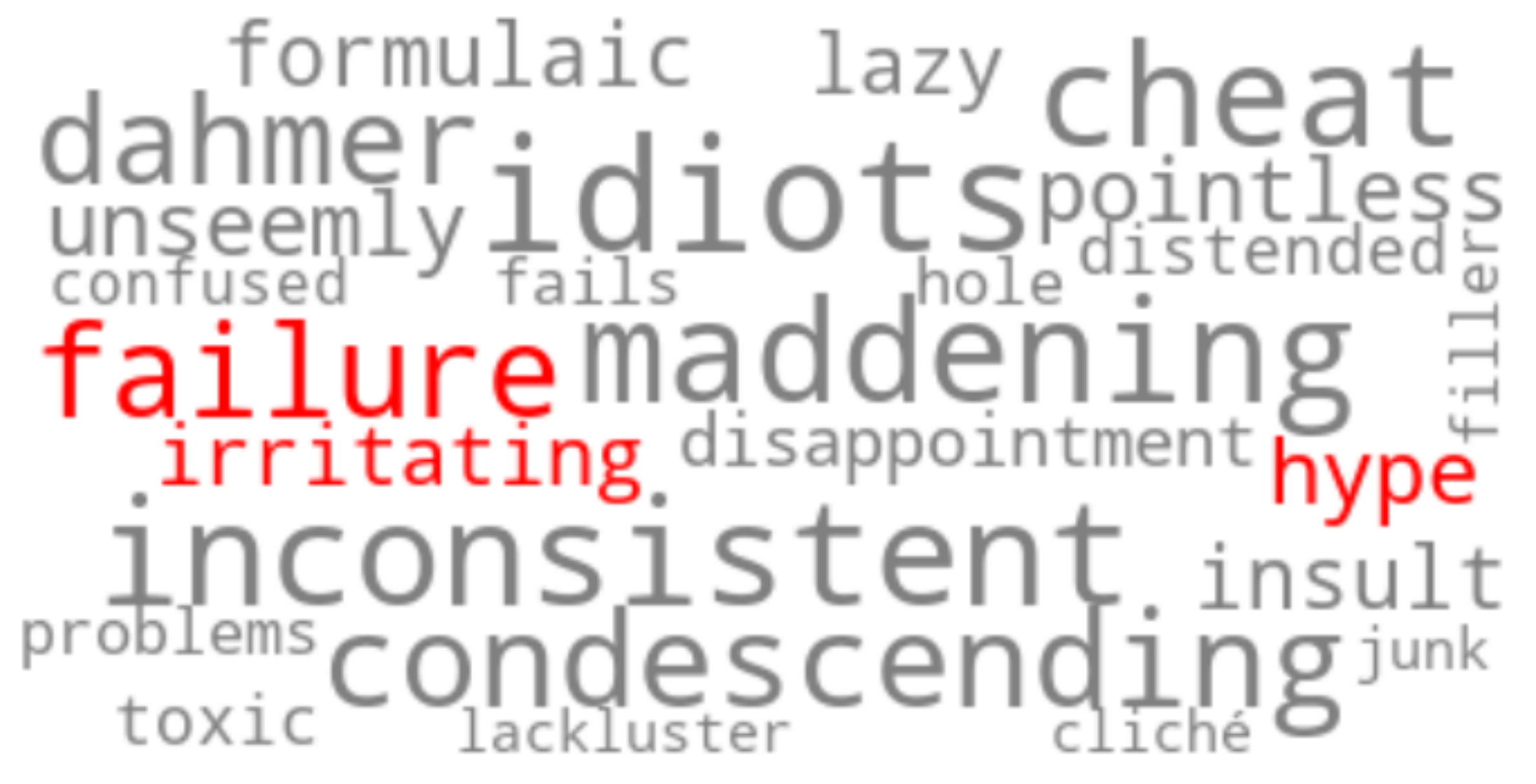}
        \end{tabular}
        \caption{}
        \label{fig:wordclouds}
    \end{subfigure} \hfill
    \begin{subfigure}[b]{.68\textwidth}
        \begin{tabular}{p{3.5cm}p{3.5cm}c}
            \toprule
            Original sentence & Counterfactual & Change in score \\
            \midrule 
            ...something \emph{likable} about the marquis... & 
            ...something \emph{irritating} about the marquis... & 
            $0.73 \rightarrow 0.34$\\
            \midrule 
            \emph{Slick} piece of cross-promotion & \emph{Hype} piece of 
            cross-promotion & $0.73\rightarrow 0.34$\\
            \midrule 
            A \emph{marvel} like none you've seen & A \emph{failure} like none 
            you've seen & $0.73 \rightarrow 0.31$\\
            \bottomrule
        \end{tabular}
        \caption{}
        \label{tab:sentiment_counterfactuals}
    \end{subfigure}
    \caption{({a}): Word cloud visualization for tokens that are 
    positively/negatively correlated with the activation of a particular 
        deep feature. 
        ({b}): Using the wordclouds from ({a}), we can make word substitutions (as highlighted in green and red) to generate counterfactuals that change the model's predicted 
        sentiment (scores below 0.5 are 
        predicted as negative).}
\end{figure*}

%% file: errors.tex
\subsection{Misclassifications}
\label{sec:errors}

Our final avenue for diagnosing unintended behaviors in models is through  
their misclassifications.
Concretely, given an image for which the model makes an incorrect prediction
(i.e., not the ground truth label as per the dataset), our goal is to pinpoint 
some aspects of the image that led to this error.


In the ImageNet setting, it turns out that over 30\% of 
misclassifications made by the 
sparse decision layer can be attributed to a single deep feature---i.e., 
manually setting this ``problematic'' feature to zero fixes the erroneous 
prediction.
For these instances, can humans understand why the problematic feature 
was triggered in the first place?
Specifically, can they recognize the \emph{pattern} in the input that
caused the error?

\ificml
\begin{figure}[t]
	\centering
	\includegraphics[width=0.95\columnwidth]{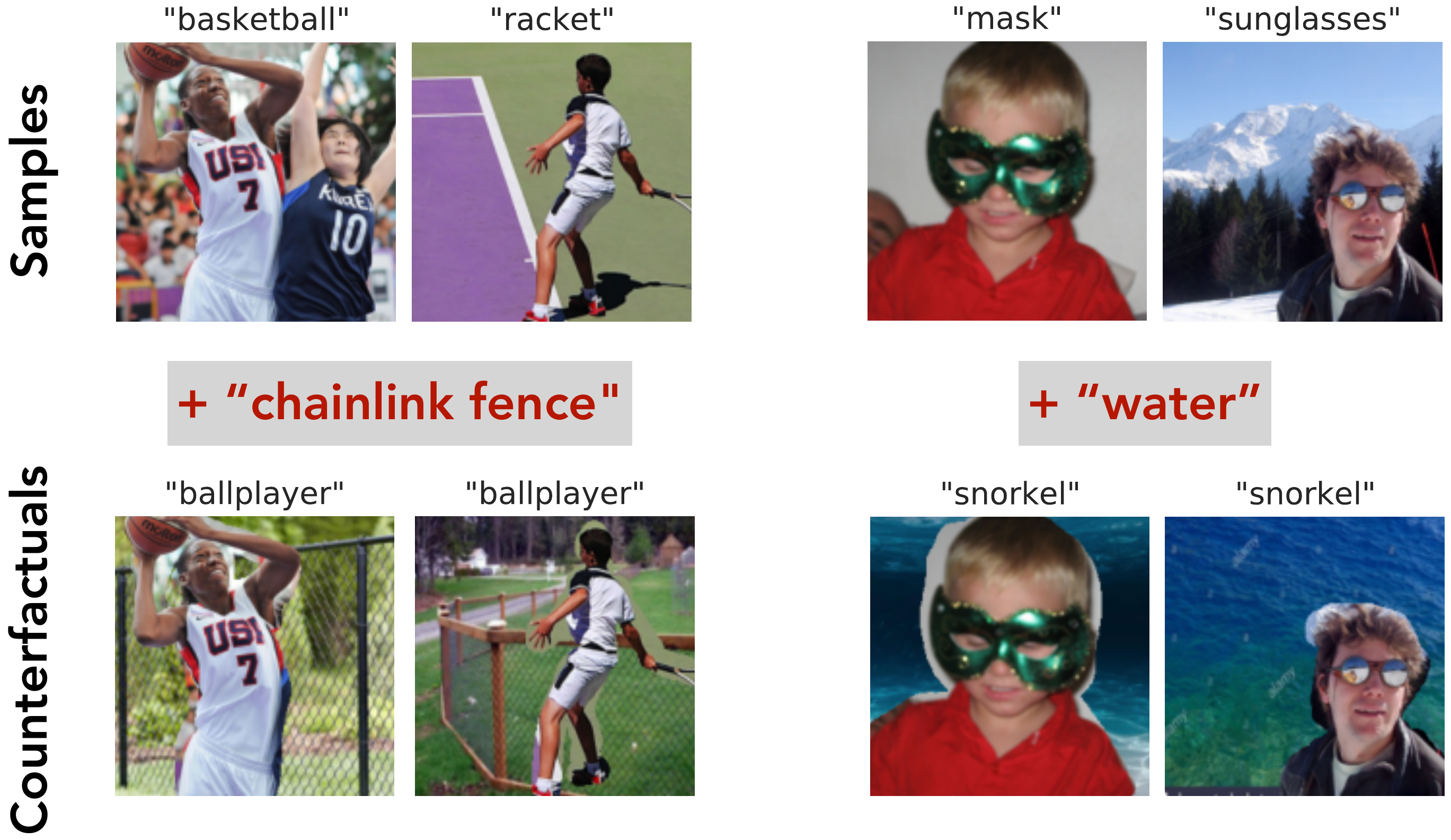}
	\caption{Counterfactual images for 
		ImageNet. We manually modify samples 
		(\emph{top row}) to contain the patterns ``chainlink fence'' and 
		``water'', 
		which annotators deem (cf. Section~\ref{sec:biases}) to be spuriously 
		correlated with the classes 
		``ballplayer'' and ``snorkel'' respectively. 
		We 
		find that these 
		counterfactuals (\emph{bottom row})  succeed in flipping 
		the prediction of the model with a 
		sparse decision layer to 
		the desired class.
	}
	\label{fig:counterfactuals_img}
\end{figure}
\else
\begin{figure}[t]
		\begin{subfigure}{0.5\textwidth}
	\centering
	\includegraphics[width=0.95\columnwidth]{figures/glm/mturk_spurious/counterfactuals}
	\caption{
	}
	\label{fig:counterfactuals_img}
\end{subfigure}
\hfill
	\begin{subfigure}{0.5\textwidth}
		\centering
		\includegraphics[width=0.8\columnwidth]{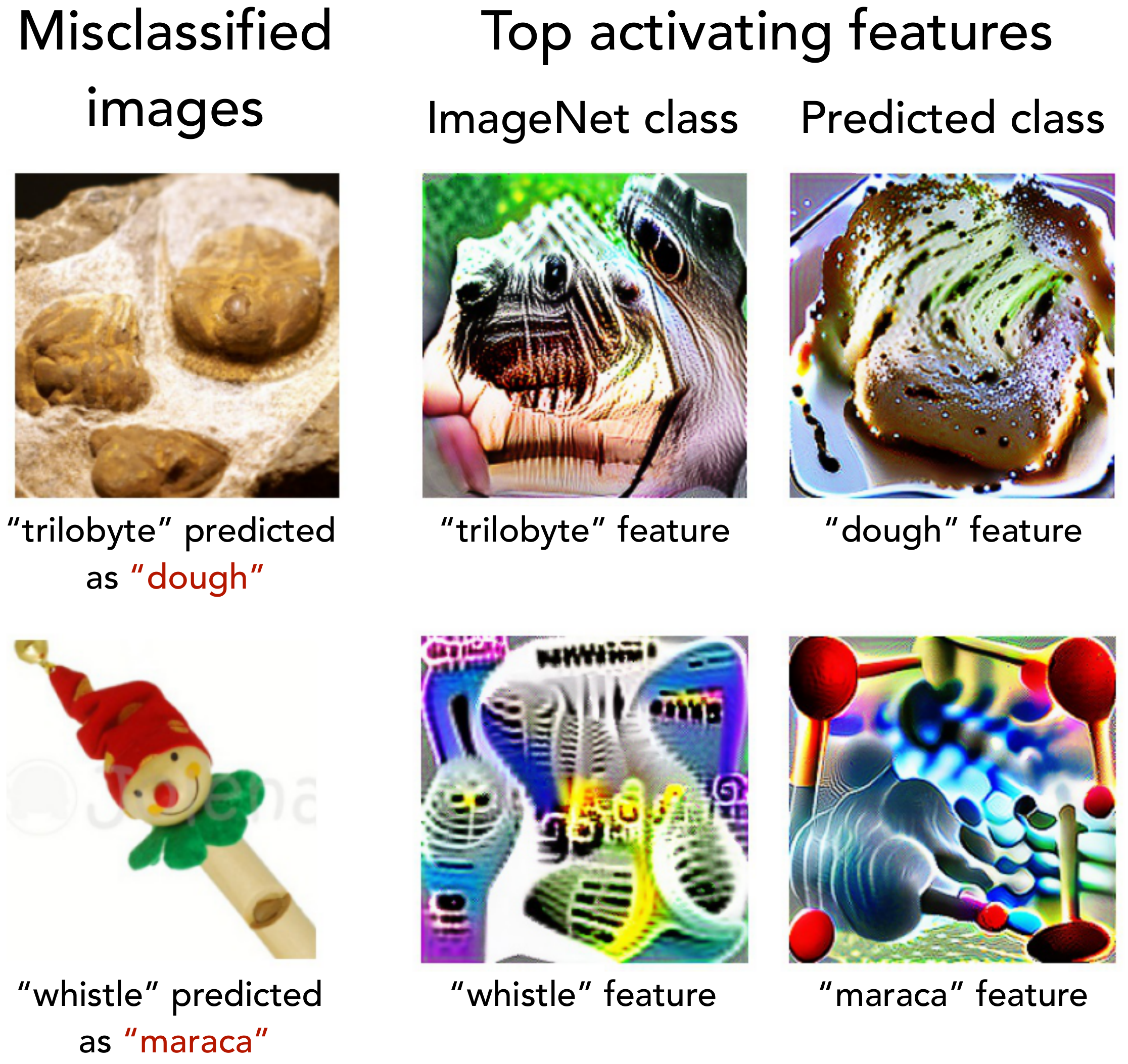}
		\caption{}
		\label{fig:misclassification}
	\end{subfigure}
\caption{(a) Counterfactual images for 
	ImageNet. We manually modify samples 
	(\emph{top row}) to contain the patterns ``chainlink fence'' and 
	``water'', 
	which annotators deem (cf. Section~\ref{sec:biases}) to be spuriously 
	correlated with the classes 
	``ballplayer'' and ``snorkel'' respectively. 
	We 
	find that these 
	counterfactuals (\emph{bottom row})  succeed in flipping 
	the prediction of the model with a 
	sparse decision layer to 
	the desired class. (b) Examples of misclassified ImageNet images for which 
	annotators 
	deem the top activated feature for the predicted class 
	(\emph{rightmost column}) 
	as a better match than the top activated feature 
	for the ground truth class (\emph{middle column}).}
\end{figure}
\fi

To test this, we present annotators on MTurk with misclassified images.
Without divulging the ground truth or predicted labels, we 
show annotators the top activated feature for 
each of the two classes via feature visualizations.
We then ask annotators to select the patterns (i.e., feature visualizations) 
that match the 
image, and to choose one that is a better 
match for the image (cf. Appendix~\ref{app:mturk_mis} for details).
As a control, we repeat the same task but replace the problematic feature 
with a randomly-chosen one.

For about 70\% of the misclassified images, annotators select 
the top feature for the predicted class as being present in the image (cf. 
Table~\ref{tab:tab_mturk_mis}).
In fact, annotators consider it a better match than the feature for the 
ground truth class 60\% of the time.
In contrast, they rarely select randomly-chosen features to be 
present in the image. 
Since annotators do not know what the underlying classes are, 
the high fraction of selections for the problematic feature indicates 
that annotators actually believe this pattern is present in the image.

We present sample misclassifications validated by annotators in 
Figure~\ref{fig:misclassification}, along with the problematic features that led 
to them.
Having access to this information can guide improvements in both 
models and datasets. For instance, model designers might consider 
augmenting the 
training data with examples of ``maracas'' without ``red tips'' to correct the 
second error in Figure~\ref{fig:misclassification}.
In Appendix~\ref{app:confusion}, we further discuss how sparse 
 decision layers can provide insight into inter-class model confusion 
matrices.

\begin{table}[!t]
	\ificml
	\caption{Fraction of misclassified images for which annotators select the 
	top feature of the predicted class to: (i) match the given image and (ii) be 
	a better match than the top feature for the ground truth class. As a 
	baseline, we also evaluate annotator selections when the top feature for 
	the predicted class is replaced by a randomly-chosen one.}
	\label{tab:tab_mturk_mis}
	\fi
	\begin{center}
		\begin{tabular}{ccc} 
			\toprule
			Features & Matches image & Best match \\ 
			\midrule
			Prediction & 70.70\% $\pm$ 3.62\% & 60.12\% $\pm$ 3.77\% \\ 
			Random & 16.63\% $\pm$ 2.91\% & 10.58\% $\pm$ 2.35\%  \\
			\bottomrule
		\end{tabular}
	\end{center}
	\ificml \else
\caption{Fraction of misclassified images for which annotators select the 
	top feature of the predicted class to: (i) match the given image and (ii) be 
	a better match than the top feature for the ground truth class. As a 
	baseline, we also evaluate annotator selections when the top feature for 
	the predicted class is replaced by a randomly-chosen one.}
\label{tab:tab_mturk_mis}
\fi
\end{table}

\ificml
\begin{figure}[t]
	\centering
	\includegraphics[width=0.72\columnwidth]{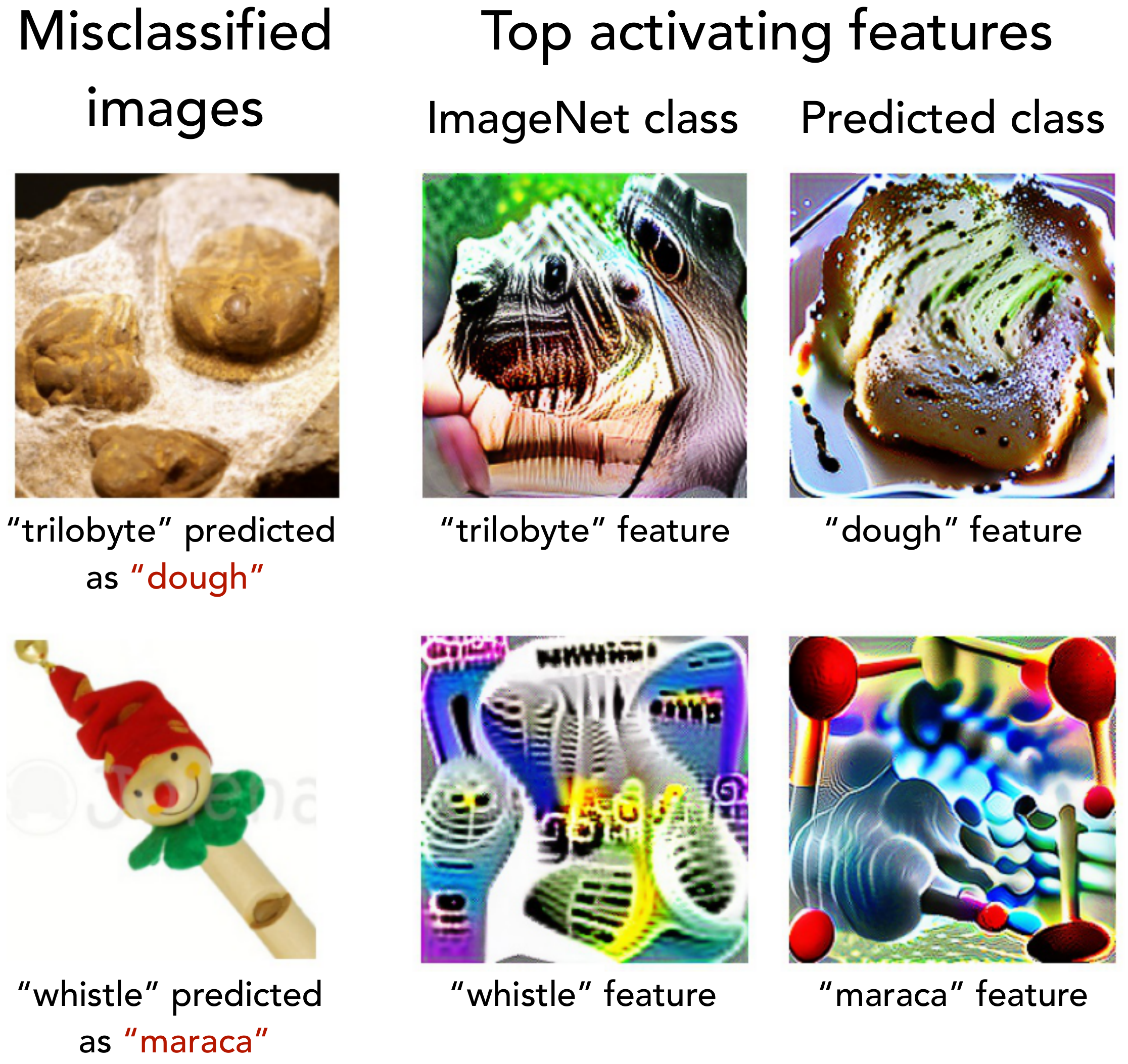}
	\caption{Examples of misclassified ImageNet images for which annotators 
	deem the top activated feature for the predicted class (\emph{rightmost}) 
	as a better match than the top activated feature 
	for the ground truth class (\emph{middle}).}
	\label{fig:misclassification}
\end{figure}
\fi

%% file: related.tex
\section{Related Work}
We now discuss prior work in interpretability and generalized 
linear models. Due 
to the large body of work in both fields, we limit the discussion to 
closely-related studies.

\paragraph{Interpretability tools.}
There have been extensive efforts towards post-hoc interpretability tools for 
deep networks. 
Feature attribution methods provide insight into model 
predictions for a specific input instance.
These include saliency 
maps~\cite{simonyan2013deep,smilkov2017smoothgrad,sundararajan2017axiomatic},
surrogate models to interpret local decision boundaries 
\cite{ribeiro2016should}, and finding influential 
\cite{koh2017understanding}, prototypical \cite{kim2016examples}, or 
counterfactual inputs \cite{goyal2019counterfactual}. 
However, as noted by various recent studies, these local attributions 
can be easy to fool~\cite{ghorbani2019interpretation,slack2020fooling} or 
may otherwise fail to capture global aspects of model 
behavior~\cite{sundararajan2017axiomatic,adebayo2018sanity,adebayo2020debugging,leavitt2020towards}.
Several methods have been proposed to interpret 
hidden units within vision networks, for example by generating feature 
visualizations~\cite{erhan2009visualizing,yosinski2015understanding, 
nguyen2016synthesizing,olah2017feature} or assigning 
semantic concepts to them~\cite{bau2017network,bau2020understanding}. 
Our work is complementary to these methods as we use them as 
primitives to probe sparse decision layers.
Another related line of work is that on concept-based explanations, which 
seeks to explain the behavior of deep networks in terms of high-level 
concepts~\cite{kim2018interpretability,ghorbani2019towards,yeh2020completeness}.
One of the drawbacks of these methods is that the detected 
concepts need not be causally linked to the model's 
predictions~\cite{goyal2019explaining}. 
In contrast, in our approach, the identified high-level concepts, i.e., the deep 
features used by the sparse decision layer, entirely determine the model's 
behavior.

Most similar is the recent work by \citep{wan2020nbdt}, which proposes 
fitting a decision tree on a deep feature representation. Network decisions 
are then explained in terms of semantic descriptions for nodes along the 
decision path. \citet{wan2020nbdt} rely on heuristics for fitting 
and labeling the decision tree, that require an existing domain-specific 
hierarchy (e.g., WordNet), causing it to be more involved and limited in its 
applicability than our approach.

\paragraph{Regularized GLMs and gradient methods.} 
Estimating GLMs with convex penalties has been studied extensively. 
Algorithms for efficiently computing regularization paths include least angle 
regression for LASSO~\citep{efron2004least} and path following 
algorithms~\citep{park2007l1} for 
$\ell_1$ regularized GLMs. 
The widely-used R package \texttt{glmnet} by 
\citet{friedman2010regularization}
provides an efficient coordinate descent-based solver for GLMs with 
elastic net regularization, and attains state-of-the-art solving times on 
CPU-based hardware. Unlike our approach, this library is best suited for 
problems with few  examples or features, and is not directly  
amenable to GPU acceleration. 
Our solver also builds off a long line of work in variance 
reduced 
proximal gradient 
methods~\citep{johnson2013accelerating,defazio2014saga,gazagnadou2019optimal},
 which have stronger theoretical convergence rates when compared to 
stochastic gradient descent.

%% file: conclusion.tex
\section{Conclusion}
We demonstrate how fitting sparse linear models over deep representations 
can result in more debuggable models, and provide a diverse set of scenarios 
showcasing the usage of this technique in practice. 
The simplicity of our approach allows it to be broadly applicable to any 
deep network with a final linear layer, and may find uses beyond the 
language and vision settings considered in this paper. 

Furthermore, we have created a number of human experiments for 
tasks such as testing model simulatiblity, detecting spurious correlations 
and validating misclassifications. Although 
primarily used in the context of 
evaluating the sparse decision layer, 
the design of these experiments may be of independent interest. 

Finally, we recognize that while deep networks are popular within 
machine learning and artifical intelligence settings, linear models 
continue to be widely used in other scientific fields. We hope that 
the development and release of our elastic net solver will find 
broader use in the scientific community for fitting large scale 
sparse linear models in contexts beyond deep learning.

%% file: acknowledgement.tex
\section*{Acknowledgements}

We thank Dimitris Tsipras for helpful discussions.

Work supported in part by the Google PhD Fellowship, Open Philanthropy, and NSF grants 
CCF-1553428 and CNS-1815221. 
This material is based upon work supported by the Defense Advanced 
Research Projects Agency (DARPA) under Contract No. HR001120C0015.
Research was sponsored by the United States Air Force Research Laboratory 
and the United States Air Force Artificial Intelligence Accelerator and was 
accomplished under Cooperative Agreement Number FA8750-19-2-1000. The 
views and conclusions contained in this document are those of the authors 
and should not be interpreted as representing the official policies, either 
expressed or implied, of the United States Air Force or the U.S. Government. 
The U.S. Government is authorized to reproduce and distribute reprints for 
Government purposes notwithstanding any copyright notation herein.

%% file: appendix_glm.tex
\appendix
\section{SAGA-based solver for generalized linear models}
\label{app:solver}
In this section, we describe in further detail our solver for learning regularized 
GLMs in relation to existing work. 
Note that many of the components underlying our solver have been 
separately studied in prior work. 
However, we are the first to effectively combine them in a way that allows for
GPU-accelerated fitting of GLMs at ImageNet-scale.
The key algorithmic primitives we leverage to this end are variance reduced 
optimization methods and path algorithms for GLMs.

Specifically, our solver uses a mini-batch derivative of the SAGA algorithm 
\citep{gazagnadou2019optimal}, which belongs to a 
class of a variance reduced proximal gradient methods. These approaches 
have several benefits: a) they are easily parallelizable via GPU, b) they enjoy 
faster convergence rates than stochastic gradient methods, and c) they 
require minimal tuning and can converge with a fixed learning rate. 

Algorithm \ref{alg:solver} provides a step-by-step description of our solver. 
Here,  the 
proximal operator for elastic net 
regularization 
is 
\begin{equation}
\textrm{Prox}_{\lambda_1, \lambda_2}(\beta) = \begin{cases}
\frac{\beta - \lambda_1}{1+\lambda_2} &\text{if } \beta > \lambda_1 \\
\frac{\beta + \lambda_1}{1+\lambda_2} &\text{if } \beta < \lambda_1 \\
0 &\text{otherwise}
\end{cases}
\end{equation}

\paragraph{Table for storing gradients}
Note that the SAGA algorithm requires saving the gradients of the model 
for each individual example. For ImageNet-sized problems, this requires a 
prohibitive amount of memory, as both the number of examples ($>$1 
million) and the size of the gradient (of the linear model) are large. 

It turns out that for linear models with $k$ outputs, it is actually possible to 
store 
all of the necessary gradient information for a single example in a vector of 
size 
$k$---as demonstrated by \citet{defazio2014saga}. 
The key idea behind this approach is that rather than storing the full gradient 
step $(x_i^T\beta + \beta_0 - y_i)x_i$, we can instead just store the scalar $a_i 
= (x_i^T\beta + \beta_0 - y_i)$ per output (i.e., a vector of length $k$ in the 
case of 
multiple outputs). Thus, for a 
dataset with $n$ examples, this reduces the memory requirements of the 
gradient table to $O(nk)$. For ImageNet, we find that the entire table easily 
fits within GPU memory limits. 

There is one caveat here: in order to use this memory trick, it is necessary to 
incorporate the $\ell_2$ regularization from the elastic net into the proximal 
operator. This is precisely why we use the proximal operator of the elastic 
net, rather than of the $\ell_1$ regularization. Unfortunately, this means that 
the smooth part of the objective (i.e. the part not used in the proximal 
operator) is no longer guaranteed to be strongly convex, and so the 
theoretical analysis of \citet{gazagnadou2019optimal} no longer strictly 
applies. Nonetheless, we find that these variance reduced methods can still 
provide strong practical convergence rates in this setting without requiring 
much tuning of batch sizes or learning rates. 

\begin{algorithm}[!t]
	\caption{GPU-accelerated solver for the elastic net for a step size 
		$\gamma$ 
		and regularization parameters $\lambda, \alpha$}
	\label{alg:solver}
	\begin{algorithmic}[1]
		\STATE Initialize table of scalars $a_i' = 0$ for $i \in [n]$
		\STATE Initialize average gradient of table $g_{avg}=0$ and 
		$g_{0avg}=0$
		\FOR{minibatch $B\subset [n]$}
		\FOR{$i \in B$}
		\STATE $a_i = x_i^T\beta + \beta_0 - y_i$
		\STATE $g_i = a_i \cdot x_i$ \textit{// calculate new gradient information}
		\STATE $g_i' = a_i' \cdot x_i$ \textit{// calculate stored gradient 
			information}
		\ENDFOR
		\STATE $g = \frac{1}{|B|}\sum_{i \in B} g_i$
		\STATE $g' = \frac{1}{|B|}\sum_{i \in B} g_i'$
		
		\STATE $g_0 = \frac{1}{|B|}\sum_{i \in B} a_i$
		\STATE $g_0' = \frac{1}{|B|}\sum_{i \in B} a_i'$
		
		\STATE $\beta = \beta - \gamma(g - g' + g_{avg})$
		\STATE $\beta_0 = \beta_0 - \gamma(g_0 - g_0' + g_{0avg})$
		\STATE $\beta = \textrm{Prox}_{\gamma\lambda\alpha, 
			\gamma\lambda(1-\alpha)}(\beta)$
		\FOR{$i\in B$}
		\STATE $a_i' = a_i$ \textit{// update table}
		\STATE $g_{avg} = g_{avg} + \frac{|B|}{n}(g - g')$ \textit{// update 
			average}
		\STATE $g_{0avg} = g_{0avg} + \frac{|B|}{n}(g_0 - g_0')$ 
		\ENDFOR
		\ENDFOR
	\end{algorithmic}
\end{algorithm}
\paragraph{Stopping criterion}
We implement two simple stopping criteria, which both take in a tolerance 
level $\epsilon_\text{tol}$. The first is a gradient-based 
stopping criteria, which terminates when: 
$$\sqrt{\|\beta^{i+1} - \beta^i\|_2^2 + \|\beta_0^{i+1} - \beta_0^{i}\|_2^2} \leq \epsilon_\text{tol}$$
Intuitively, this stops when the change in the estimated coefficients is small. Our 
second stopping criteria is more conservative and uses a longer search horizon, and 
stops when the training loss has not improved by more than $\epsilon_\text{tol}$ 
for more than $T$ epochs for some $T$, which we call the lookbehind stopping criteria. 

In practice, we find that the gradient-based stopping criteria with 
$\epsilon_\text{tol}=10^{-4}$ is sufficient for most cases (i.e. the solver has 
converged sufficiently such that the number of non-zero entries will no 
longer change). For significantly larger 
problems such as ImageNet, where individual batch sizes can have much larger variability in progressing the training objective, we find that the lookbehind stopping criteria is 
sufficient with $\epsilon_\text{tol}=10^{-4}$ and $T=5$. 

\paragraph{Relation of the solver to existing work} We now discuss 
how our solver borrows and differs from existing work. First, note that the 
original SAGA algorithm \citep{defazio2014saga} analyzes the regularized 
form but updates its gradient estimate with one sample at a time, which is 
not amenable to GPU parallelism. On the other hand, 
\citet{gazagnadou2019optimal} analyze a minibatch variant of SAGA but 
without regularization. In our solver, we use a straightforward adaptation of 
minibatch SAGA to its regularized equivalent by including a proximal step for 
the elastic net regularization after the gradient step. 

To compute the regularization paths, we closely follow the framework of 
\citet{friedman2010regularization}. Specifically, we compute solutions for a 
decreasing sequence of regularization, using the solution of the previous 
regularization as a warm start for the next. The maximum regularization value 
which fits only the bias term is calculated as the fixed point of the coordinate 
descent iteration as 
\begin{equation}
\lambda_{max} = \max_j \frac{1}{N\alpha} \left|\sum_{i=1}^n x_{ij}y_i\right|
\end{equation}
and scheduled down to $\lambda_{min} = \epsilon\lambda_{max}$ over a 
sequence of $K$ values on a log scale, as done by 
\citet{friedman2010regularization}. Typical suggested values are to take 
$K=100$ and $\epsilon=0.001$, which are what we use in all of our experiments. 
For extensions to logistic and multinomial 
regression, we refer the reader to \citet{friedman2010regularization}, and 
note that our approach is the same but substituting our SAGA-based solver 
in liue of the coordinate descent-based solver. 

\subsection{Timing Experiments}
\label{app:timing}
In this section, we discuss how the runtime of our solver scales with 
the problem size.
To be able to compare our solver with existing approaches, the experiments 
performed here are at a smaller scale than those in the main body of the 
paper.

\paragraph{Problem setting \& hyperparameters.}
The problem we examine is that of fitting a linear decision layer for the 
CIFAR-10 dataset using the deep feature representation of an 
ImageNet-trained ResNet-50 (2048-dimensional features). 
We then vary the number of training examples (from 1k to 50k) and fit an 
elastic net regularized GLM using various methods. We compare 
\texttt{glmnet} (state-of-the-art, coordinate 
descent-based solver) on a 9th generation Intel Core i7 with 6 cores clocked at 2.6Ghz, 
and our approach \texttt{glm-saga} using a GeForce GTX 1080ti. We note that in these 
small-scale experiments, the graphics card remains at around 10-20\% utilization, 
indicating that the problem size is too small to fully utilize the GPU. 

We fix $\alpha=0.99$, $\epsilon = 10^{-4}$, set aside 10\% of the training 
data for validation, and calculate regularization paths for $k=100$ different values, which are the defaults for \texttt{glmnet}. 
For our approach, we additionally use a mini-batch size 
of 512, a learning rate of 0.1, and a tolerance level of $10^{-4}$ for the gradient-based 
stopping criteria. 

\paragraph{Improvements in scalability}
As expected, on smaller problem instances with a couple thousand examples, 
\texttt{glmnet} is faster than our solver---cf. Table~\ref{tab:app_timing}. This 
is largely due to the increased 
base running time of our solver---a consequence of  
gradient based methods requiring some time to converge. 
However, as the 
problem 
size grows, the runtime of \texttt{glmnet} increases rapidly, and exceeds the running 
time of \texttt{glm-saga} at 3,000 datapoints. For example, it takes almost 
40 minutes to fit 4,000 data points with \texttt{glmnet}, an increase of 20x 
the running time for 4x the data relative to the running time for 1,000 data points. In contrast, our solver only needs 19 
minutes to fit 4,000 datapoints, an increase of 2x the running time for 4x 
the data.
Consequently, while \texttt{glmnet} takes a considerable amount of time to fit 
the full CIFAR10 problem size 
(50,000 datapoints)---nearly 13 hours---our solver can do the same in only 
33 minutes. Notably, our solver can fit the regularization paths 
of the decision layer for the full ImageNet dataset (1 million examples with 
2048 features) in approximately 6 hours.  

\begin{table}
\setcounter{table}{35}
\begin{center}
\caption{Runtime in minutes for \texttt{glmnet} and \texttt{glm-saga} for 
fitting a sparse decision layer on the CIFAR-10 dataset using deep 
representations (2048D) for a pre-trained ResNet-50. Here, we assess how 
the runtime of different solvers scales as a function of training data points.}
\label{tab:app_timing}
\begin{tabular}{lcccccc}
\toprule
& \multicolumn{6}{c}{Number of examples}\\
\cline{2-7}
Solver & 1k & 2k & 3k & 4k & 5k & 50k \\
\midrule
\texttt{glmnet} & 2 & 7 & 25 & 39 & 58 & 776\\
\texttt{glm-saga} & 9 & 13 & 17 & 19 & 22 & 33\\
\bottomrule
\end{tabular}
\end{center}
\setcounter{table}{11}
\end{table}


\paragraph{Backpropagation libraries} One more alternative to fitting 
linear models at scale is to use a standard autodifferentiation library such 
as PyTorch or Tensorflow. However, typical optimizers used in these libraries 
do not handle non-smooth regularizers well (i.e., the $\ell_1$ penalty of the elastic
net). In practice, these types of approaches must gradually schedule learning 
rates down to zero in order to converge, and take too long to compute 
regularization paths. For example, the fixed-feature transfer experiments from 
\citet{salman2020adversarially} takes approximately 4 hours to fit the same CIFAR10 
timing experiment for a single regularization value. In contrast, the 
SAGA-based optimizers enables a flexible range of learning rates that can converge 
rapidly without needing to tune or decay the learning rate over time. 

\subsection{Elastic net, $\ell_1$, and $\ell_2$ regularization}
The elastic net is known to combine the benefits of both $\ell_1$ and $\ell_2$ regularization for linear models. The $\ell_1$ regularization, often seen in the LASSO, primarily provides sparsity in the solution. The $\ell_2$ regularization, often seen as ridge regression, brings improved performance, a unique solution via strong convexity, and a grouping effect of similar neurons. Due to this last property of $\ell_2$ regularization, highly correlated features will become non-zero at the same time over the regularization path. The elastic net combines all of these strengths, and we refer the reader to \citet{tibshirani2017sparsity} for further discussion on the interaction between elastic net, $\ell_1$, and $\ell_2$. 

\subsection{Feature ordering}
\label{app:order}
In the main body of the paper, we utilized regularization paths obtained via 
elastic net to obtain a sparse decision layer over deep features.
We now discuss an additional use case of regularization paths---as a
means to assess relative (deep) feature importance within the decision layer 
of a standard deep network. 
Such an ordering could, for instance, provide an alternative criteria for 
feature selection in "feature-highlighting" 
explanations~\citep{barocas2020hidden}. 

The underlying mechanism that allows us to do this is the $\ell_1$ 
regularization in the elastic net, which imposes sparsity properties on the 
coefficients of the resulting linear model \citep{tibshirani1994regression}. 
Specifically, the coefficients for each feature 
become non-zero at discrete points in the regularization path, as  
$\lambda$ tends to zero.  
Informally, one can view features that are assigned non-zero coefficients 
earlier as being more useful from an accuracy standpoint, given the sparsity regularization.

\begin{figure}[!h]
	
\begin{subfigure}{0.3\textwidth}
	\centering
	\includegraphics[width=1\columnwidth]{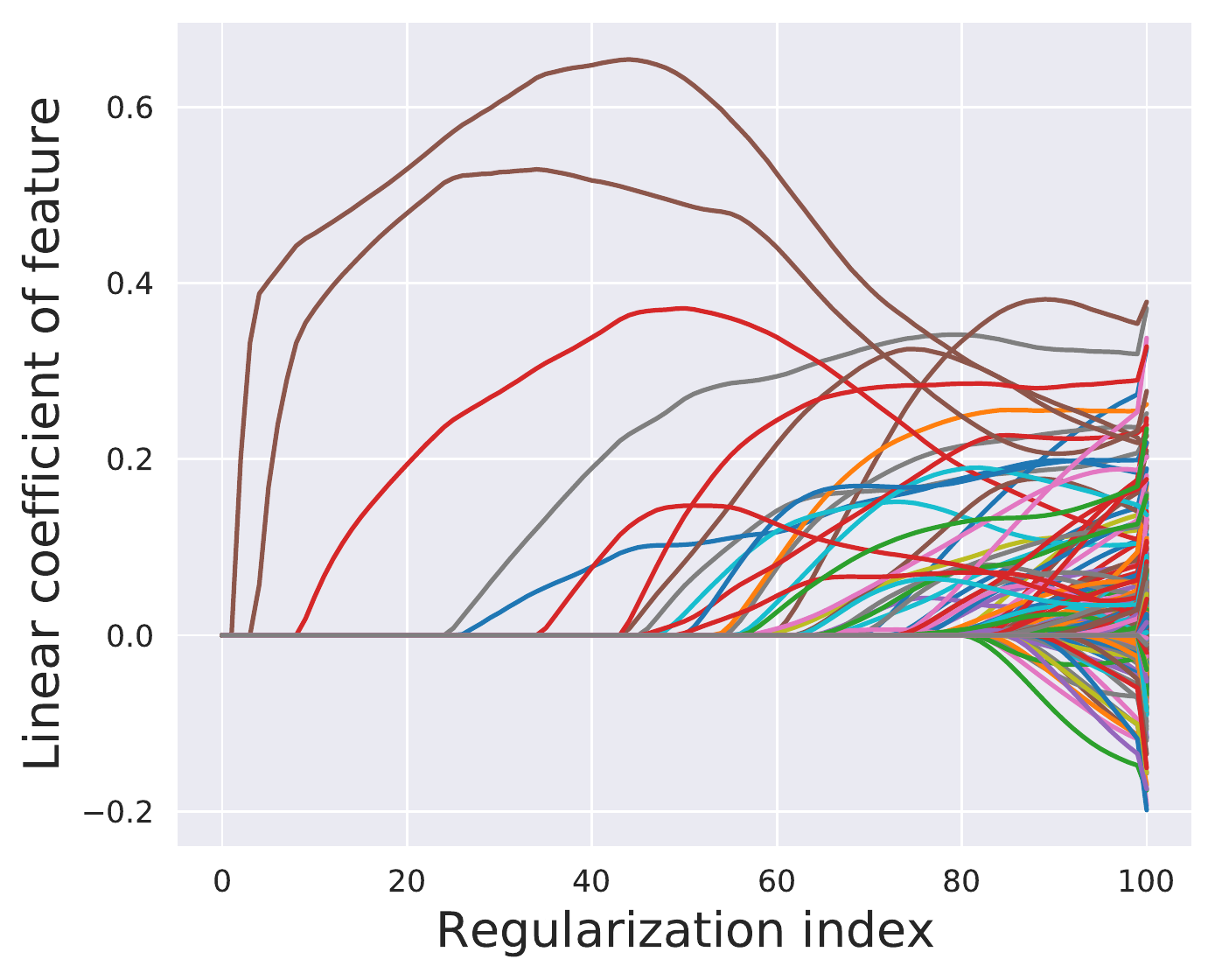}
\end{subfigure}
\hfill
\begin{subfigure}{0.64\textwidth}
	\centering
	\includegraphics[width=1\columnwidth]{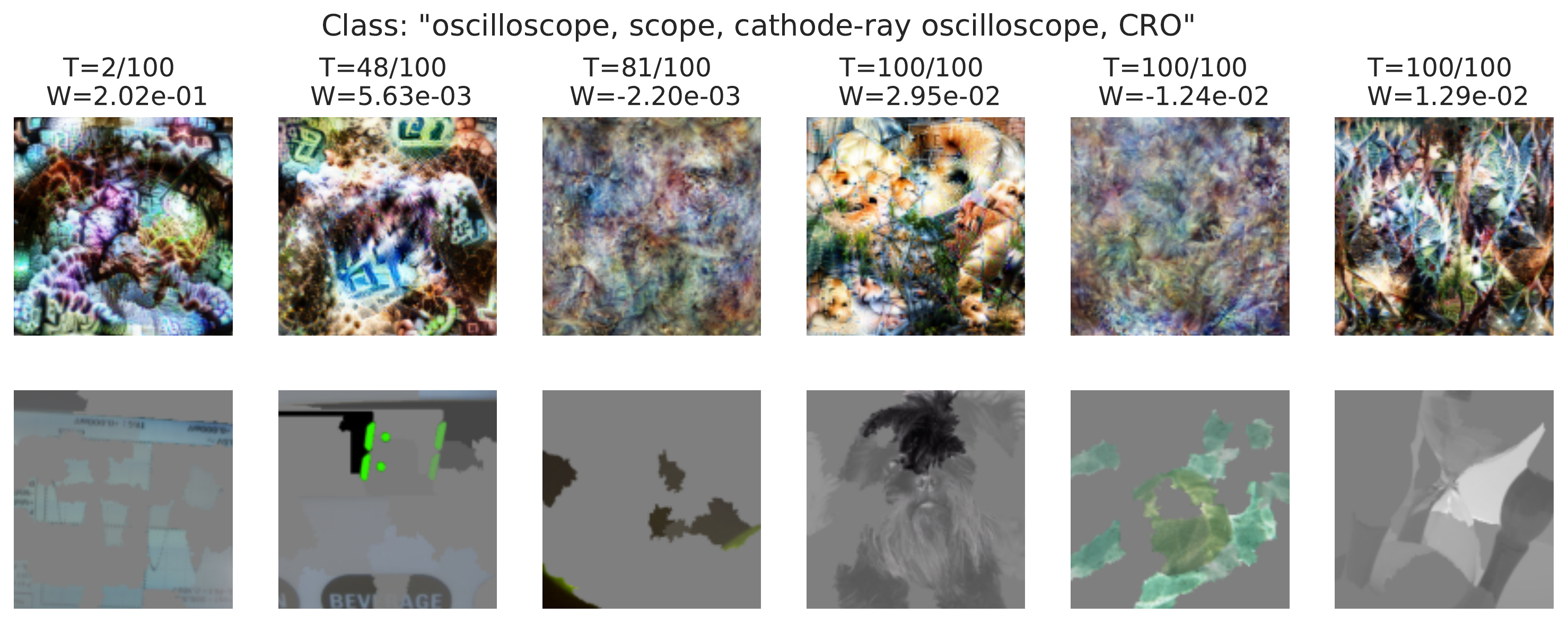}
\end{subfigure}
\begin{subfigure}{0.3\textwidth}
	\centering
	\includegraphics[width=1\columnwidth]{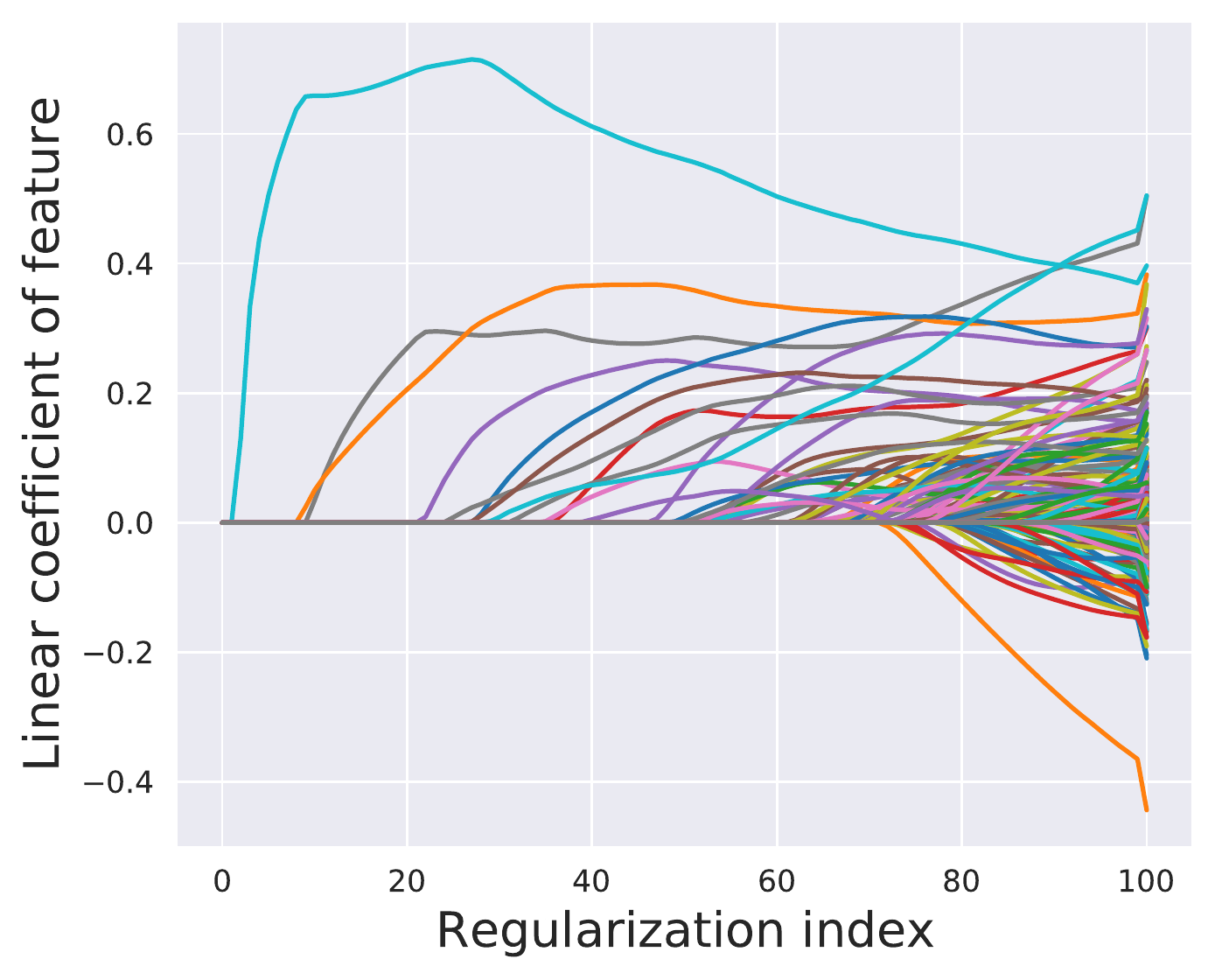}
\end{subfigure}
\hfill
\begin{subfigure}{0.64\textwidth}
	\centering
	\includegraphics[width=1\columnwidth]{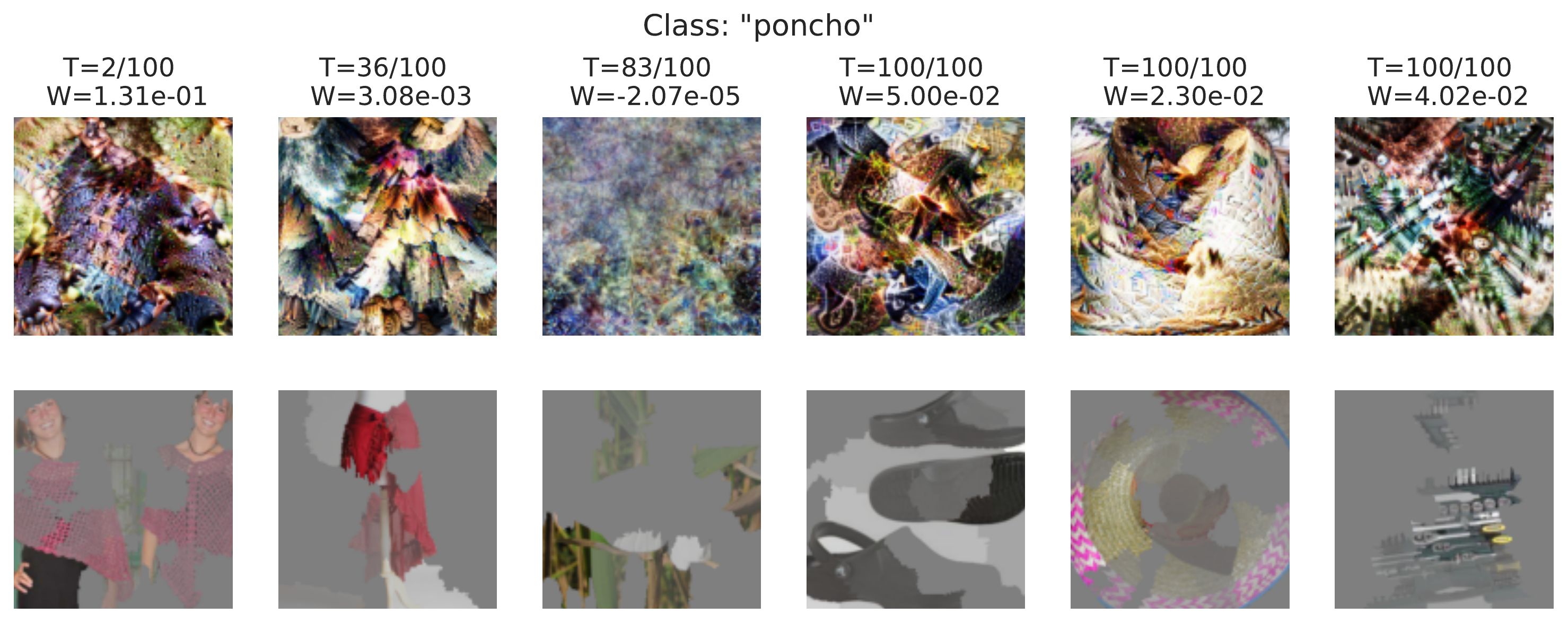}
\end{subfigure}
\caption{Sample regularization paths (\emph{left}) and feature ordering 
(\emph{right}) for sparse decision layers trained  on deep 
features of a ResNet-50 classifier for two ImageNet classes. Regularization 
paths highlight when different deep features are incorporated into the 
decision 
layer as the sparsity regularization is reduced. Sample features 
(as feature visualizations and LIME superpixels) included into the 
decision layer at increasing regularization indices (T) are shown on the right.}
	\label{fig:app_order_std_in}
\end{figure}

\begin{figure}[!h]
	
	\begin{subfigure}{0.3\textwidth}
		\centering
		\includegraphics[width=1\columnwidth]{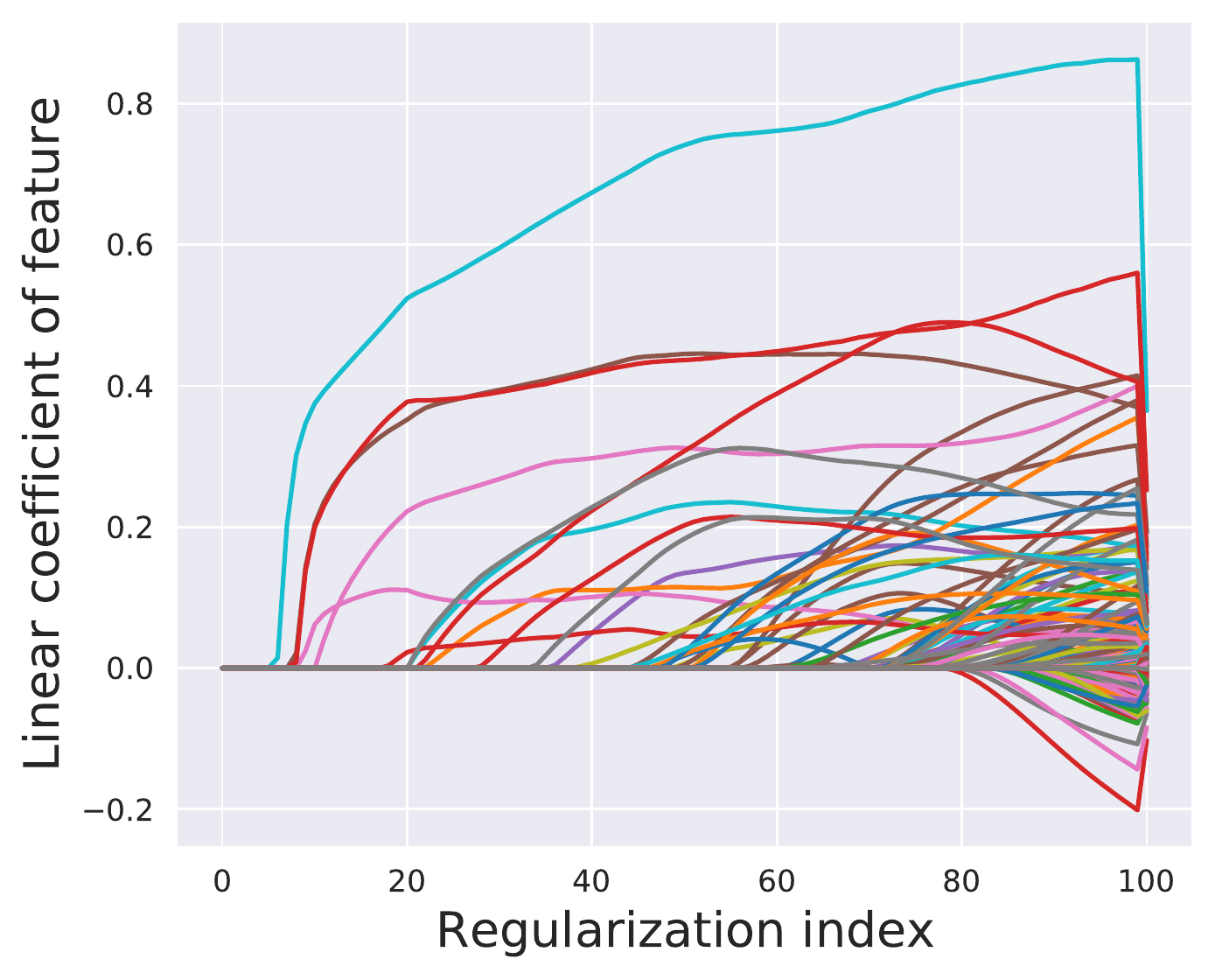}
	\end{subfigure}
	\hfill
	\begin{subfigure}{0.64\textwidth}
		\centering
		\includegraphics[width=1\columnwidth]{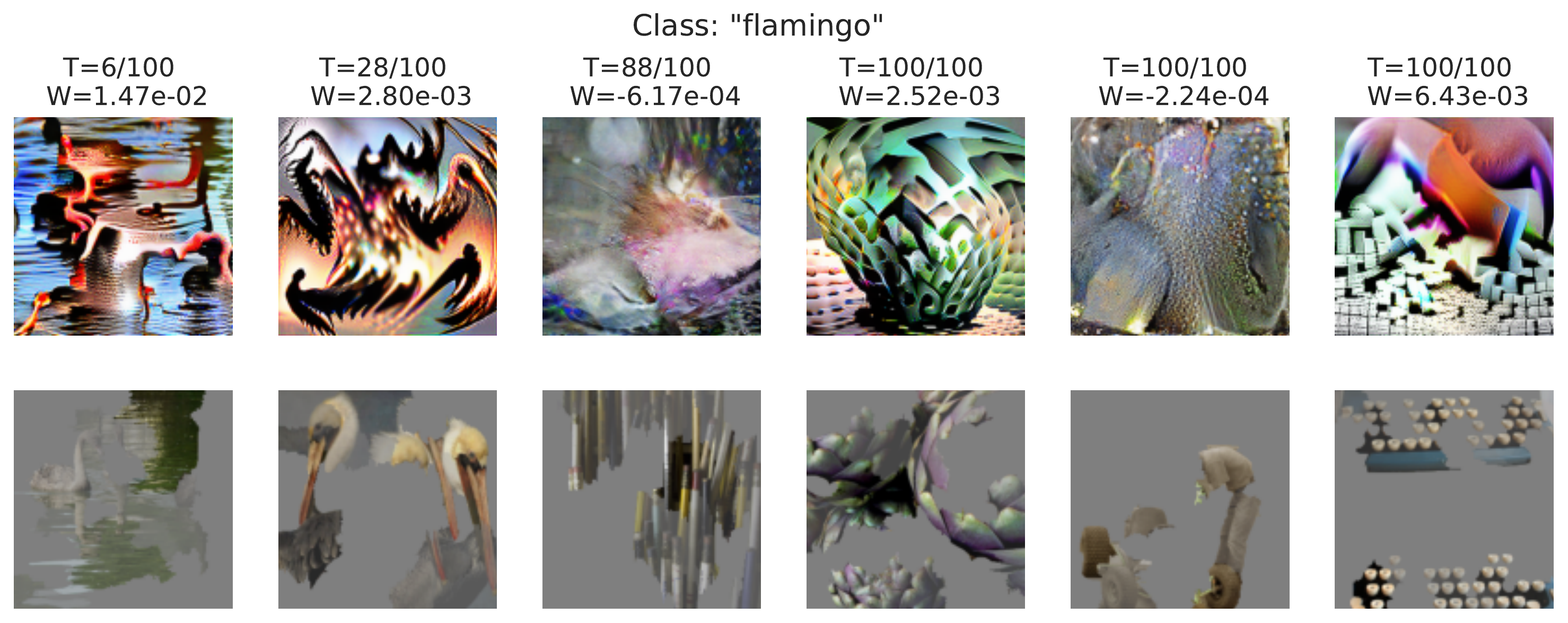}
	\end{subfigure}
	\begin{subfigure}{0.3\textwidth}
		\centering
		\includegraphics[width=1\columnwidth]{figures/glm/feature_ordering/reg_pathimagenet_3_dense_flamingo}
	\end{subfigure}
	\hfill
	\begin{subfigure}{0.64\textwidth}
		\centering
		\includegraphics[width=1\columnwidth]{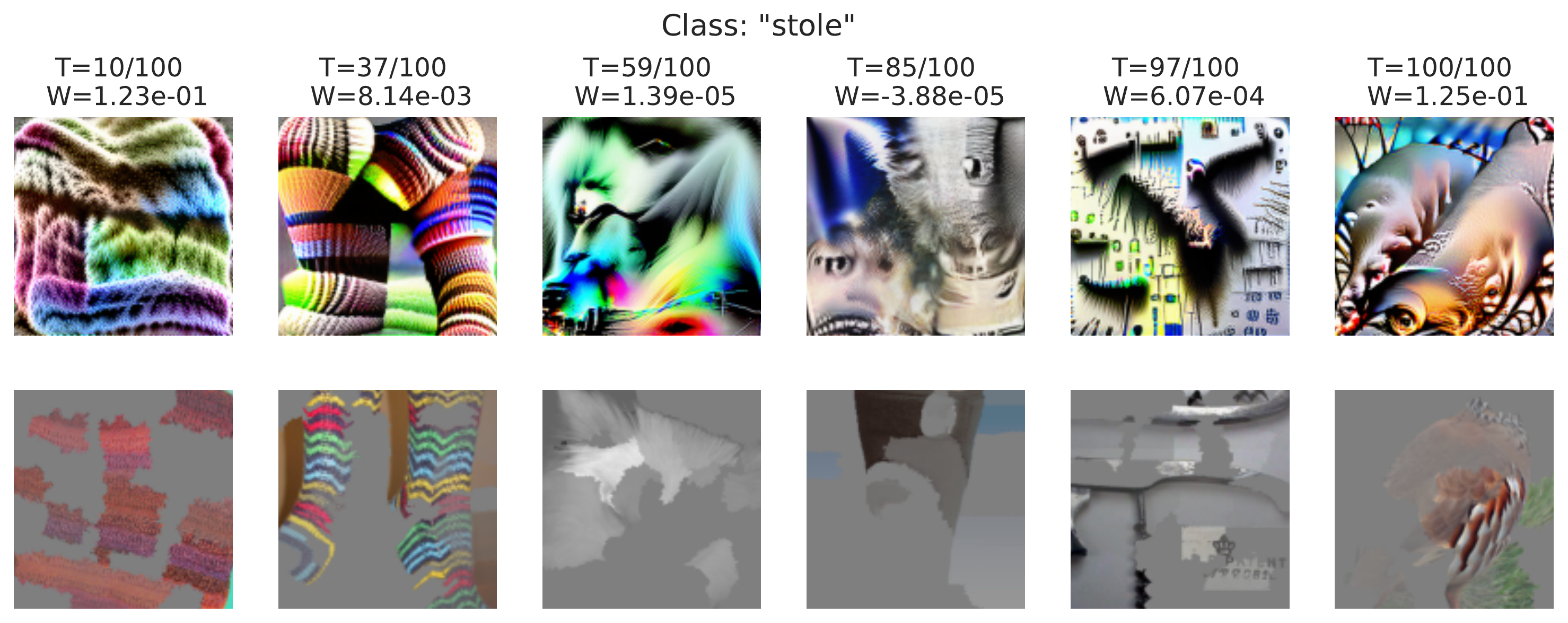}
	\end{subfigure}
	\caption{Sample regularization paths (\emph{left}) and feature ordering 
		(\emph{right}) for sparse decision layers trained  on deep 
		features of a robust ResNet-50 classifier for two ImageNet classes. 
		Regularization 
		paths highlight when different deep features are incorporated into the 
		decision 
		layer as the sparsity regularization is reduced. Sample features 
		(as feature visualizations and LIME superpixels) included into the 
		decision layer at increasing regularization indices (T) are shown on the 
		right.}
	\label{fig:app_order_rob_in}
\end{figure}

\begin{figure}[!h]
	
	\begin{subfigure}{0.3\textwidth}
		\centering
		\includegraphics[width=1\columnwidth]{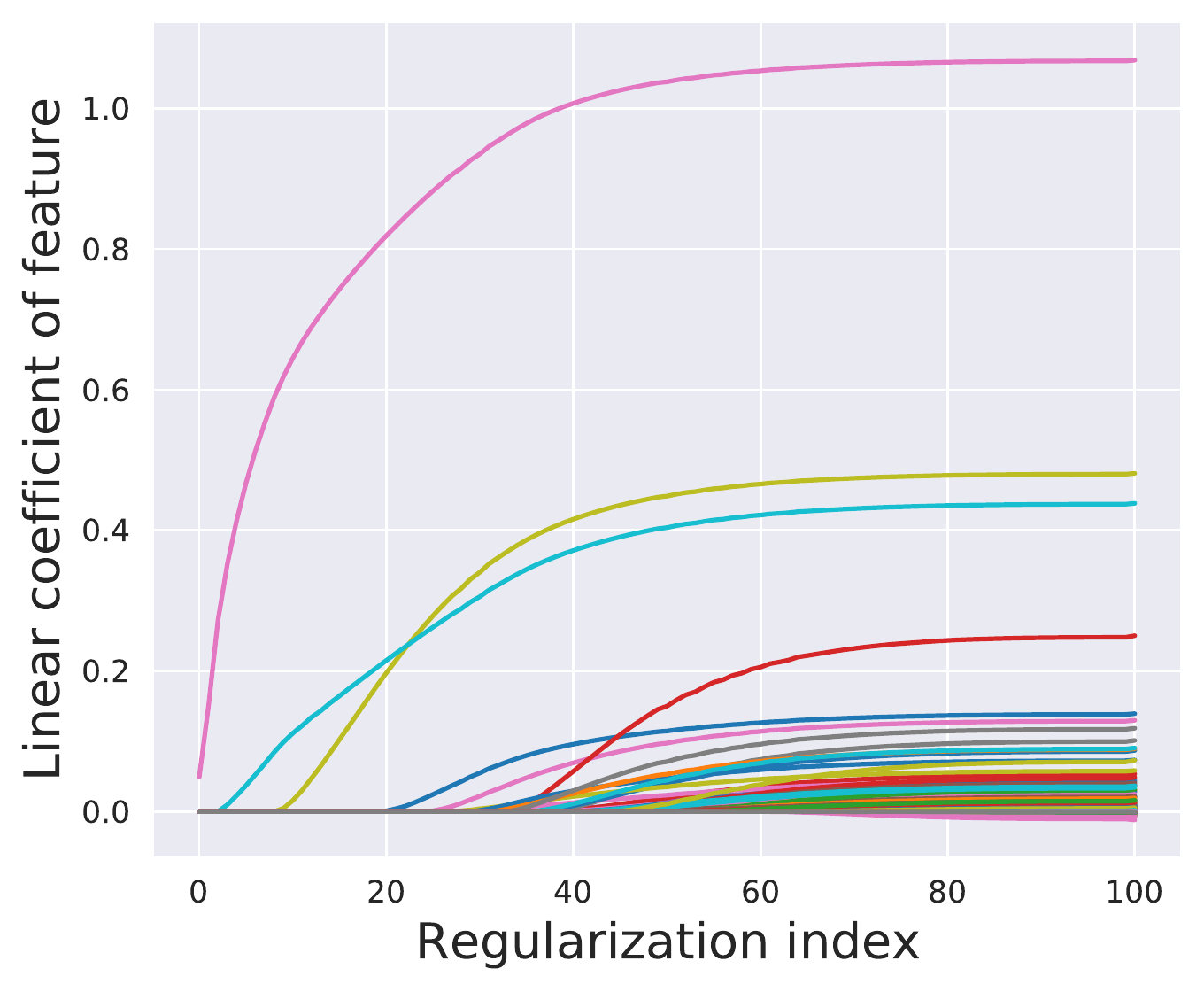}
	\end{subfigure}
	\hfill
	\begin{subfigure}{0.64\textwidth}
		\centering
		\includegraphics[width=1\columnwidth]{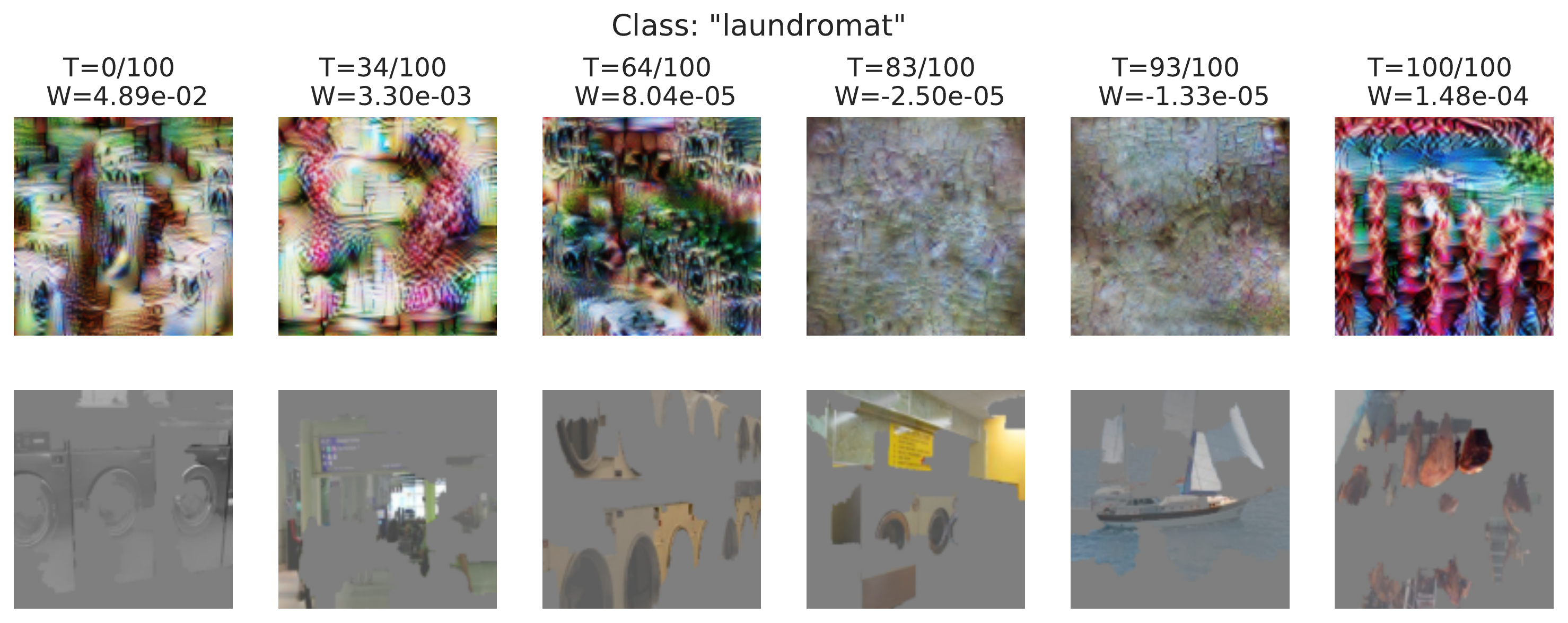}
	\end{subfigure}
	\begin{subfigure}{0.3\textwidth}
		\centering
		\includegraphics[width=1\columnwidth]{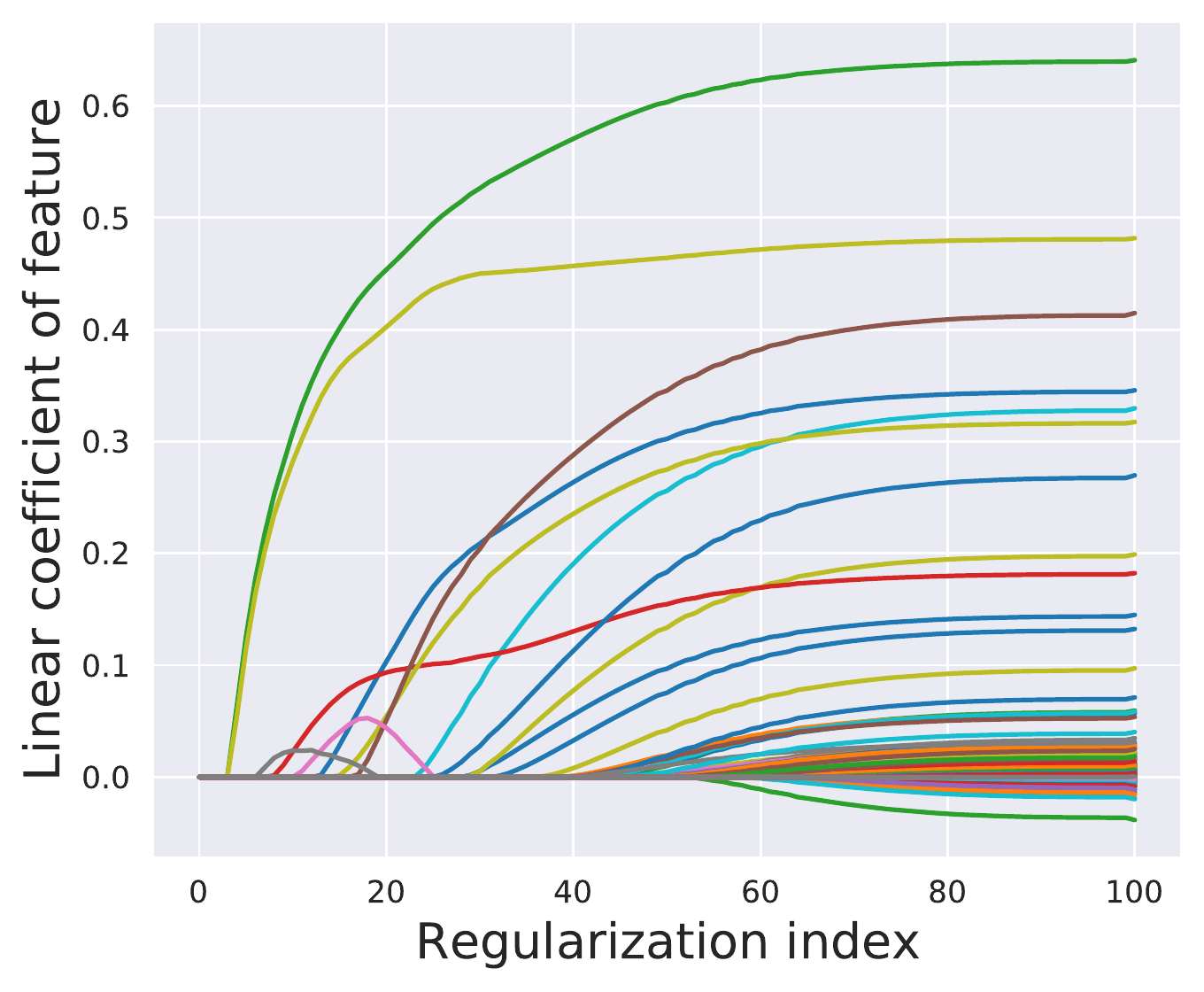}
	\end{subfigure}
	\hfill
	\begin{subfigure}{0.64\textwidth}
		\centering
		\includegraphics[width=1\columnwidth]{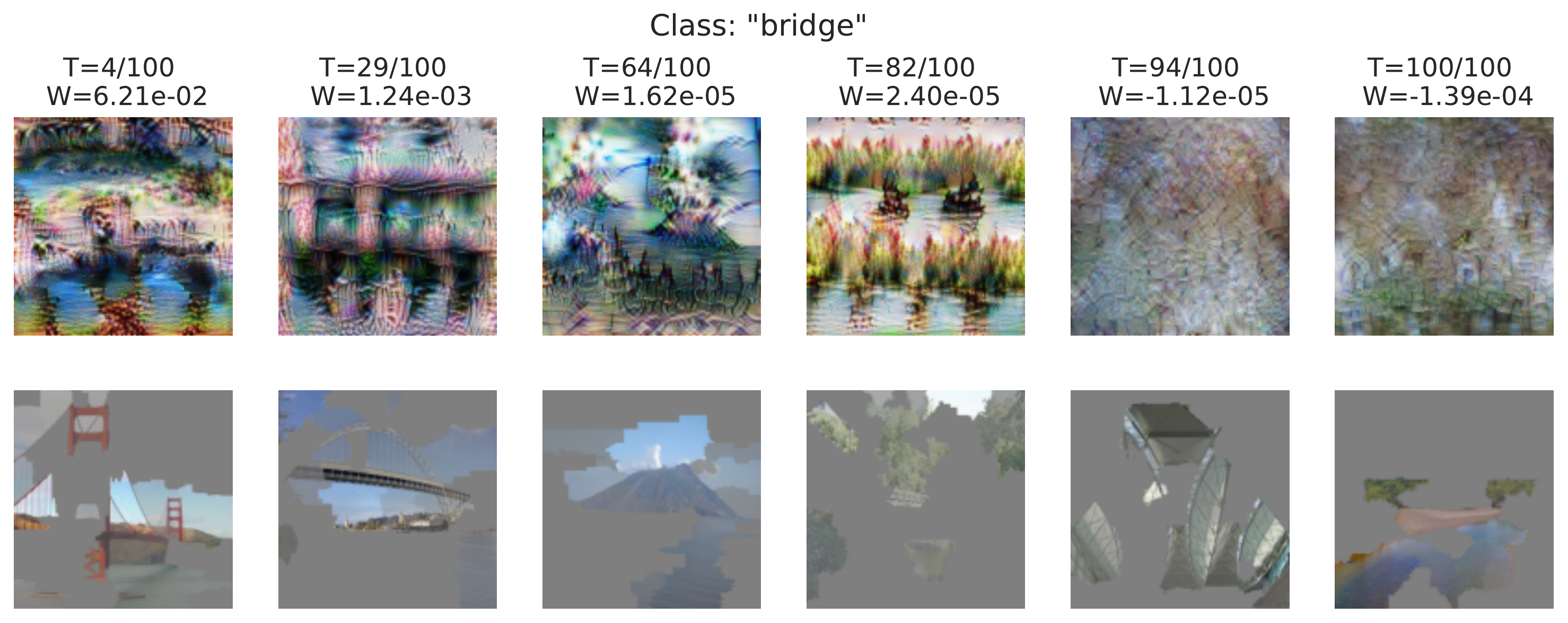}
	\end{subfigure}
	\caption{Sample regularization paths (\emph{left}) and feature ordering 
		(\emph{right}) for sparse decision layers trained  on deep 
		features of a ResNet-50 classifier for two Places-10 classes. 
		Regularization 
		paths highlight when different deep features are incorporated into the 
		decision 
		layer as the sparsity regularization is reduced. Sample features 
		(as feature visualizations and LIME superpixels) included into the 
		decision layer at increasing regularization indices (T) are shown on the 
		right.}
	\label{fig:app_order_std_places}
\end{figure}

\begin{figure}[!h]
	
	\begin{subfigure}{0.3\textwidth}
		\centering
		\includegraphics[width=1\columnwidth]{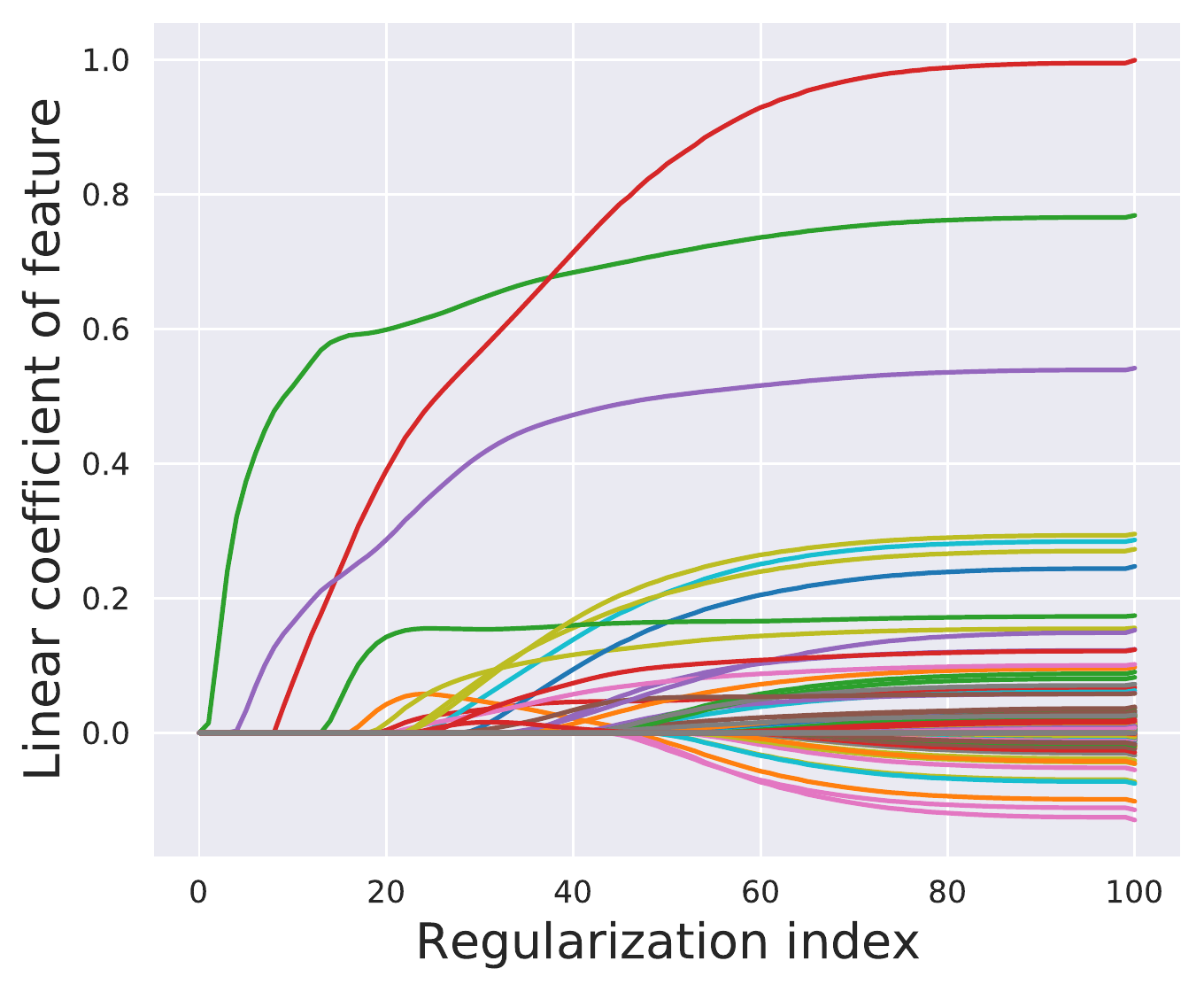}
	\end{subfigure}
	\hfill
	\begin{subfigure}{0.64\textwidth}
		\centering
		\includegraphics[width=1\columnwidth]{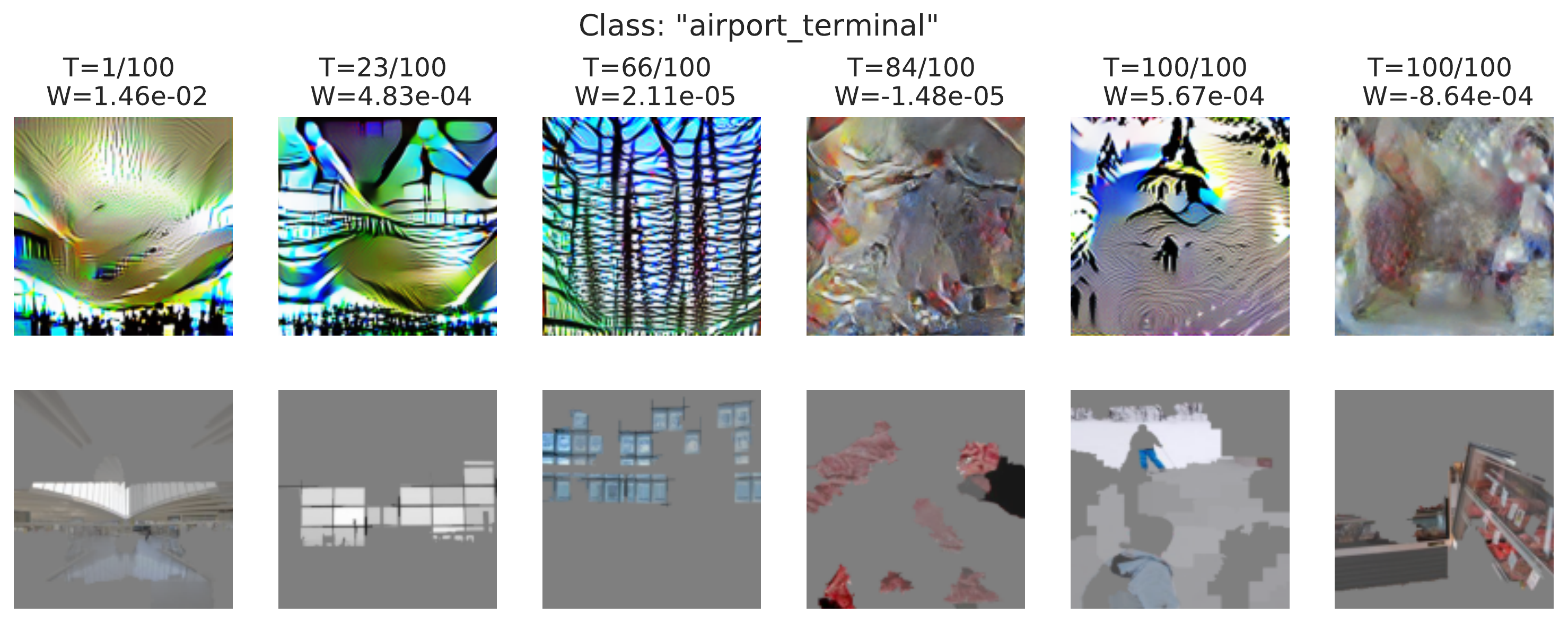}
	\end{subfigure}
	\begin{subfigure}{0.3\textwidth}
		\centering
		\includegraphics[width=1\columnwidth]{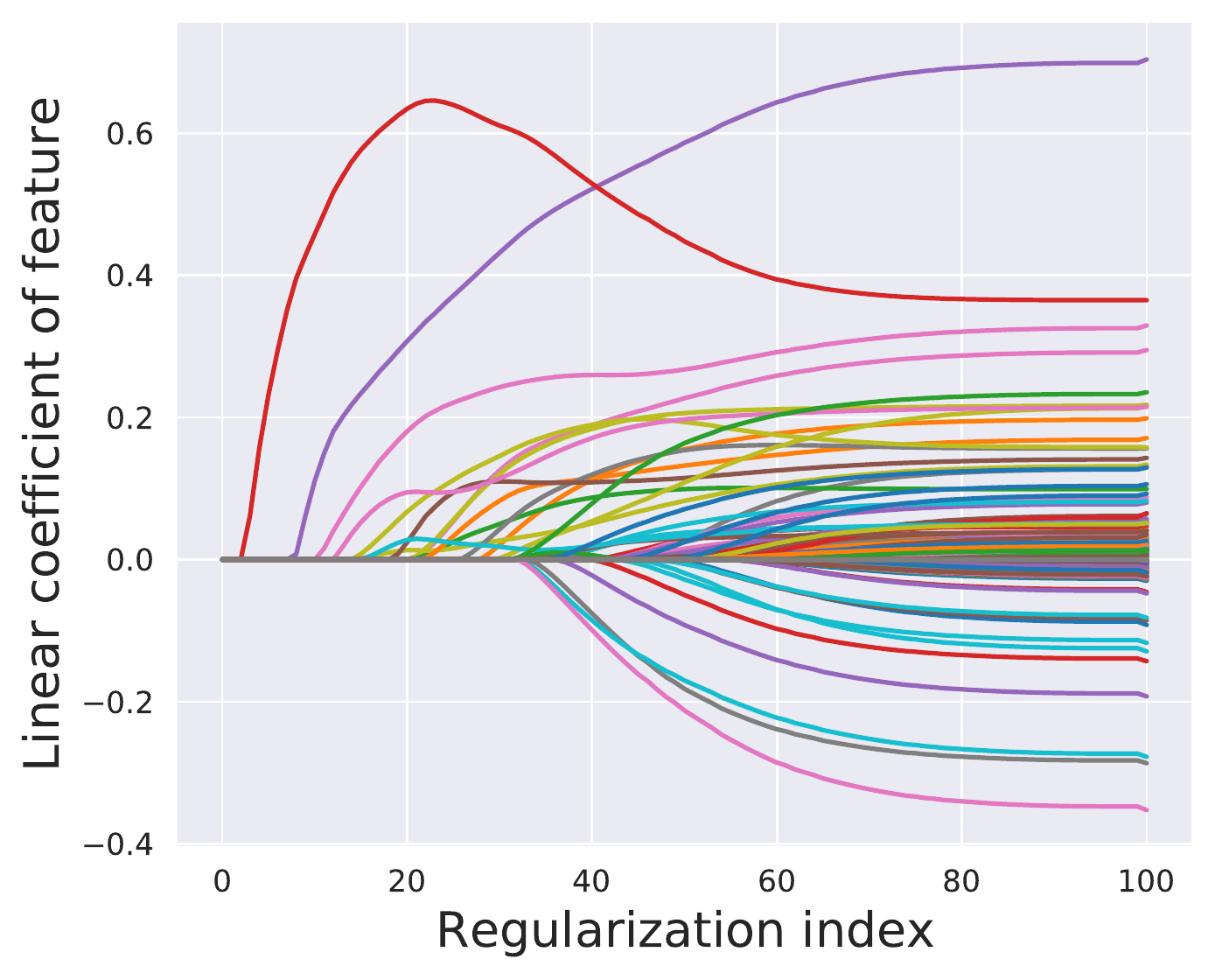}
	\end{subfigure}
	\hfill
	\begin{subfigure}{0.64\textwidth}
		\centering
		\includegraphics[width=1\columnwidth]{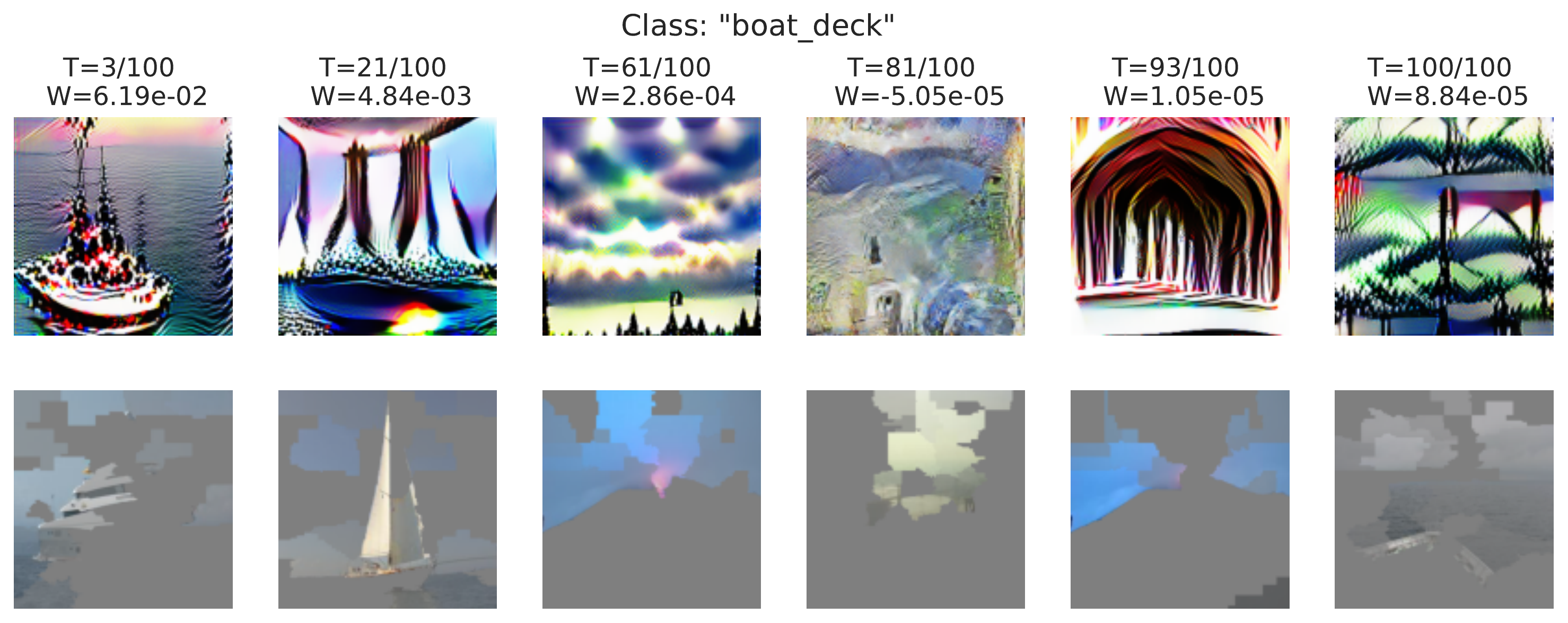}
	\end{subfigure}
	\caption{Sample regularization paths (\emph{left}) and feature ordering 
		(\emph{right}) for sparse decision layers trained  on deep 
		features of a robust ResNet-50 classifier for two Places-10 classes. 
		Regularization 
		paths highlight when different deep features are incorporated into the 
		decision 
		layer as the sparsity regularization is reduced. Sample features 
		(as feature visualizations and LIME superpixels) included into the 
		decision layer at increasing regularization indices (T) are shown on the 
		right.}
	\label{fig:app_order_rob_places}
\end{figure}

Consequently, the order in which (deep) features are incorporated into the 
sparse decision layers, within the regularization path, may shed light on their 
relative utility within the \emph{standard} deep network.
In Figures~\ref{fig:app_order_std_in}-~\ref{fig:app_order_rob_places}, we 
illustrate regularization paths along with the derived feature ordering for 
standard and robust ResNet-50 classifiers trained on ImageNet and 
Places-10 datasets. 
For all the models, it appears that features that are incorporated earlier into 
the regularization path (for a class) are actually more semantically aligned 
with the corresponding object category.

\clearpage

%% file: appendix_lime.tex
\section{Feature interpretations}
\label{app:feature_interpretation}

We now discuss in depth our procedure for generating feature 
interpretations for deep features in the vision and language settings.

\subsection{Feature visualization}

Feature visualization is a popular approach to interpret individual neurons 
within a deep network. Here, the objective is to synthesize inputs (via 
optimization in pixel space) that highly activate the neuron of interest.
Unfortunately, for standard networks trained via empirical risk minimization, 
it is well-known that vanilla feature visualization---using just gradient 
descent in input space---fails to produce semantically-meaningful 
interpretations.
In fact, these visualizations frequently suffer from artifacts and high 
frequency patterns~\cite{olah2017feature}.
One cause for this could be the reliance of standard models on 
input features that are imperceptible or unintuitive, as has been noted in 
recent studies~\cite{ilyas2019adversarial}.

To mitigate this challenge, there has been a long line of work on defining 
modified optimization objectives to produce more meaningful feature 
visualizations~\cite{olah2017feature}.
In this work, we use the Tensorflow-based Lucid 
library\footnote{\url{https://github.com/tensorflow/lucid}} to produce feature 
visualizations for standard models.
Therein, the optimization objective contains additional regularizations to 
penalize high-frequency changes in pixel space and to encourage 
transformation robustness.
Further, gradient descent is performed in the Fourier basis to further 
discourage high-frequency input patterns.
We defer the reader to~\citet{olah2017feature} for a more complete 
presentation.

In contrast, a different line of 
work~\cite{tsipras2019robustness,engstrom2019learning} has shown that 
robust (adversarially-trained) models tend to have better feature 
representations than their standard counterparts.
Thus, for robust models, gradient descent in pixel space is already 
sufficient to find semantically-meaningful feature visualizations.

\subsection{LIME}
\label{app:lime}

\paragraph{Image superpixels.}
Traditionally, LIME is used to obtain instance-specific explanations---i.e., to 
identify the superpixels in a given test image that are most responsible for 
the model's prediction.
However, in our setting, we would like to obtain a global understanding of 
deep features, independent of specific test examples.
Thus, we use the following two step-procedure to obtain LIME-based feature 
interpretations:
\begin{enumerate}
	\item Rank test set images based on how strongly they activate the feature 
	of interest. Then select the top-$k$ (or conversely bottom-$k$) images as 
	the most prototypical examples for positive (negative) activation of the 
	feature.
	\item Run LIME on each of these examples to identify relevant superpixels. 
	At a high level, this involves performing linear 
	regression to map image superpixels to the (normalized) activation of the 
	deep feature (rather than the probability of a specific class as is typical).
\end{enumerate}
Due to space constraints, we use $k=1$ in all our figures. However, in our 
analysis, we found the superpixels identified with $k=1$ to be representative 
of those obtained with higher values.

\paragraph{Word clouds for language models}
For language models, off-the-shelf neuron interpretability tools are 
somewhat more limited than their vision counterparts. 
Of the tools listed above, only LIME is used in the language domain to 
produce sentence-specific explanations.
Similar to our methodology for vision models, we apply LIME to 
a  given deep feature representation rather than the output neuron. However, 
rather than selecting prototypical images, we instead aggregate LIME explanations 
over the entire validation set. 

Specifically, for a given feature, we 
average the LIME weighting for each word over all of the sentences that the word appears in. This allows us to identify words that strongly activate/deactivate the given 
feature globally over the entire validation set,  which we then visualize using word clouds. In practice, since a word cloud has limited space, we provide the top 30 most highly weighted words to the word cloud generator. The exact procedure is shown in Algorithm \ref{alg:limeaggregate}, and we use the word cloud generator from \url{https://github.com/amueller/word_cloud}. 

\begin{algorithm}[!t]
	\caption{Word cloud feature visualization for language models for a 
		vocabulary $V$, 
		a corpus $w_{ij}$ for $i,j \in [m]\times[n]$ of $m$ sentences with $n$ 
		words}
	\label{alg:limeaggregate}
	\begin{algorithmic}[1]
		\FOR{$i = 1\dots m$}
		\STATE $\beta_i = \texttt{LIME}(w_i)$ \textit{// generate LIME explanation 
			for each sentence}
		\ENDFOR
		\FOR{$w \in V$}
		\STATE $K_w = \sum_{ij : w = w_{ij}} 1$ \textit{// count number of 
			occurances of word}
		\STATE $\hat \beta_w = \frac{1}{K_w}\sum_{ij : w = w_{ij}} \beta_{ij}$  
		\textit{// calculate average LIME explanation of word}
		\ENDFOR
		\STATE \Return $\texttt{Wordcloud}(\beta,V)$ \textit{// generate word 
			cloud for vocabulary $V$ weighted by $\beta$}
	\end{algorithmic}
\end{algorithm}



%% file: appendix_datasets.tex
\newpage
\section{Datasets and Models}
\label{app:datasets}

\subsection{Datasets}
We perform our experiments on the following widely-used vision and 
language datasets.

\begin{itemize}
	\item ImageNet-1k~\cite{deng2009imagenet,russakovsky2015imagenet}.
	\item Places-10: A subset of Places365~\cite{zhou2017places} containing 
	the classes ``airport terminal'', ``boat deck'', ``bridge'',
	 ``butcher's shop'', ``church-outdoor'', ``hotel room'', ``laundromat'',
	``river'', ``ski slope'' and ``volcano''.
	\item Stanford Sentiment Treebank (SST)~\cite{socher2013recursive} with 
	labels for ``positive'' and ``negative'' sentiment.
	\item Toxic Comments~\cite{wulczyn2017ex} with labels for ``toxic'', 
	``severe toxic'', ``obscene'', ``'threat'', ``insult'', and `identity hate''.
\end{itemize}

\paragraph{Balancing the comment classification task.} The toxic 
comments classification task has a highly unbalanced test set, and is largely 
skewed towards non-toxic comments. Consequently, the baseline accuracy for simply  
predicting the non-toxic label is often upwards of 90\% on the unbalanced test set. 
To get a more interpretable and 
usable performance metric, we instead randomly subsample the test set to be balanced with 
50\% each of toxic and non-toxic comments from the corresponding toxicity 
category. 
Thus, the baseline accuracy for random chance for 
toxic comment classification in our experiments  is 50\%. 

\subsection{Models}
We consider ResNet-50~\cite{he2016deep} classifiers 
and BERT~\cite{devlin2018bert} models for vision and language tasks 
respectively.
In the vision setting, we consider both standard  and 
robust models~\cite{madry2018towards}.

\paragraph{Vision.}  All the models are trained 
for 90 epochs, weight decay 1e-4 and momentum 
0.9.  We used a batch size of 512 for ImageNet and 128 for Places-10. The 
initial learning rate is 0.1 and is dropped by a factor of 10 every 30 
epochs.
The robust models were obtained using adversarial training with a 
$\ell_2$ 
PGD adversary~\cite{madry2018towards} with $\eps=3$, 3 attack steps and 
attack step size of $\frac{2 \times \eps}{3}$.

\paragraph{Language.}
The language models are all pretrained and available from the HuggingFace library, 
and use the standard BERT base architecture. 
Specifically, the sentiment classification model is from \url{https://huggingface.co/barissayil/bert-sentiment-analysis-sst} and the toxic comment models (both Toxic-BERT and Debiased-BERT) come from \url{https://huggingface.co/unitary/toxic-bert}. 


%% file: appendix_verification.tex
\section{Evaluating sparse decision layers}
\label{app:verification}

\subsection{Trade-offs for all datasets}
\label{app:tradeoffs}

\begin{figure}[!h]
	
	\centering

\begin{subfigure}{0.33\textwidth}
	\centering
	\includegraphics[width=1\columnwidth]{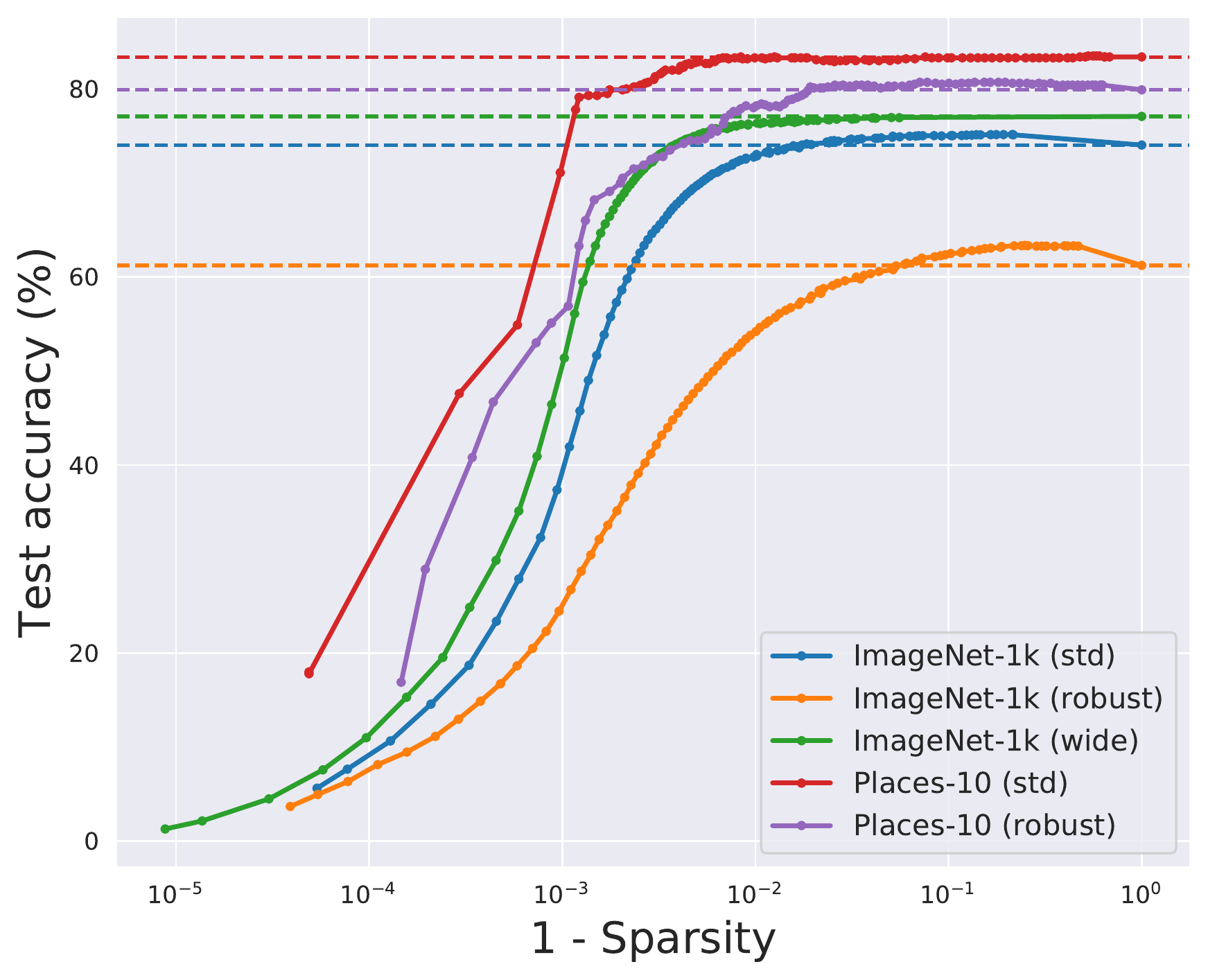}
	\caption{}
\end{subfigure}
\begin{subfigure}{0.33\textwidth}
	\centering
	\includegraphics[width=1\columnwidth]{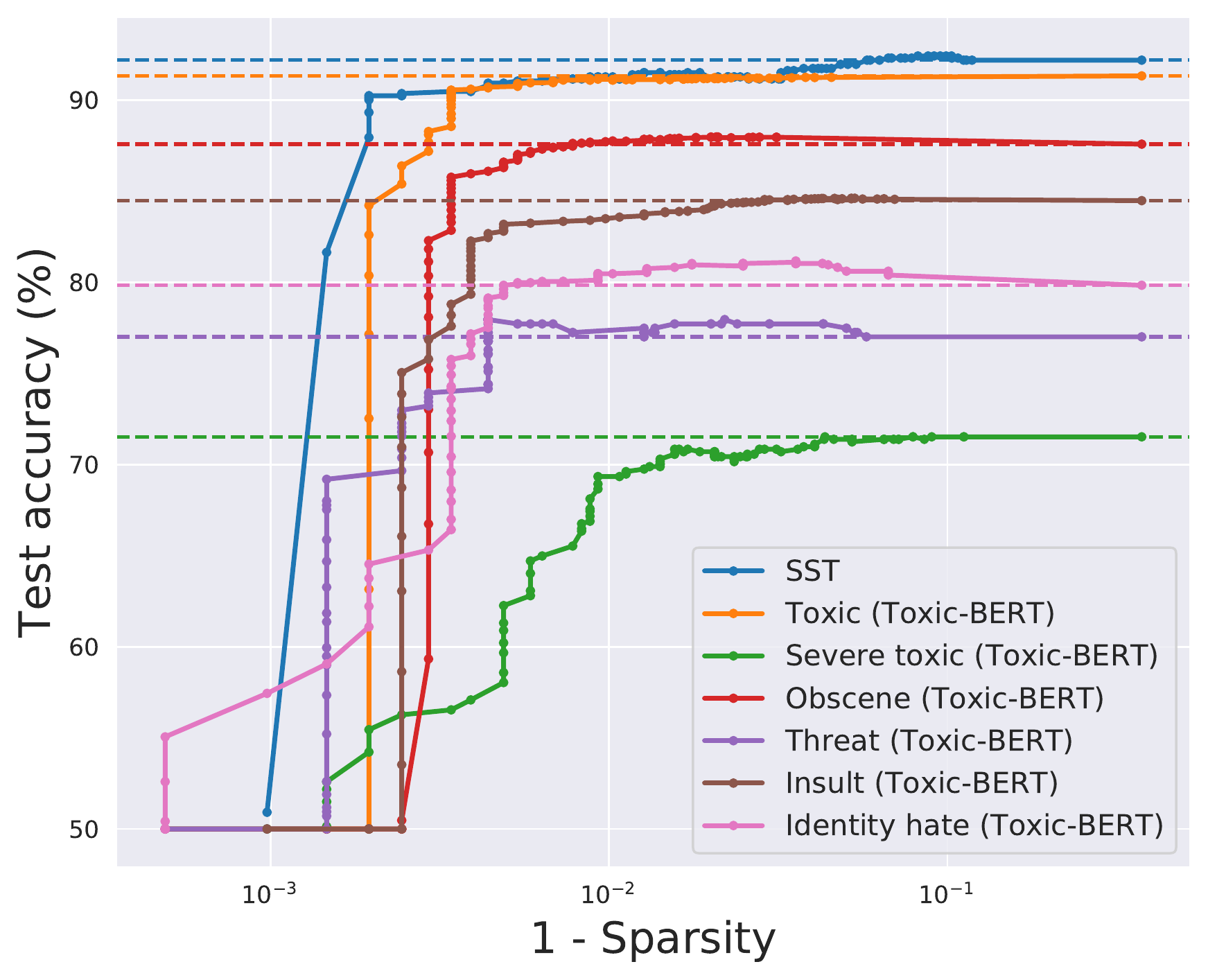}
	\caption{}
\end{subfigure}
\hfil
\begin{subfigure}{0.33\textwidth}
	\centering
	\includegraphics[width=1\columnwidth]{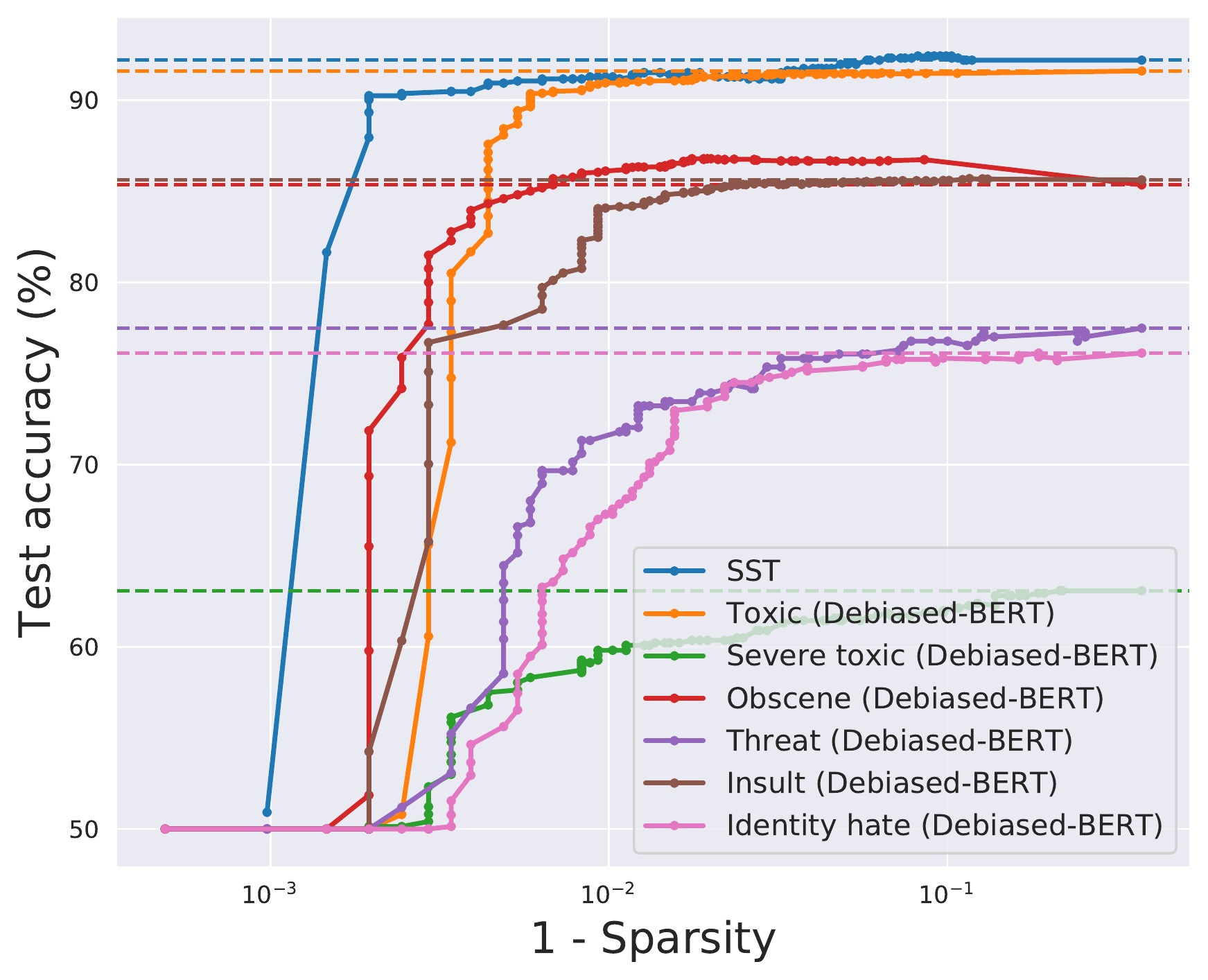}
	\caption{}
\end{subfigure}
\caption{Sparsity vs. accuracy trade-offs of models with sparse decision 
layers for (a) vision and (b,c) language tasks. }
	\label{fig:app_sparsity}
\end{figure}

In Figure~\ref{fig:app_sparsity}, we present an extended version of 
Figure~\ref{fig:sparsity}---including all the tasks and models we consider in 
both the vision and language setting. Each 
point on the curve corresponds to single 
	linear classifier from the regularization path in 
	Equation (\ref{eq:path}). Note that we include the (same) SST curve in 
	both 
	language plots for the Toxic and Debiased BERT models.

\subsubsection{Selecting a single sparse model}
\label{app:single}

As discussed in Section~\ref{sec:glm_explain}, the elastic net yields a 
sequence of linear models---with varying accuracy and sparsity---also 
known as the regularization path. In practice, performance of these models 
on a hold-out validation set can be used to guide model selection based on 
application-specific criteria. In our experiments, we set aside 10\% of the train 
set for this purpose. 

\paragraph{Our model selection thresholds.}
For both vision and NLP tasks, we use the validation set to identify the 
{sparsest} decision layer, whose accuracy is no more than 5\% lower 
on the validation set, compared to the best performing decision layer. 
As discussed in the paper, these thresholds are meant to be illustrative 
and can be varied depending on the specific application.
We now visualize the per-class distribution  of deep features for the sparse 
decision layers selected in Table~\ref{tab:ablation}. (We omit  the NLP tasks 
as they entail only two classes.)

\begin{figure}[!h]
    \centering
    \includegraphics[width=0.9\columnwidth]{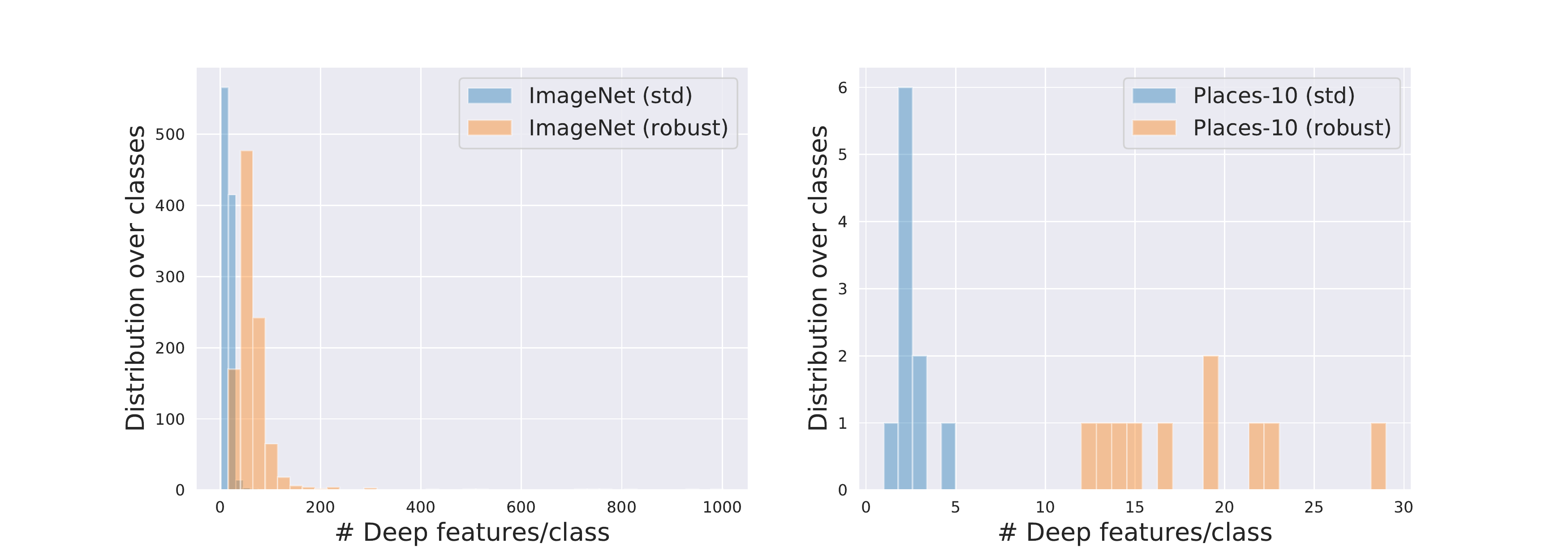}
    \caption{Distribution of the number of deep features used per class by 
    sparse decision 
    layers of vision models. Note that a standard (dense) decision layer 
    uses all 2048 deep features to predict every class.}
    \label{fig:app_per_class_sparsity}
\end{figure}

\subsection{Feature highlighting}
In Table \ref{tab:app_accuracy_extended} we show an extended version of 
Table \ref{tab:ablation}, which now contains an additional wide ImageNet 
representation as well as 3 additional toxic comment categories for each 
toxic comment classifier. The overall test accuracy of a subset of these 
models (before sparsification) is under `Dense $\rightarrow$ All' in 
Figure~\ref{tab:ablation}.

\begin{table}[!t]
	\setcounter{table}{36}
	\begin{center}
		\caption{Extended version of Table~\ref{tab:ablation}: Comparison of 
		the accuracy of dense/sparse decision layers 
			when 
			they are constrained to utilize only the top-$k$ deep features 
			(based on 
			weight magnitude). We also show overall model accuracy, and the 
			accuracy gained by using the remaining deep features.}
		\label{tab:app_accuracy_extended}
		\begin{tabular}{lccccccc}
			\toprule
			\multirow{2}{*}{Dataset/Model} & &\multicolumn{3}{c}{Dense} & 
			\multicolumn{3}{c}{Sparse} \\
			\cmidrule(lr){3-5}\cmidrule(lr){6-8} 
			& $k$ & All & Top-$k$ & Rest & All & Top-$k$ & Rest \\
			\midrule
			ImageNet (std) & \multirow{3}{*}{10} &   74.03 &  58.46 &  55.22 &   
			72.24 &  69.78 
			&  10.84 
			\\
			ImageNet (wide, std) & &   77.07 &  72.42 &  48.75 &   73.48 &  
			73.45 &   0.91 \\
			ImageNet (robust) & &   61.23 &  28.99 &  34.65 &   59.99 &  
			45.82 &  
			19.83 \\
			\midrule
			Places-10 (std) & \multirow{2}{*}{10} &  83.30 &  83.60 &  81.20 &   
			77.40 &  77.40 &  
			10.00 
			\\
			Places-10 (robust) &  &  80.20 &  76.10 &  76.40 &   77.80 &  
			76.60 &  
			40.20 \\
			\midrule
			SST              & 5 &   91.51 &  53.21 &  91.17 &   90.71 &  90.48 &  
			50.92 \\
			\midrule
			Toxic-BERT (toxic)  & \multirow{6}{*}{5} & 83.33 &  55.35 &  57.87 
			&   82.47 &  82.33 &  50.00 \\
			Toxic-BERT (severe toxic)  &&   71.53 &  50.00 &  50.14 &   67.57 &  
			50.00 &  50.00 \\
			Toxic-BERT (obscene) && 80.41 &  50.03 &  50.00 &   77.32 &  
			72.39 &  50.00 \\ %
			Toxic-BERT (threat)  &&   77.01 &  50.00 &  50.00 &   76.30 &  74.17 
			&  50.00 \\
			Toxic-BERT (insult) && 72.72 &  50.00 &  50.00 &   77.14 &  75.80 & 
			50.00 \\
			Toxic-BERT (identity hate) &&   79.85 &  57.87 &  50.00 &   74.93 &  
			71.49 &  50.00 \\
			\midrule
			Debiased-BERT (toxic) & \multirow{6}{*}{5} &   91.61 &  50.00 &  
			83.26 &   87.59 &  78.58 &  50.00 \\
			Debiased-BERT (severe toxic)  &&   63.08 &  50.00 &  50.00 &   
			55.86 &  53.81 &  50.00 \\
			Debiased-BERT (obscene) &&   85.36 &  50.00 &  58.36 &   81.50 &  
			81.17 &  50.00 \\
			Debiased-BERT (threat)  &&   77.49 &  50.00 &  50.00 &   68.96 &  
			50.00 &  50.00 \\
			Debiased-BERT (insult)  &&   85.63 &  50.00 &  59.95 &   79.28 &  
			71.48 &  50.00 \\
			Debiased-BERT (identity hate) &&   76.12 &  50.00 &  50.84 &   71.98 
			&  50.00 &  50.00 \\
			\bottomrule
		\end{tabular}
	\end{center}
	\setcounter{table}{15}
\end{table}

\input{appendix_visualization}

\clearpage
\subsection{Human evaluation}
\label{app:mturk_sim}

\begin{figure}[!t]
	\centering
	
	\includegraphics[width=0.7\columnwidth]{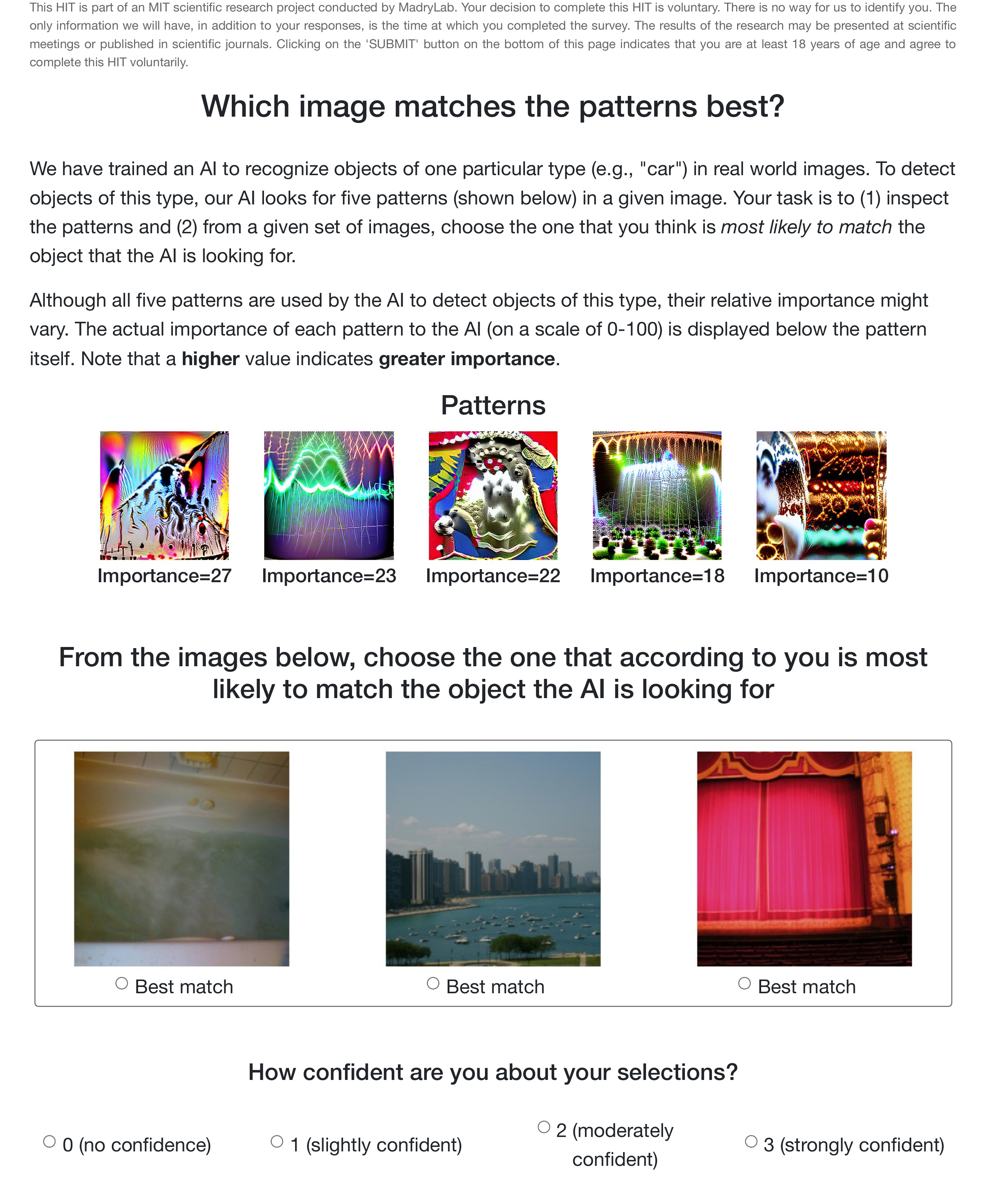}
	\caption{Sample MTurk task to assess how amenable models with 
		dense/sparse decision layers are to human understanding. }
	\label{fig:app_task_sim}
\end{figure}

We now detail the setup of our MTurk study from Section~\ref{sec:human}.  
For our analysis, we use a ResNet-50 that has been adversarially-trained 
($\eps=3$) on the ImageNet dataset. 
To obtain a sparse decision layer, we then train a sequence of GLMs via 
elastic net (cf. Section~\ref{sec:glm_explain}) on the deep representation of 
this network. Based on a validation set, we choose a single sparse 
decision layer---with 57.65\% test accuracy and 39.18 deep 
features/class on average. 

\begin{figure}[!t]
	\begin{subfigure}[b]{1\textwidth}
		\centering
		\includegraphics[width=0.9\columnwidth]{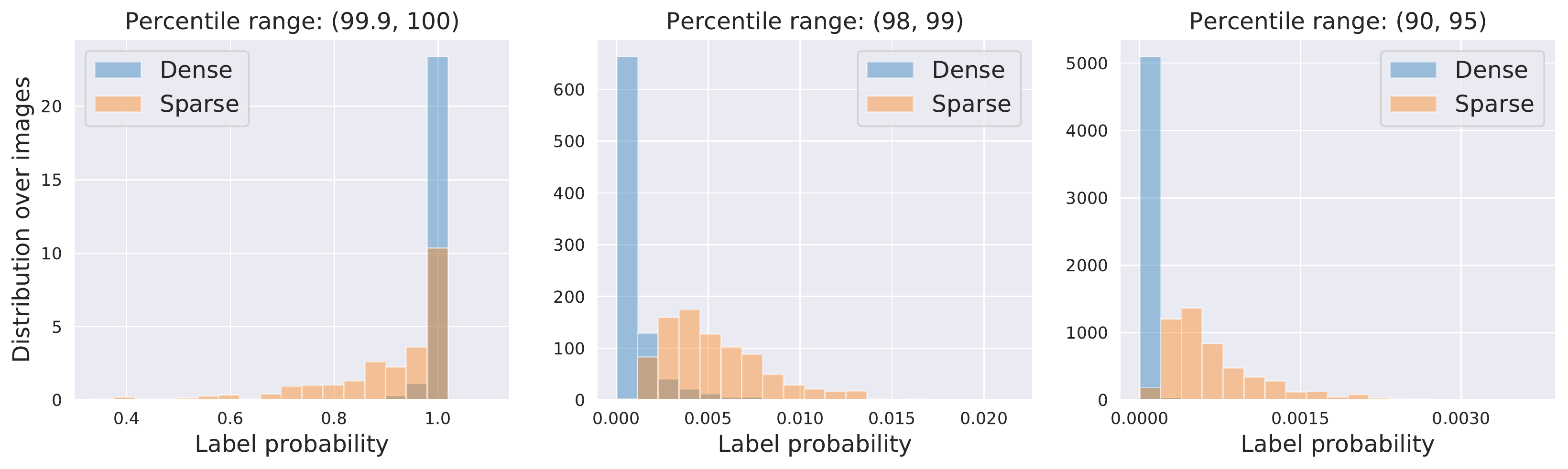}
		\caption{}
		\label{fig:logits}
	\end{subfigure}
	\begin{subfigure}[b]{0.48\textwidth}
		\centering
		\includegraphics[width=0.8\columnwidth]{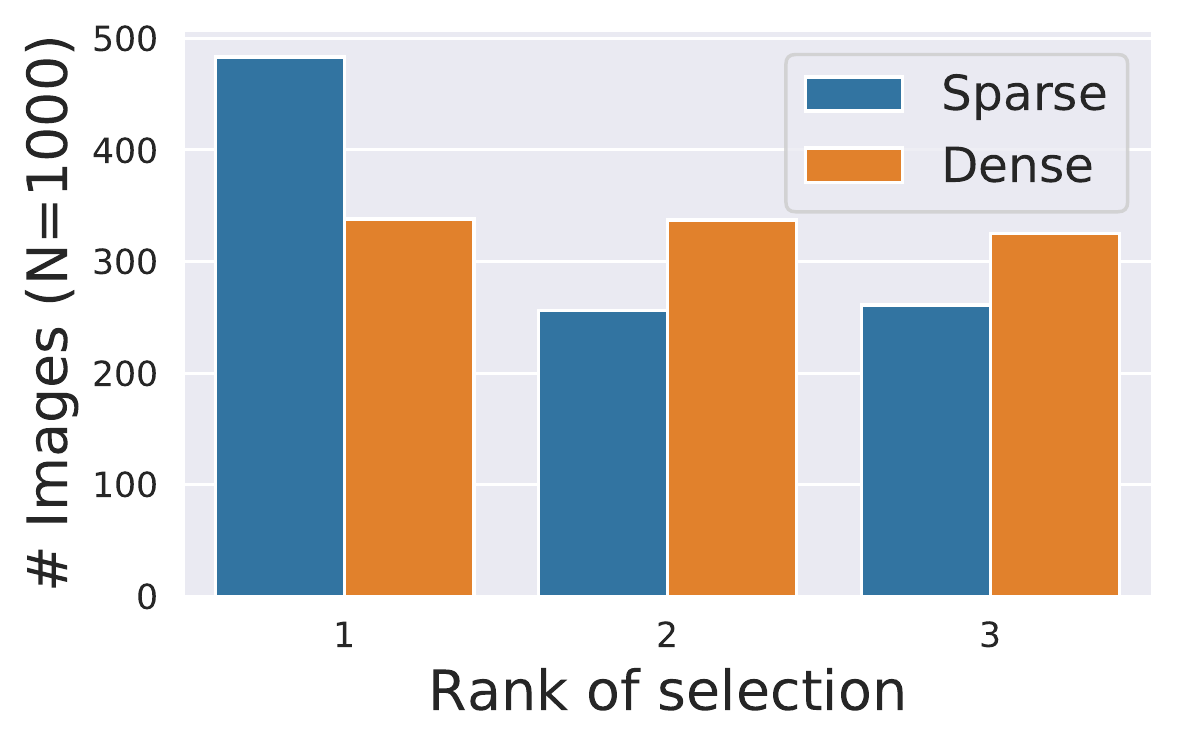}
		\caption{}
		\label{fig:ranking_comparison}
	\end{subfigure}
	\begin{subfigure}[b]{0.48\textwidth}
		\centering
		\includegraphics[width=0.7\columnwidth]{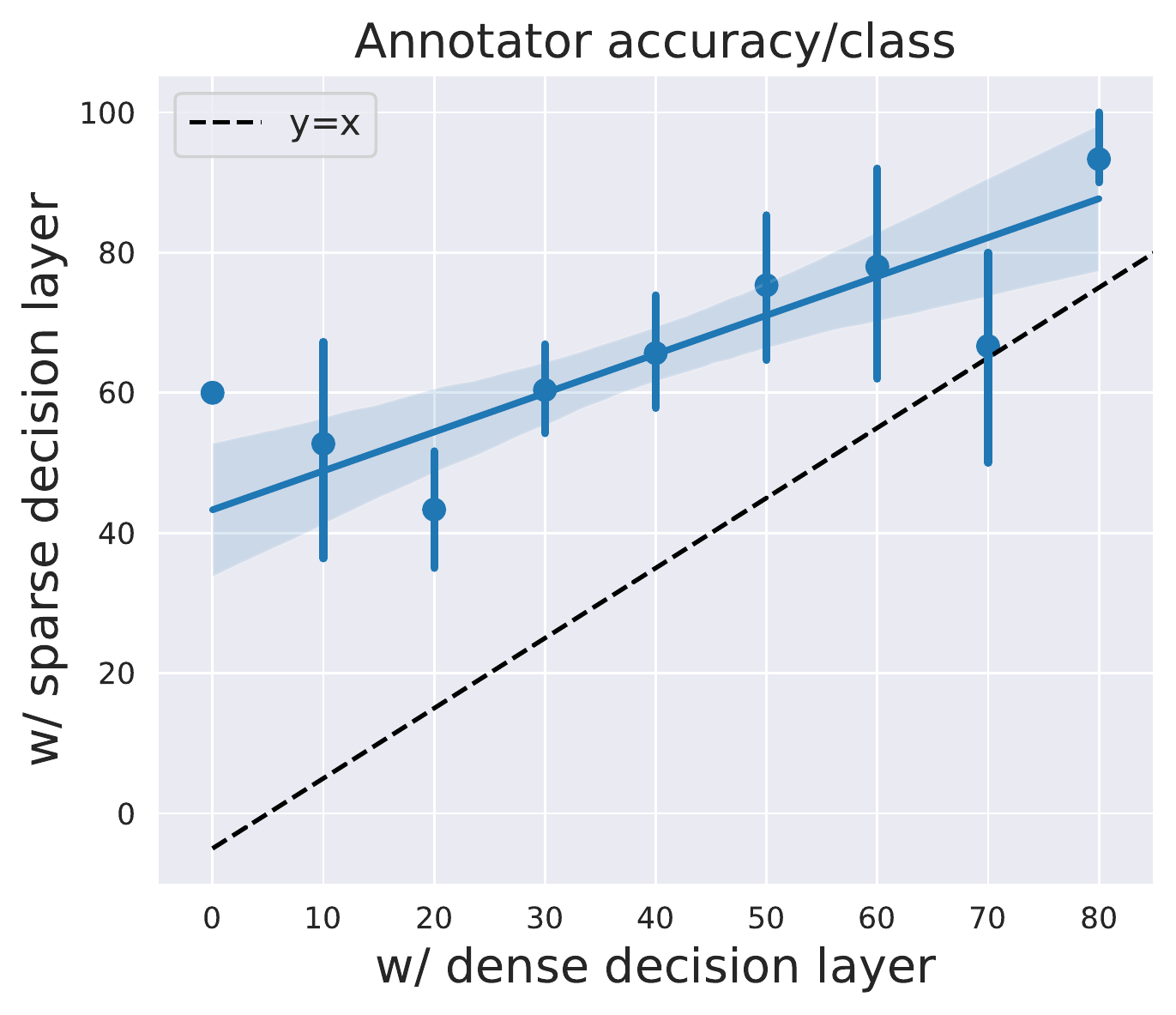}
		\caption{}
		\label{fig:relative_accuracy}
	\end{subfigure}
	\caption{(a) Distribution of the target label probability assigned by the 
		model (with a dense/sparse decision layer) to image candidates used in 
		our 
		MTurk Study. (b) Distribution of images 
		selected by annotators in terms of  
		the ranking of their target class probability. Here, a rank of $k$ 
		implies that the selected image has the $k$-th highest 
		probability of the target 
		class (out of the 3 images) according to the model. (c) Per-class 
		accuracy 
		of annotators in simulating the predictions of models with 
		dense/sparse decision layer.}
\end{figure}

\paragraph{Task setup}
Recall that our objective is to assess how effectively annotators are able to 
simulate the predictions of a model when they are exposed to its (dense or sparse) decision layer.
To this end, we first randomly select 100 ImageNet classes.
Then, for each such `target class' and decision layer (dense/sparse) pair, we 
created a task by: 

\begin{enumerate}
	\item \textbf{Selecting deep features:} We  randomly-select \emph{five} 
	deep features utilized by the 
	decision layer to recognize objects of the target class. To make the 
	comparison more fair, we restrict our attention to deep features that are 
	assigned significant weight (>5\% of the maximum) by the 
	corresponding model. We then present these deep 
	features to annotators via feature visualizations. Also shown 
	alongside are the (normalized and rescaled) linear coefficients for each 
	deep feature. 
	
	\item \textbf{Selecting test inputs:} We rank all the test set ImageNet 
	images based on the probability assigned by the corresponding model 
	(i.e., the ResNet-50 with a dense/sparse decision layer) to the target class.
	We then randomly select three images, such that they lie in the following 
	percentile ranges in terms of target class probability: (90, 95), (98, 99) 
	and (99.99, 100). Note that since ImageNet has 1000 diverse object 
	categories, the 
	target class probability of a randomly sampled image from the dataset is 
	likely to be extremely small. 
	Thus, fixing the percentiles as described above allows us to pick image 
	candidates that are: (i) somewhat relevant to the target class; and (ii) of 
	comparable difficulty for both types of decision layers. In 
	Figure~\ref{fig:logits}, we present the target probability distribution as per 
	the model for image candidates selected in this manner.
\end{enumerate}

Finally, annotators are presented with the deep features chosen 
above---describing them as patterns used by an AI model to recognize 
objects of a certain (unspecified) type.
They are then asked to pick one of the image candidates 
(randomly-permuted) that best matches the patterns. 
Annotators are also asked to mark their confidence  
on a likert scale. 
A sample task is shown in Figure~\ref{fig:app_task_sim}.

For each target label-decision layer pair, we obtain 10 tasks by repeating the 
random selection process above. This results in a total of 2000 tasks (100 
classes x 2 models x 10 tasks/(class, model)).
Each task is presented to 5 annotators, compensated at \$0.04 per task.

\paragraph{Quality control}
For each task, we aggregated results over all the annotators. While doing so, 
we eliminated individual instances where a particular annotator made no 
selections. We also completely eliminated instances corresponding to 
annotators who consistently (>80\% of the times) left the tasks blank. 
Finally, 
while reporting our 
results, we only keep tasks for which we have selections from at least two (of 
five) annotators. We determine the final selection based on a majority vote 
over annotators, weighted by their confidence.

\paragraph{Results}

In Table~\ref{tab:app_mturk_sim}, we report annotator accuracy---in terms 
of their ability to correctly identify the image with the highest target class 
probability as per the model. We also present a break down of the overall 
accuracy depending on whether or not the ``correct image'' is from the 
target class.
We find that sparsity significantly boosts annotators' ability to intuit 
(simulate) the model---by nearly 30\%. In fact, their performance on models 
with dense decision layers is close to chance (33\%). Note also that for  
models with sparse decision layers, annotators are able to correctly simulate 
the predictions \emph{even} when the correct image belongs to a different 
class.

\begin{table}[!h]
	\begin{center}
		\begin{tabular}{ cc>{\bfseries}c } 
			\toprule
			Accuracy (\%) & Dense & \normalfont{Sparse}\\ 
			\midrule
			Overall & 35.61 $\pm$ 3.09& 63.02 $\pm$ 3.02 \\
			From target class & 44.02 $\pm$ 5.02 & 72.22 $\pm$ 4.74  \\ 
			From another class & 30.64 $\pm$ 3.65 & 57.33 $\pm$ 4.00 \\ 
			\bottomrule
		\end{tabular}
	\end{center}
	\caption{Accuracy of annotators at simulating the model given 
		explanations from the dense and sparse classifiers.}
	\label{tab:app_mturk_sim}
\end{table}

In Figure~\ref{fig:ranking_comparison}, we visualize how the image 
selected by annotators ranks in terms of the model's 
target class probability, over all tasks. Note that a rank of one implies that 
the annotators 
correctly selected the image which the model predicts as having highest 
target class probability. This figure largely corroborates the findings 
in~\ref{tab:app_mturk_sim}---in particular, highlighting that for the standard 
(dense) decision layer, annotator selections are near-random. In 
Figure~\ref{fig:relative_accuracy}, we visualize annotator 
accuracy---aggregated per (the 10 tasks for a) target class---for models with 
dense and sparse decision layers.

%% file: appendix_visualization.tex
\subsection{Additional comparisons of features}
\label{app:visualizations}
\label{app:feature_int}

\begin{figure}[h]
	\centering
	\begin{subfigure}{0.45\textwidth}
		\centering
		\includegraphics[width=\columnwidth]{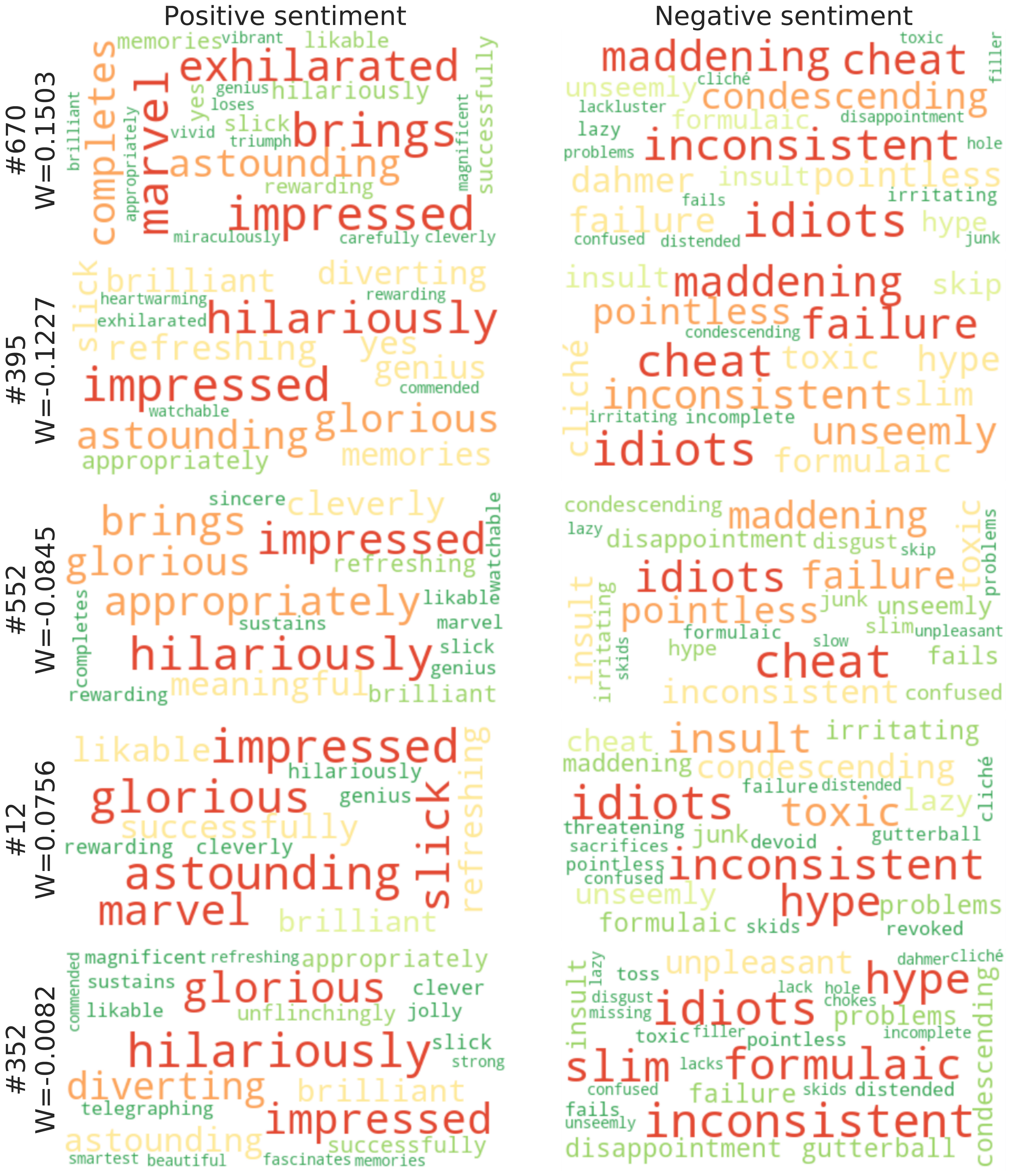}
		\caption{}
	\end{subfigure}
	\hfill
	\begin{subfigure}{0.45\textwidth}
		\centering
		\includegraphics[width=\columnwidth]{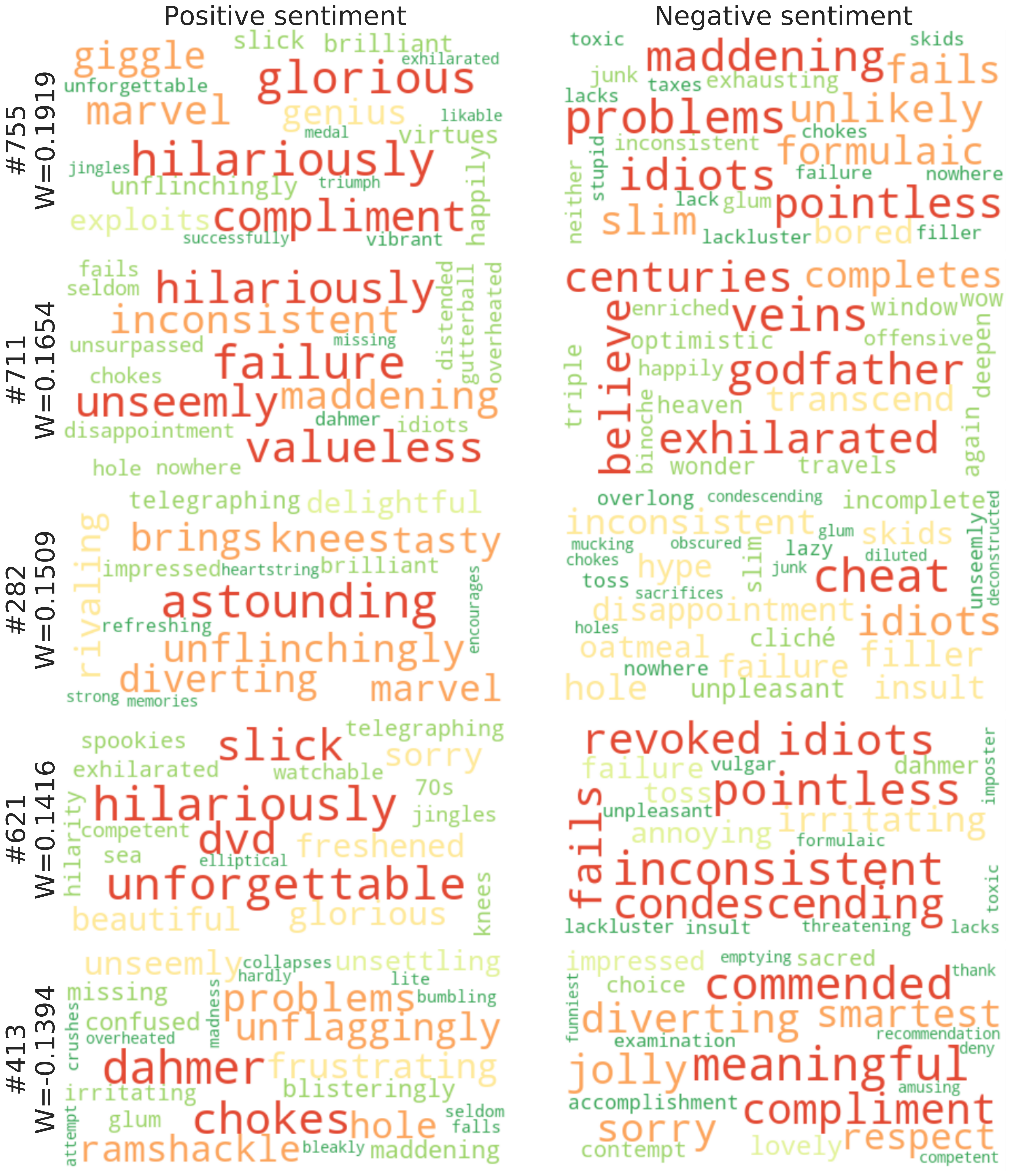}
		\caption{}
	\end{subfigure}
	\begin{subfigure}{0.9\textwidth}
		\centering
		\includegraphics[width=\columnwidth]{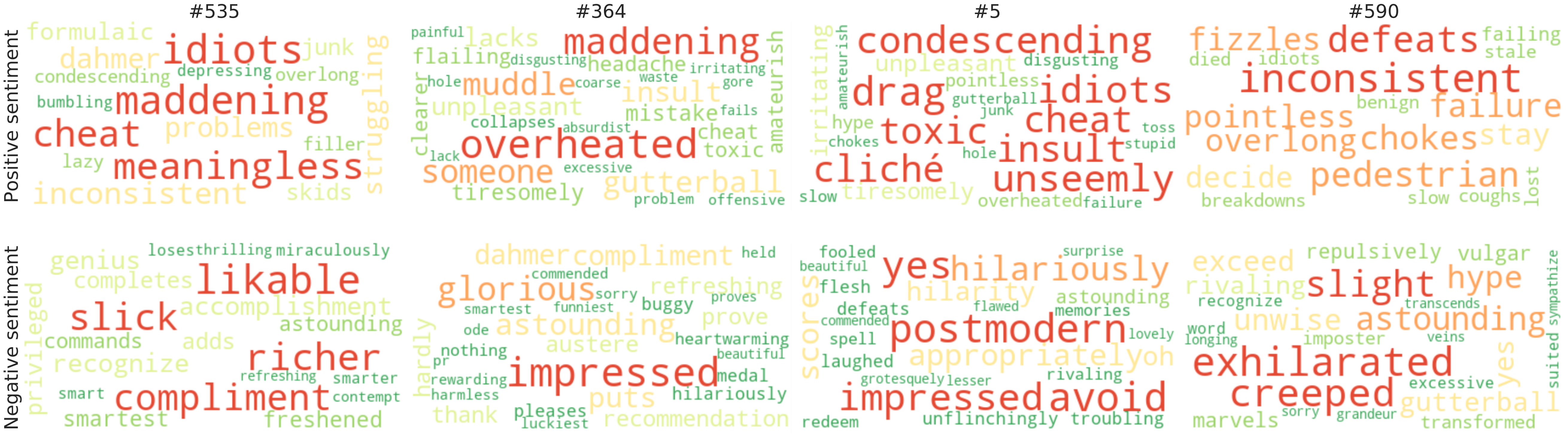}
		\caption{}
	\end{subfigure}
	\caption{Additional SST word clouds visualizing the positive and negative 
	activations for the top 5 features of the (a) sparse decision layer, (b) 
	dense decision layer, and (c) additional randomly-selected features (positive or negative weighting is according to the dense decision layer). While 
	the sparse model focuses on features that have clear positive and negative 
	semantic meaning in their word clouds, the dense model and the other 
	randomly-selected features are noticeably more mixed in sentiment. }
	\label{fig:app_wordclouds}
\end{figure}

\begin{figure}[!h]
	\begin{subfigure}{1\textwidth}
		\centering
		\includegraphics[width=0.8\columnwidth]{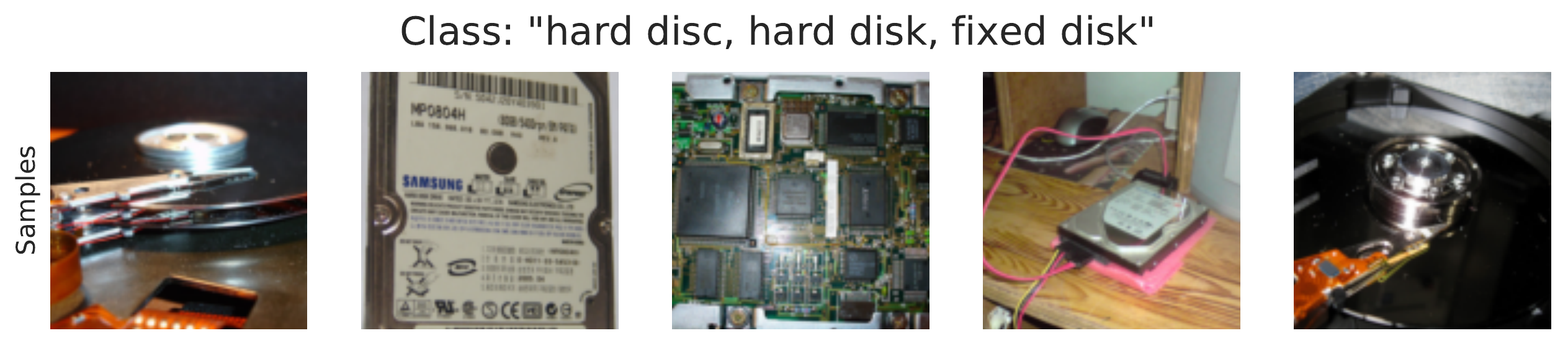}
		\caption{Class samples}
	\end{subfigure}
	\begin{subfigure}{1\textwidth}
		\centering
		\includegraphics[width=0.8\columnwidth]{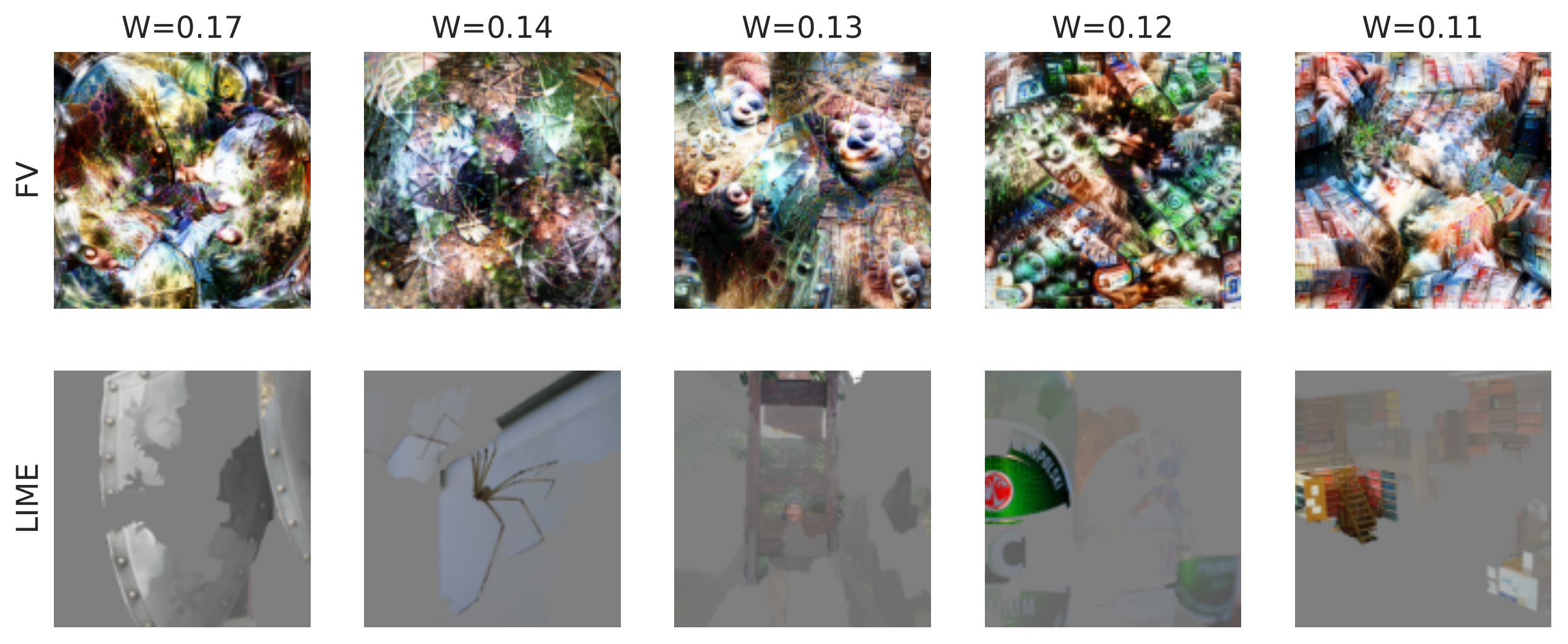}
		\caption{Dense}	
	\end{subfigure}
	\begin{subfigure}{1\textwidth}
		\centering
		\includegraphics[width=0.8\columnwidth]{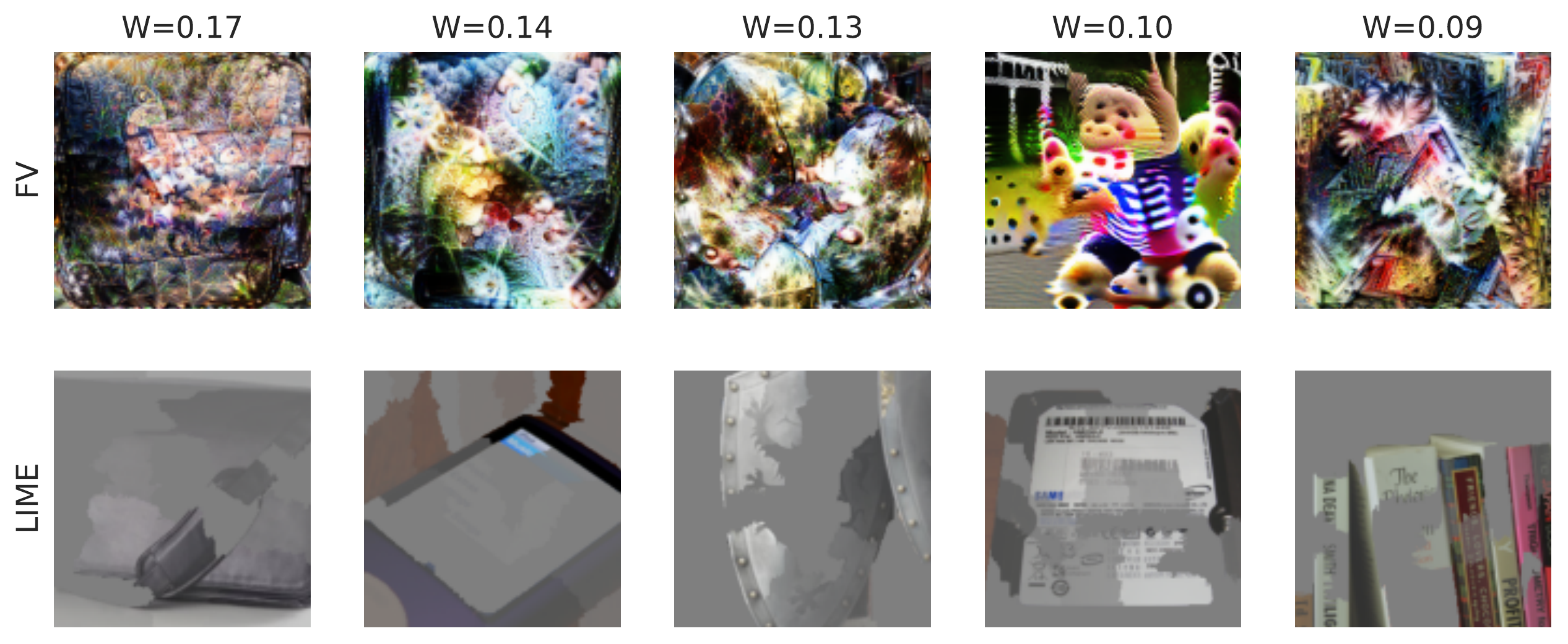}
		\caption{Sparse}	
	\end{subfigure}
	\caption{Deep features used by a standard ($\eps=0$) 
		ResNet-50 with dense (\emph{middle}) and 
		\sparsemod s 
		(\emph{bottom})  for a randomly-chosen 
		ImageNet class. For each (deep) feature, we show its corresponding 
		linear coefficient in the decision layer (W), along with feature 
		interpretations in the form of 
		feature 
		visualizations (FV) and LIME superpixels.}
	\label{fig:app_fv_std_in_harddisk}
\end{figure}

\begin{figure}[!h]
	\begin{subfigure}{1\textwidth}
		\centering
		\includegraphics[width=0.8\columnwidth]{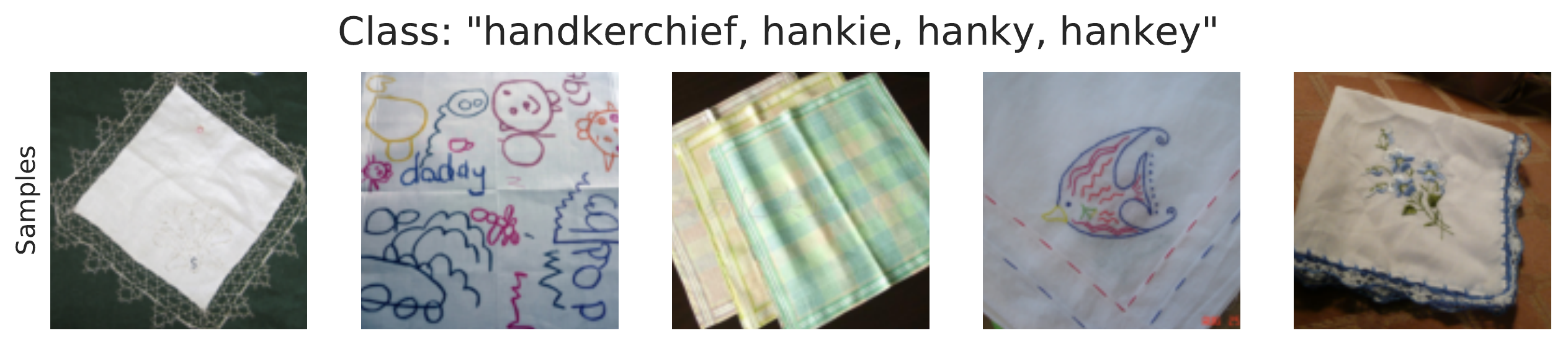}
		\caption{Class samples}
	\end{subfigure}
	\begin{subfigure}{1\textwidth}
		\centering
		\includegraphics[width=0.8\columnwidth]{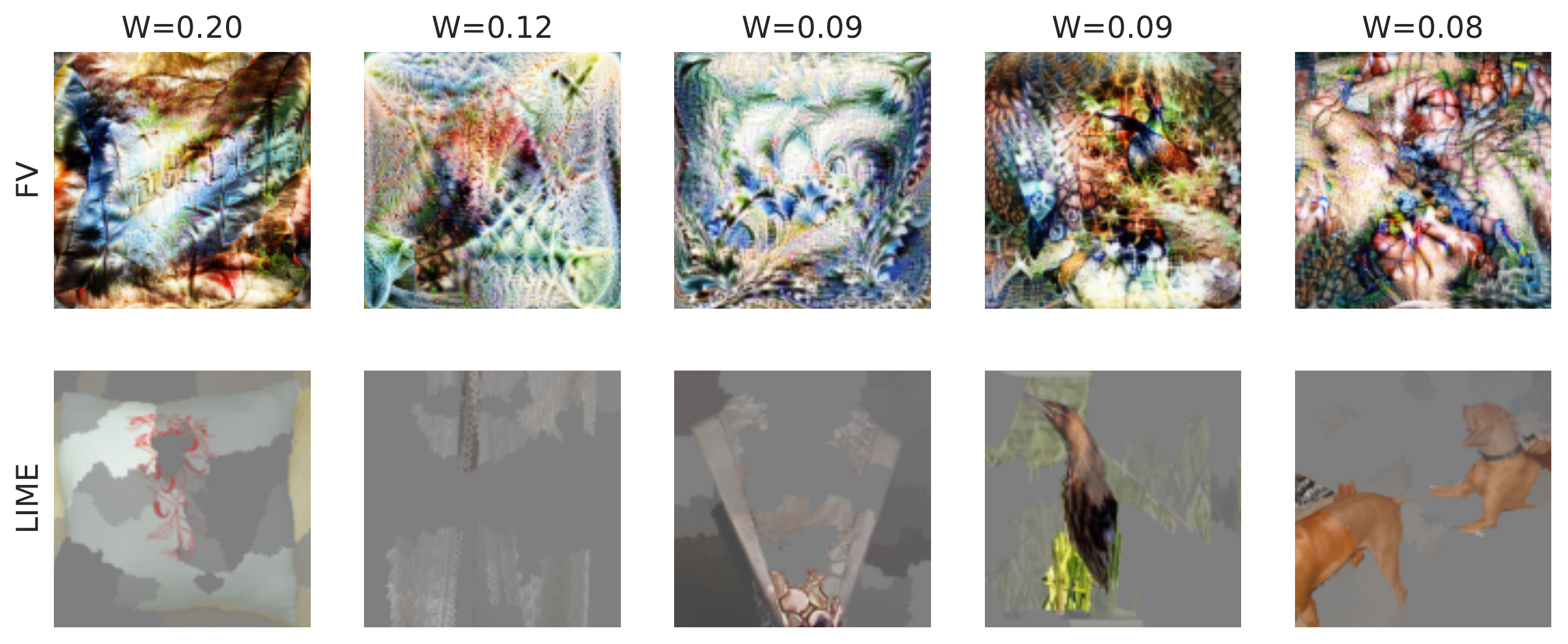}
		\caption{Dense}	
	\end{subfigure}
	\begin{subfigure}{1\textwidth}
		\centering
		\includegraphics[width=0.8\columnwidth]{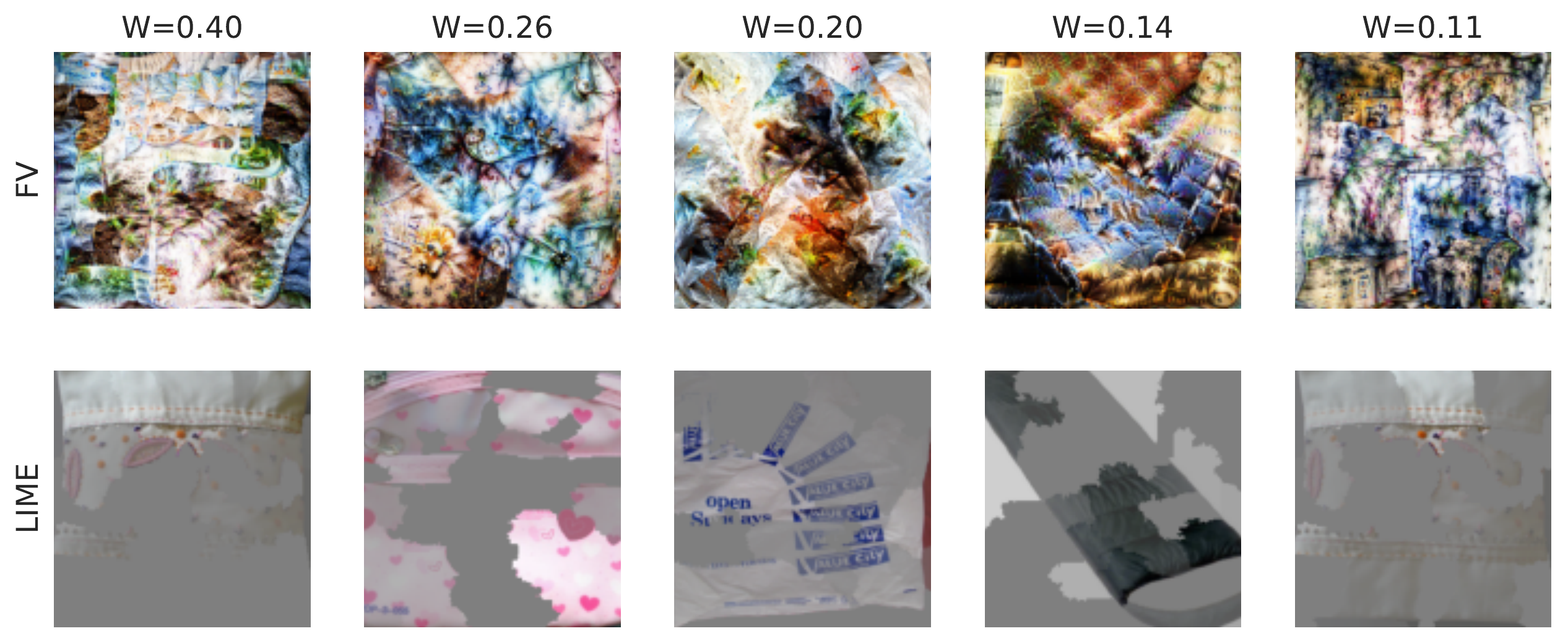}
		\caption{Sparse}	
	\end{subfigure}
	\caption{Deep features used by a standard ($\eps=0$) 
		ResNet-50 with dense (\emph{middle}) and 
		\sparsemod s 
		(\emph{bottom})  for a randomly-chosen 
		ImageNet class. For each (deep) feature, we show its corresponding 
		linear coefficient in the decision layer (W), along with feature 
		interpretations in the form of 
		feature 
		visualizations (FV) and LIME superpixels.}
	\label{fig:app_fv_std_in_hanky}
\end{figure}

\begin{figure}[!h]
	\begin{subfigure}{1\textwidth}
		\centering
		\includegraphics[width=0.8\columnwidth]{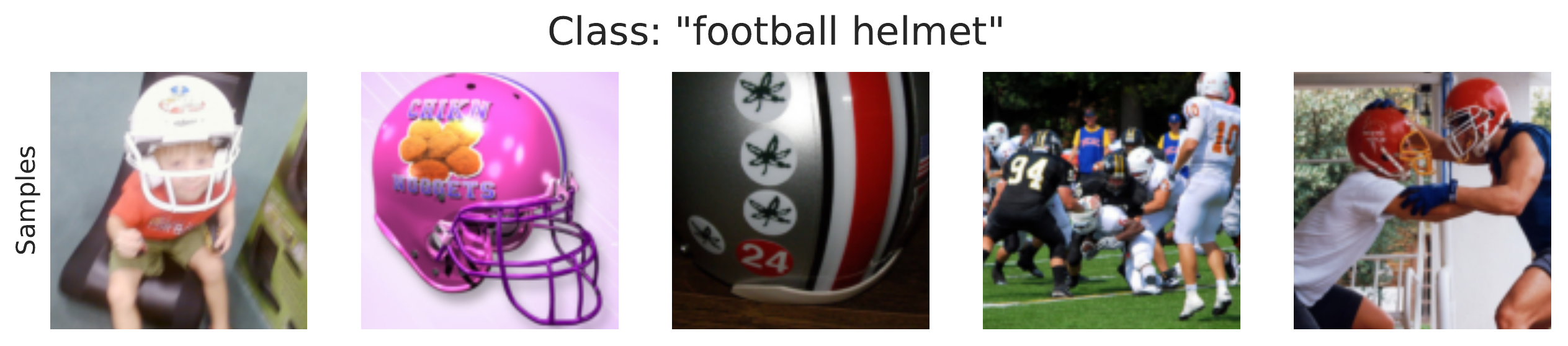}
		\caption{Class samples}
	\end{subfigure}
	\begin{subfigure}{1\textwidth}
		\centering
		\includegraphics[width=0.8\columnwidth]{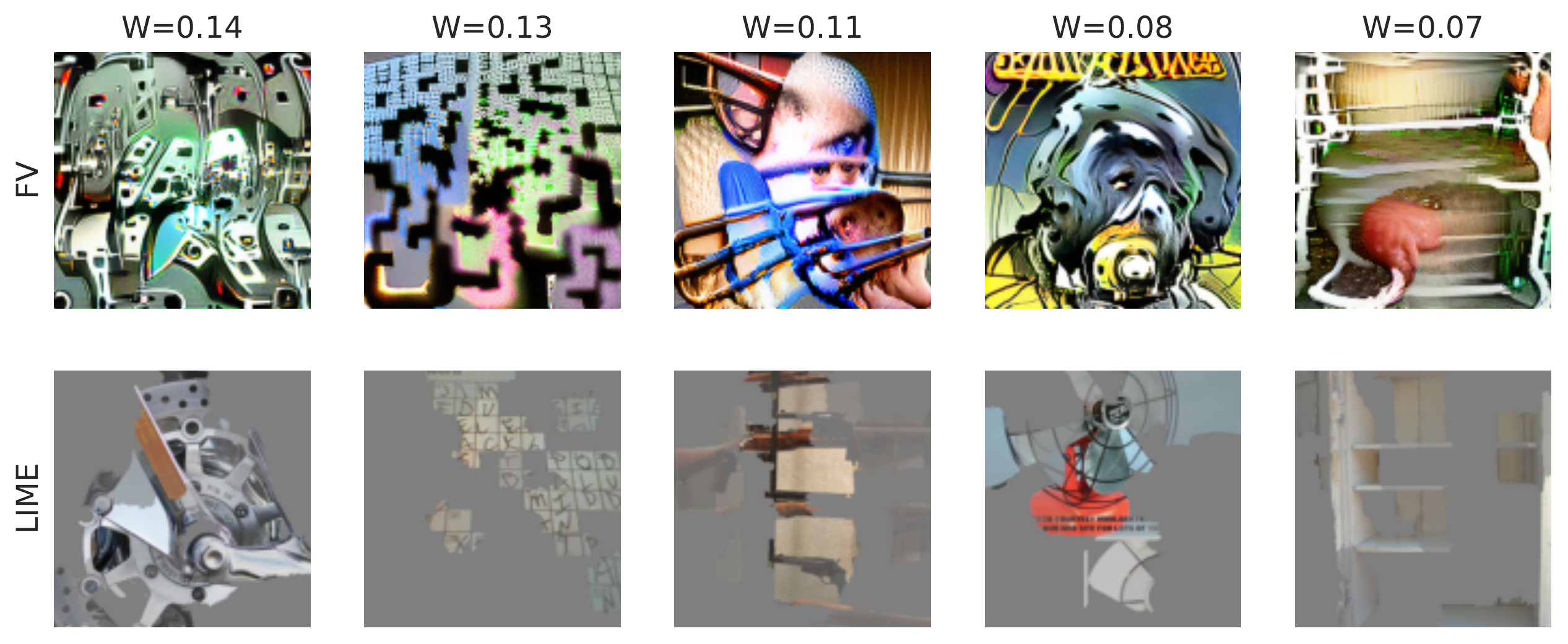}
	   \caption{Dense}	
  \end{subfigure}
	\begin{subfigure}{1\textwidth}
		\centering
		\includegraphics[width=0.8\columnwidth]{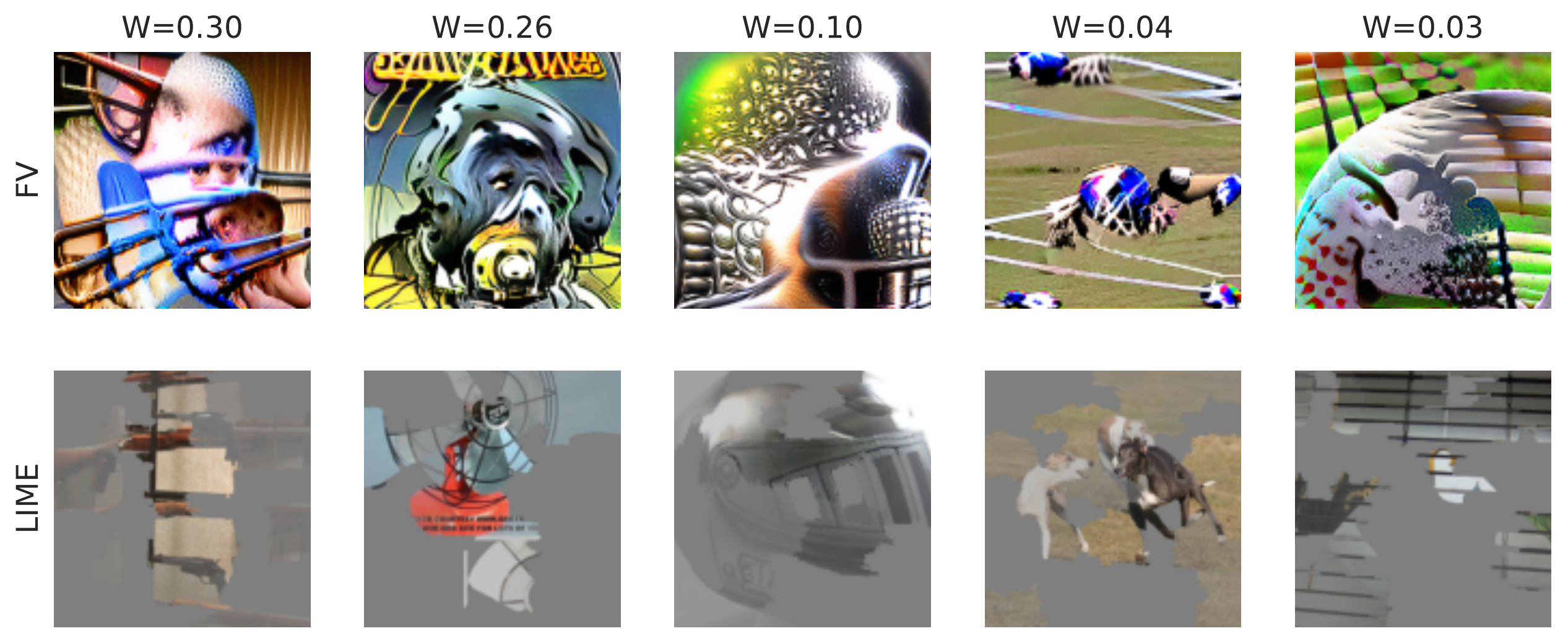}
		\caption{Sparse}	
  \end{subfigure}
	\caption{Deep features used by a adversarially-trained ($\eps=3$) 
		ResNet-50 with dense (\emph{middle}) and 
		\sparsemod s 
		(\emph{bottom})  for a randomly-chosen 
		ImageNet class. For each (deep) feature, we show its corresponding 
		linear coefficient in the decision layer (W), along with feature 
		interpretations in the form of 
		feature 
		visualizations (FV) and LIME superpixels.}
	\label{fig:app_fv_rob_in_helmet}
\end{figure}

\begin{figure}[!h]
	\begin{subfigure}{1\textwidth}
		\centering
		\includegraphics[width=0.8\columnwidth]{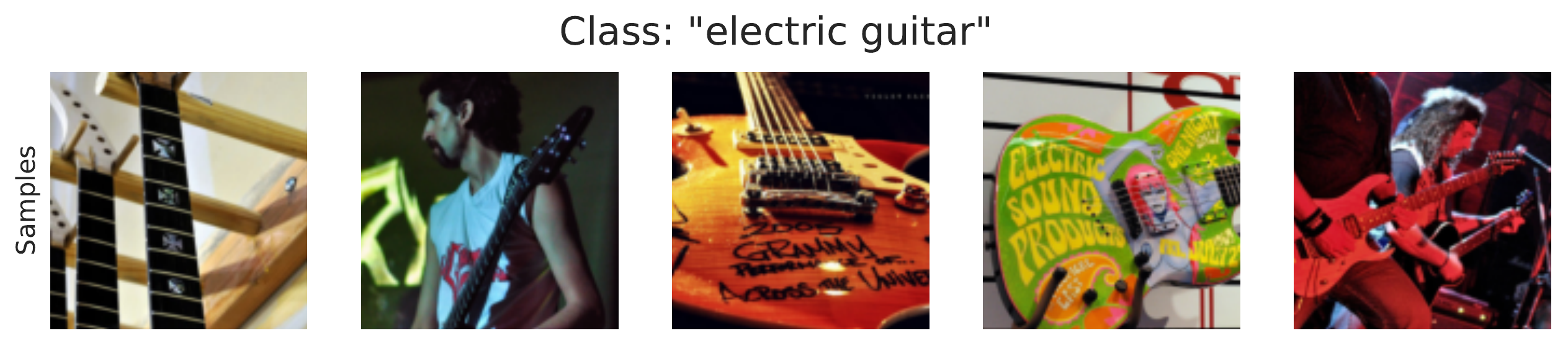}
		\caption{Class samples}
	\end{subfigure}
	\begin{subfigure}{1\textwidth}
		\centering
		\includegraphics[width=0.8\columnwidth]{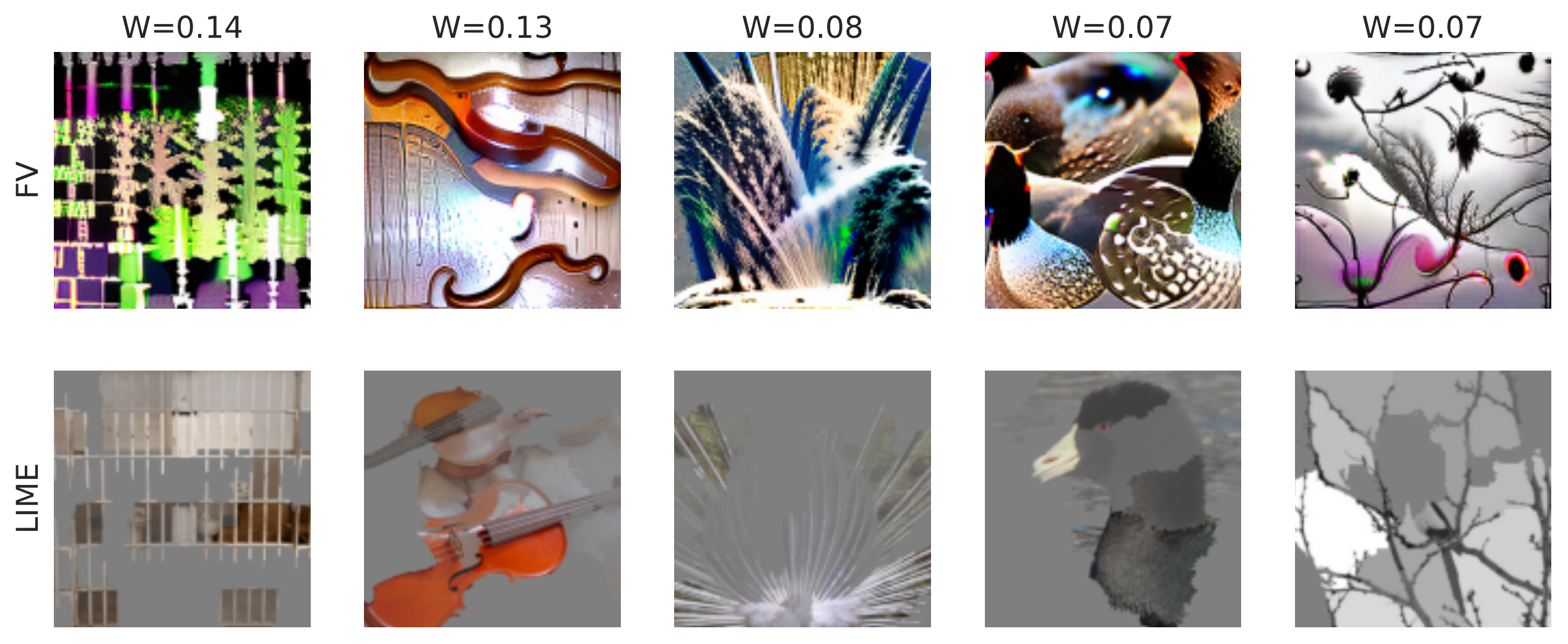}
		\caption{Dense}	
	\end{subfigure}
	\begin{subfigure}{1\textwidth}
		\centering
		\includegraphics[width=0.8\columnwidth]{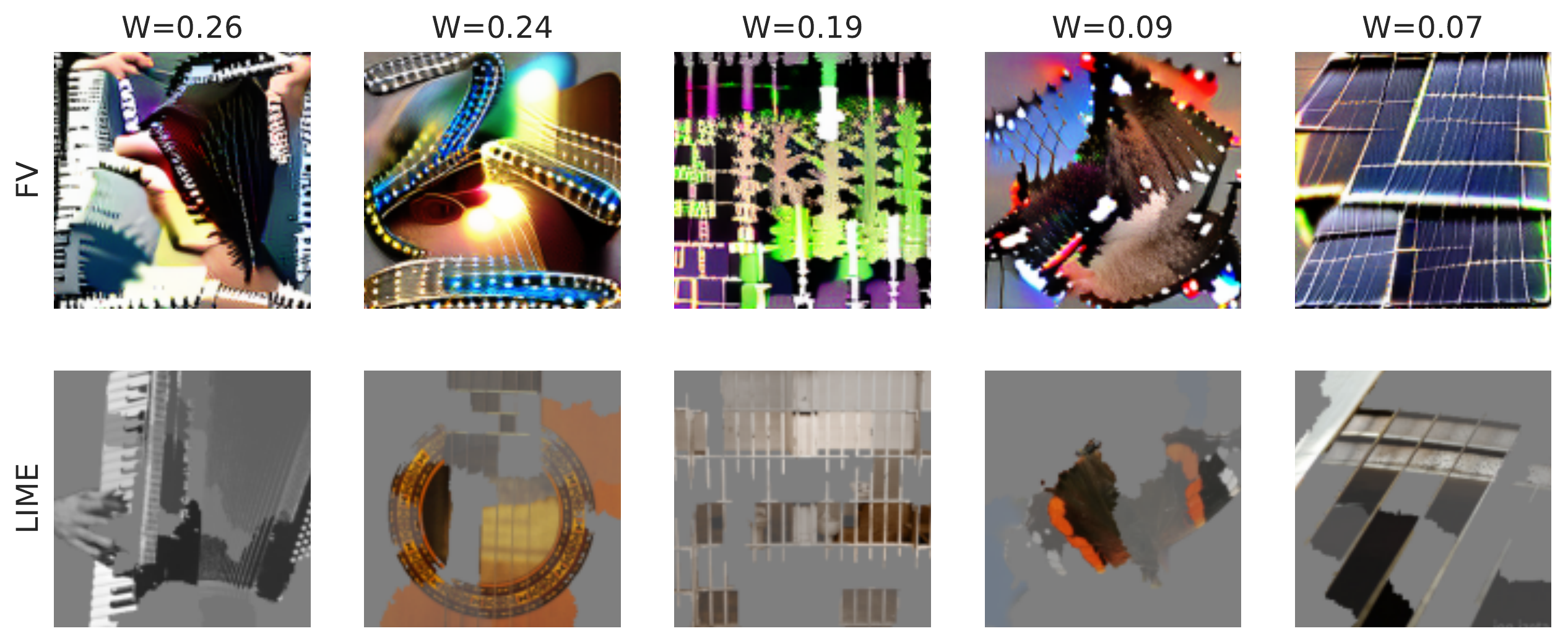}
		\caption{Sparse}	
	\end{subfigure}
	\caption{Deep features used by a adversarially-trained ($\eps=3$) 
		ResNet-50 with dense (\emph{middle}) and 
		\sparsemod s 
		(\emph{bottom})  for a randomly-chosen 
		ImageNet class. For each (deep) feature, we show its corresponding 
		linear coefficient in the decision layer (W), along with feature 
		interpretations in the form of 
		feature 
		visualizations (FV) and LIME superpixels. }
	\label{fig:app_fv_rob_in_electricguitar}
\end{figure}

\begin{figure}[!h]
	
	\begin{subfigure}{1\textwidth}
		\centering
		\includegraphics[width=0.8\columnwidth]{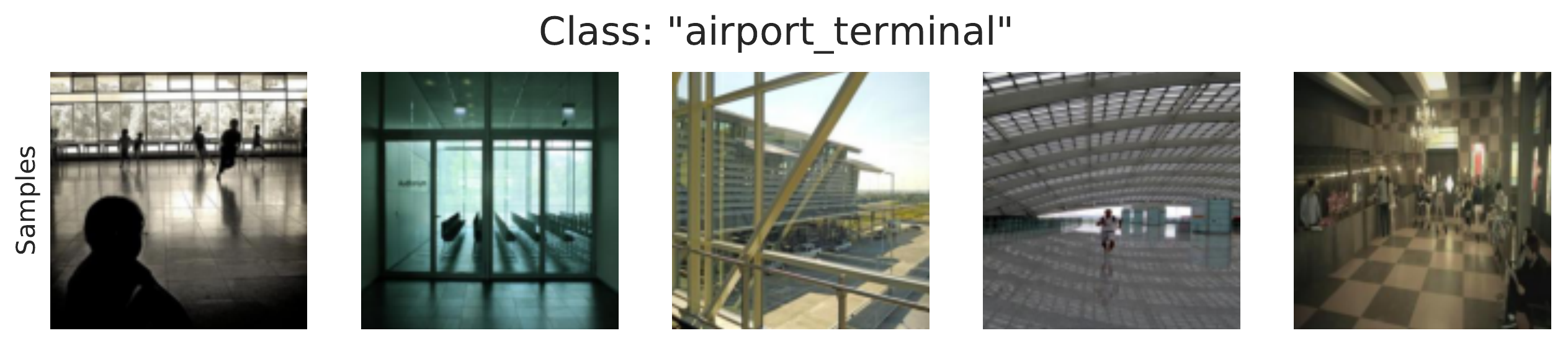}
		\caption{Class samples}	
	\end{subfigure}
	\begin{subfigure}{1\textwidth}
		\centering
		\includegraphics[width=0.2\columnwidth]{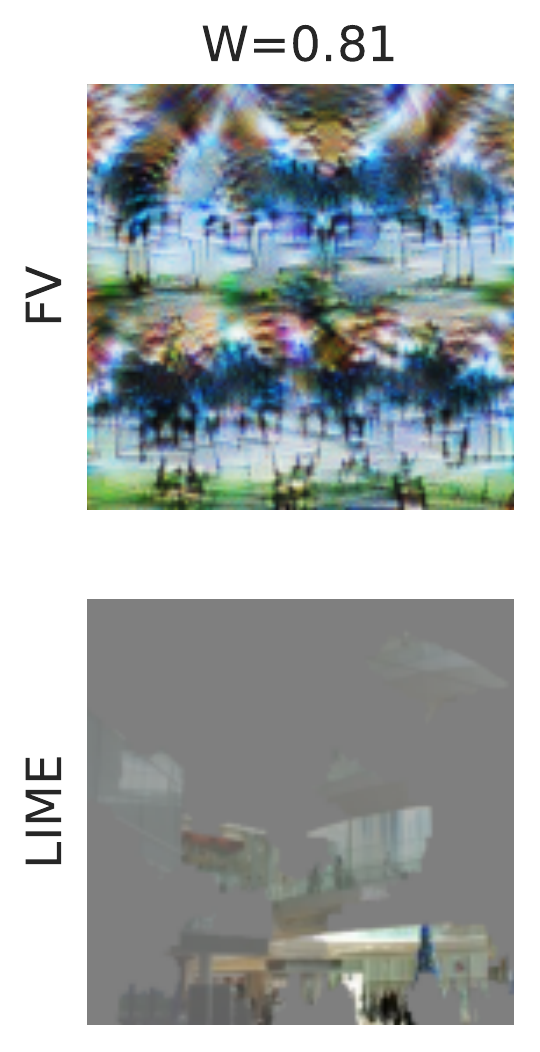}
		\caption{Dense}	
	\end{subfigure}
	\begin{subfigure}{1\textwidth}
		\centering
		\includegraphics[width=0.2\columnwidth]{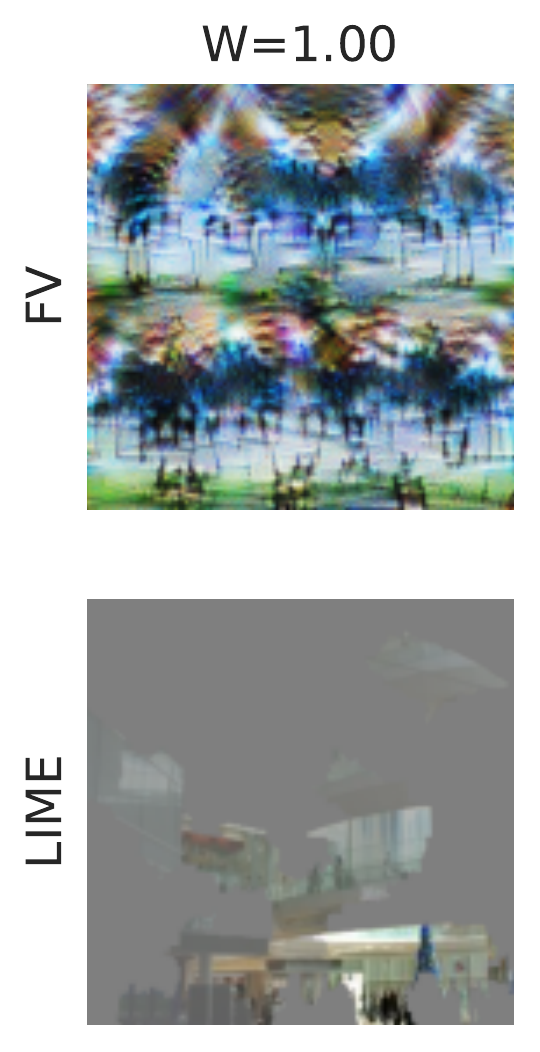}
		\caption{Sparse}	
	\end{subfigure}
	\caption{Deep features used by a standard ($\eps=0$) 
		ResNet-50 with dense (\emph{middle}) and 
		\sparsemod s 
		(\emph{bottom})  for a randomly-chosen 
		Places-10 class. For each (deep) feature, we show its corresponding 
		linear coefficient in the decision layer (W), along with feature 
		interpretations in the form of 
		feature 
		visualizations (FV) and LIME superpixels.}
	\label{fig:app_fv_std_places}
\end{figure}

\begin{figure}[!h]
	
	\begin{subfigure}{1\textwidth}
		\centering
		\includegraphics[width=0.8\columnwidth]{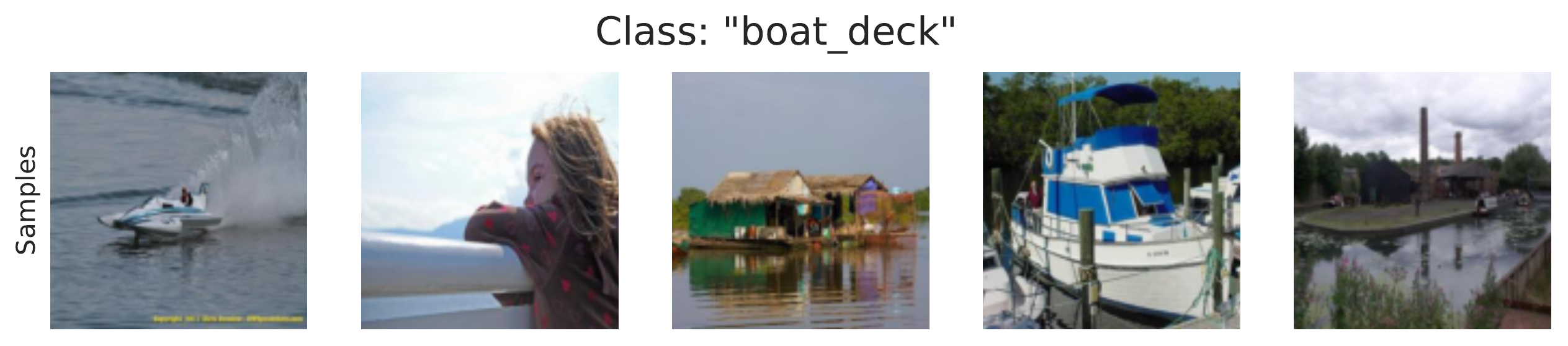}
		\caption{Class samples}	
	\end{subfigure}
	\begin{subfigure}{1\textwidth}
		\centering
		\includegraphics[width=0.8\columnwidth]{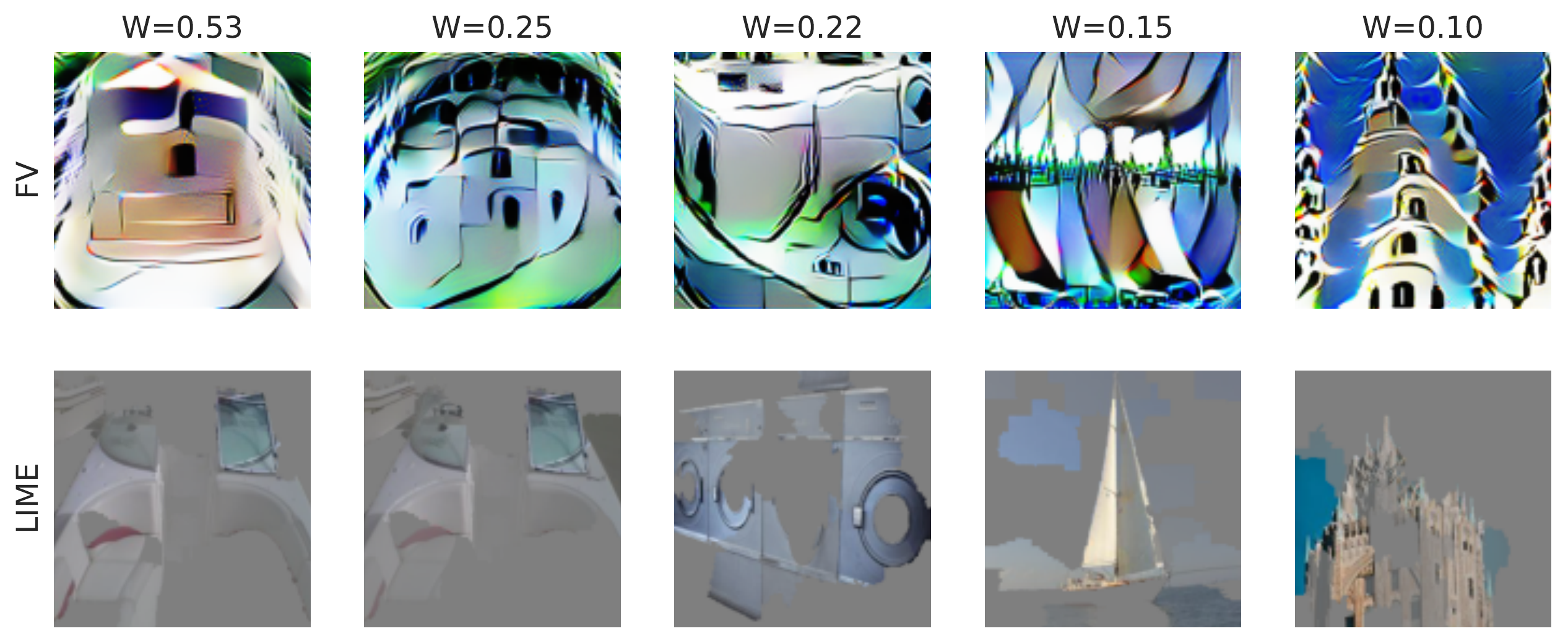}
		\caption{Dense}	
	\end{subfigure}
	\begin{subfigure}{1\textwidth}
		\centering
		\includegraphics[width=0.8\columnwidth]{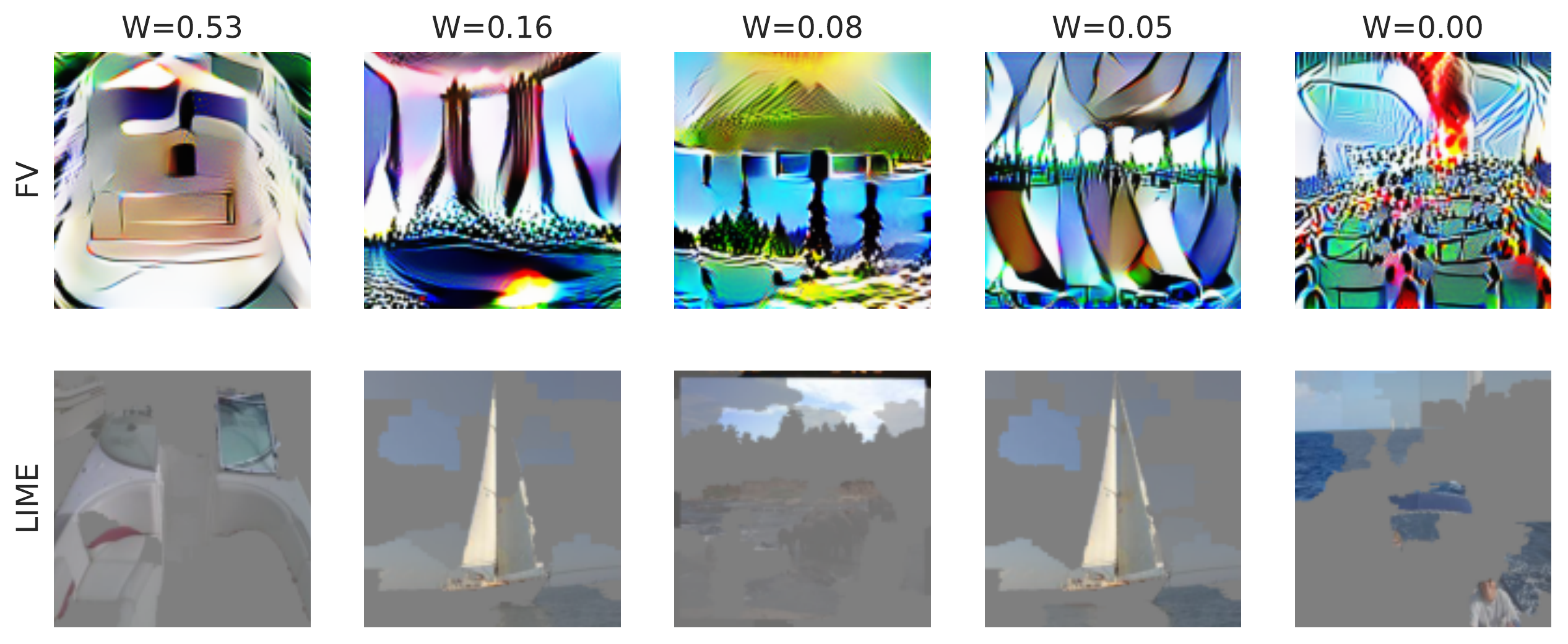}
		\caption{Sparse}	
   \end{subfigure}
	\caption{Deep features used by a adversarially-trained ($\eps=3$) 
		ResNet-50 with dense (\emph{middle}) and 
		\sparsemod s 
		(\emph{bottom})  for a randomly-chosen 
		Places-10 class. For each (deep) feature, we show its corresponding 
		linear coefficient in the decision layer (W), along with feature 
		interpretations in the form of 
		feature 
		visualizations (FV) and LIME superpixels.}
	\label{fig:app_fv_rob_places}
\end{figure}

In Figure~\ref{fig:app_wordclouds}, we visualize additional deep features 
used by BERT models with \sparsemod s for the SST sentiment analysis task.
Figures~\ref{fig:app_fv_std_in_harddisk}-~\ref{fig:app_fv_rob_places} show 
feature 
interpretations of deep features used by ResNet-50 classifiers with 
\sparsemod s trained on ImageNet and Places-10. 
Due to space constraints, we limit the feature interpretations for vision 
models to (at most) five 
randomly-chosen deep features used by the dense/\sparsemod{} in 
Figure~\ref{fig:suite} and 
Figures~\ref{fig:app_fv_std_in_harddisk}-~\ref{fig:app_fv_rob_places}. To 
allow for a fair comparison between the two decision layers, we sample these 
features as follows. Given a target class, we first determine the number of 
deep features ($k$) used by the \sparsemod{} to recognize objects of that 
class. Then, for both decision layers, we randomly sample five deep features 
from the top-$k$ highest weighted ones (for that class).

\newpage
\subsubsection{Language models}
\label{app:sst_wordclouds}

\clearpage
\subsubsection{Vision models}
\label{app:imagenet_visualizations}

%% file: appendix_bias.tex
\section{Model biases and spurious correlations}
\label{app:bias}

\subsection{Toxic comments}
\label{app:toxic}
In this section, we visualize the word clouds for the toxic comment classifiers which 
reveal the biases that the model has learned from the data. Note that these figures are heavily redacted due to the nature of these comments. 

\begin{figure}[!h]
	\centering
	\begin{subfigure}{0.45\textwidth}
		\centering
		\includegraphics[width=\columnwidth]{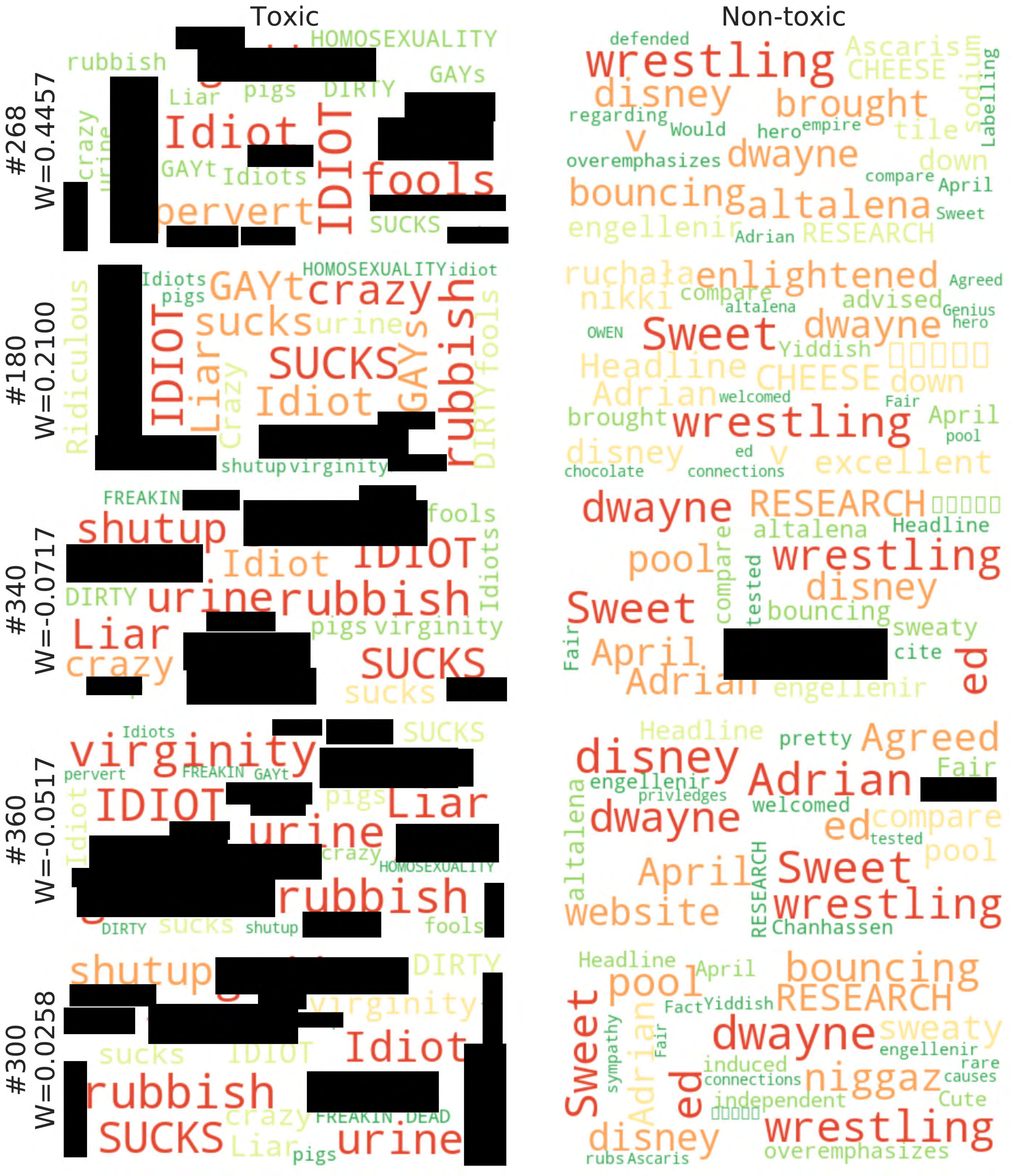}
		\caption{}
		\label{fig:app_toxic_bert_sparse}
	\end{subfigure}
	\hfill
	\begin{subfigure}{0.45\textwidth}
		\centering
		\includegraphics[width=\columnwidth]{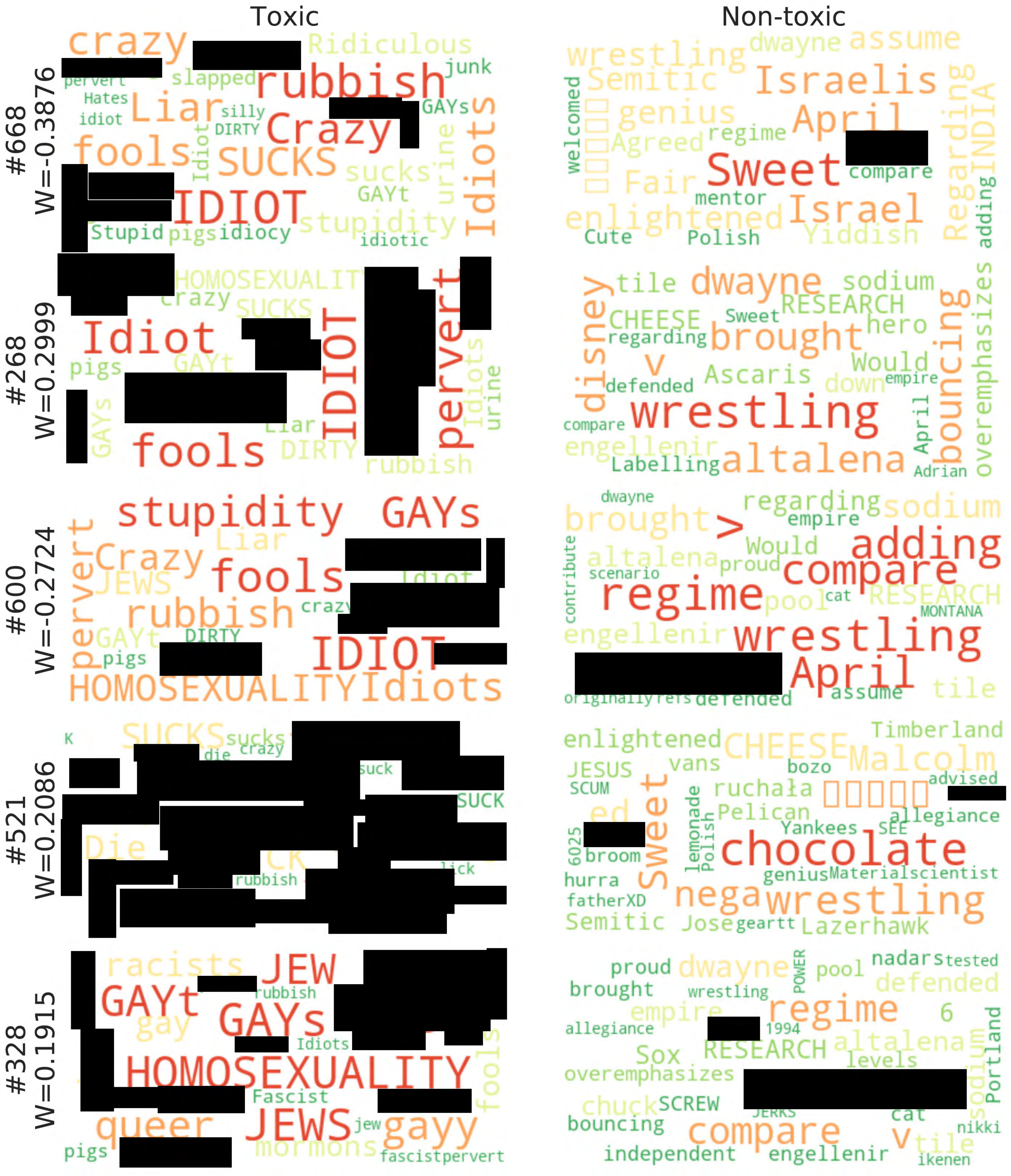}
		\caption{}
		\label{fig:app_toxic_bert_dense}
	\end{subfigure}
	\caption{Word cloud visualizations of the top 5 deep features in 
	Toxic-BERT for the (a) sparse decision layer and (b) dense decision layer}
	\label{fig:app_toxic_bert}
\end{figure}

In Figure~\ref{fig:app_toxic_bert}, we visualize the top five features for the 
sparse (Figure \ref{fig:app_toxic_bert_sparse}) and dense (Figure 
\ref{fig:app_toxic_bert_dense}) decision layers of Toxic-Bert. We note that 
more of the words in the sparse decision layer refer to identity groups, 
whereas this is less clear in the dense decision layer. Even if we expand our 
interpretation to the top 10 neurons with the largest weight, only 7.5\% 
of the words refer to identity groups for the model with a dense decision 
layer. 

\begin{figure}[!h]
	\centering
	\begin{subfigure}{0.48\textwidth}
		\centering
		\includegraphics[width=0.95\columnwidth]{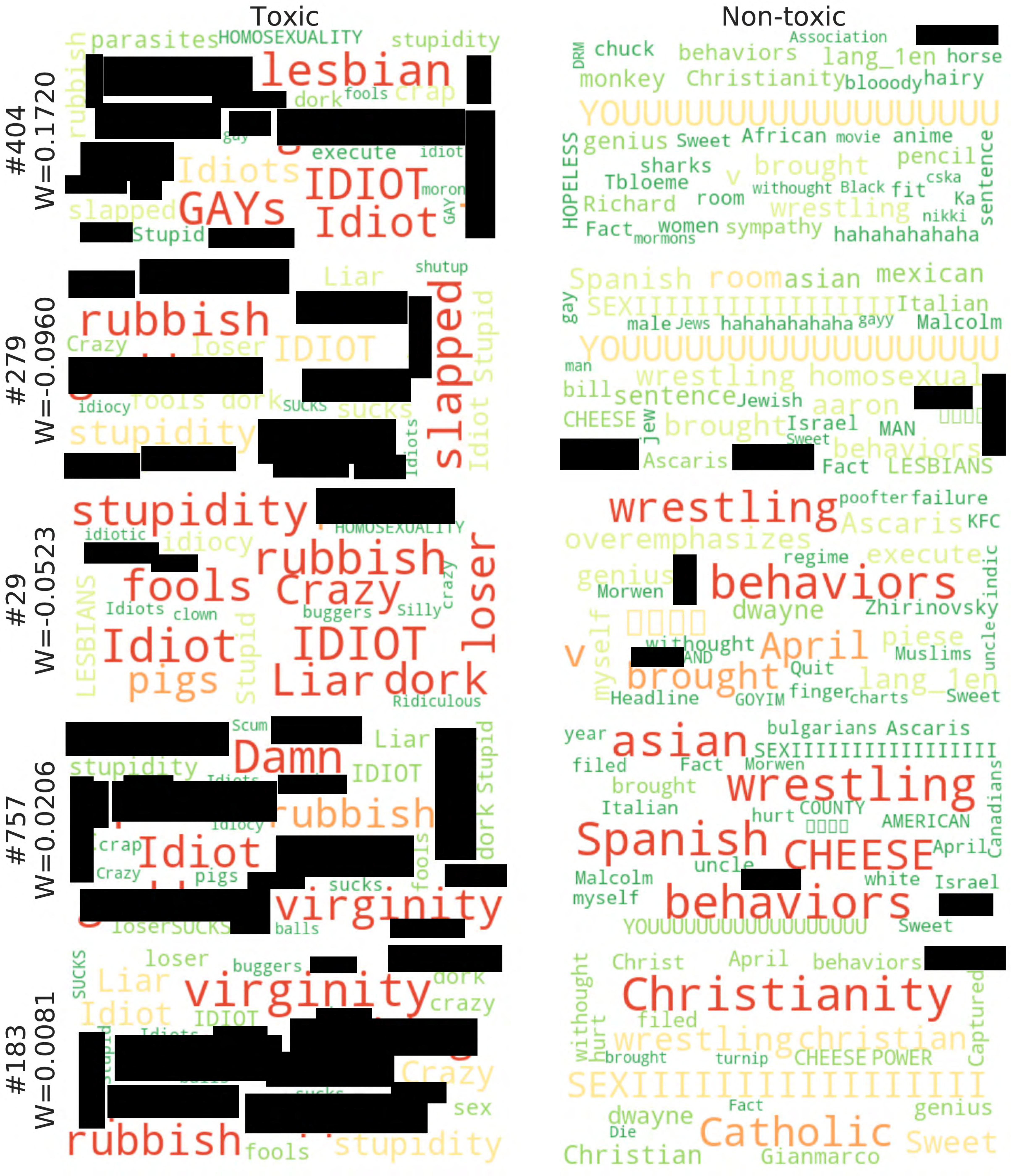}
		\caption{}
		\label{fig:app_debiased_bert_sparse}
	\end{subfigure}
	\hfill
	\begin{subfigure}{0.48\textwidth}
		\centering
		\includegraphics[width=0.95\columnwidth]{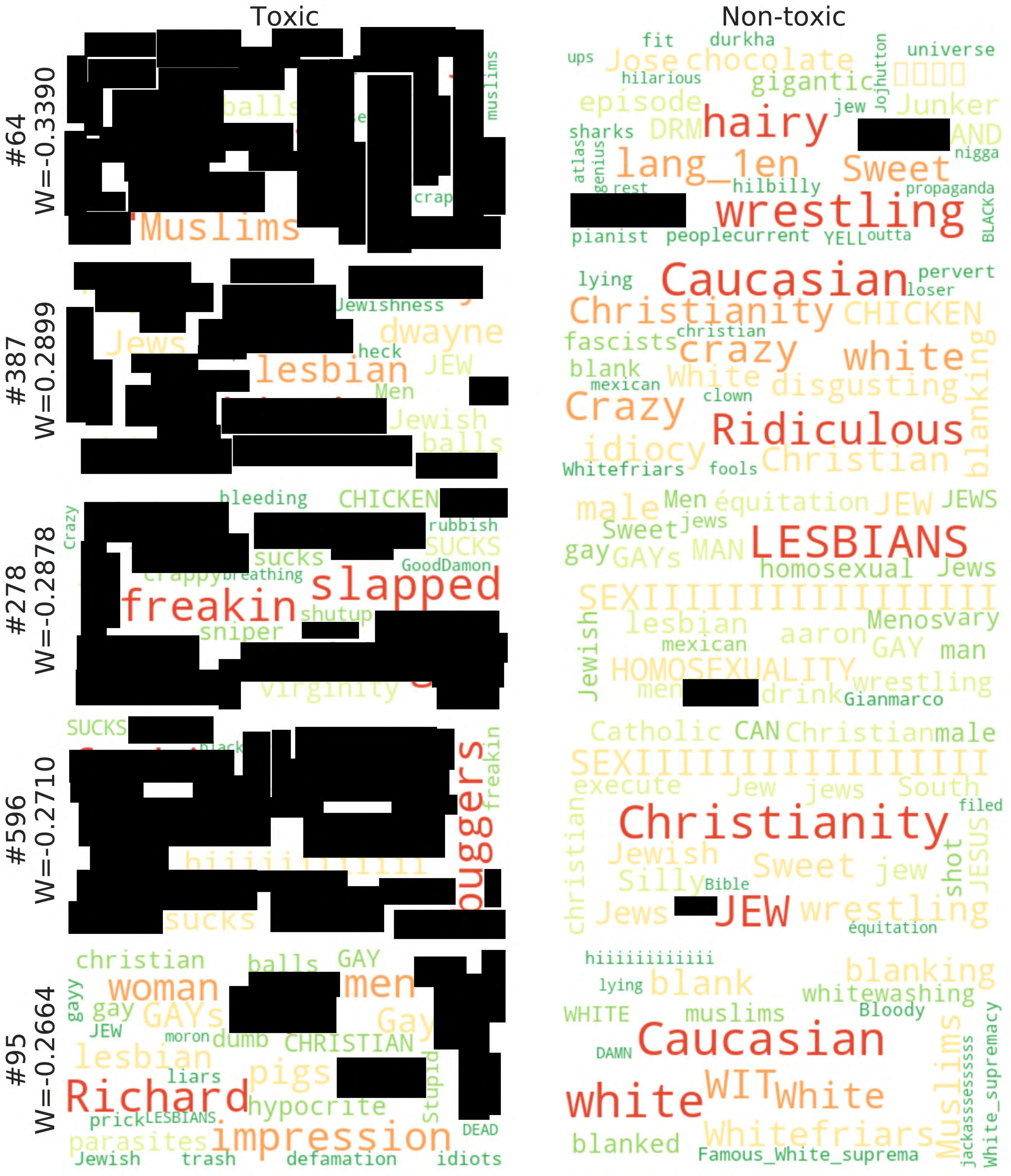}
		\caption{}
		\label{fig:app_debiased_bert_dense}
	\end{subfigure}
	\caption{Word cloud visualizations of the top 5 deep features in 
	Debiased-BERT for the (a) sparse decision layer and (b) dense decision 
	layer}
	\label{fig:app_debiased_bert}
\end{figure}

In Figure~\ref{fig:app_debiased_bert}, we perform a similar visualization as for 
the Toxic-BERT model, but for the Debiased-BERT model. The word clouds for 
the sparse decision layer (Figure \ref{fig:app_debiased_bert_sparse}) provide 
evidence that the Debiased-BERT model no longer uses identity words as 
prevalently for identifying toxic comments. However, it is especially clear 
from the word clouds for the sparse decision layer that a significant fraction 
of the non-toxic word clouds contain identity words. This suggests that the 
model now uses these identity words as strong evidence for non-toxicity, 
which can be also reflected to a lesser degree in the wordclouds for the 
dense decision layer (Figure \ref{fig:app_debiased_bert_dense}). 

\subsection{ImageNet}
\label{app:imagenet_biases}
\label{app:spurious}

\subsubsection{Human study}
\label{app:mturk_spurious}

We now detail the setup of our MTurk study from Section~\ref{sec:biases}.  
For our analysis, we use a standard ResNet-50 trained on the ImageNet 
dataset---with the default (dense) decision layer, as well as its sparse 
counterpart from Figure~\ref{tab:ablation}.

\paragraph{Task setup.}
This task is designed to semi-automatically identify learned correlations in 
classifiers with dense/sparse decision layers.
To this end, we randomly-select 1000 class pairs from each model, such that 
the classes share a common deep feature in the decision layer. We only 
consider features to 
which the model assigns a substantial weight for both classes (>5\% 
maximum weight).
Then, for each class (from the pair), we select the three images that 
maximally activate the deep feature of interest.
Doing so allows us to identify the most prototypical images from each class 
for the given deep feature.

\begin{figure}[!t]
	\centering
	
	\includegraphics[width=0.7\columnwidth]{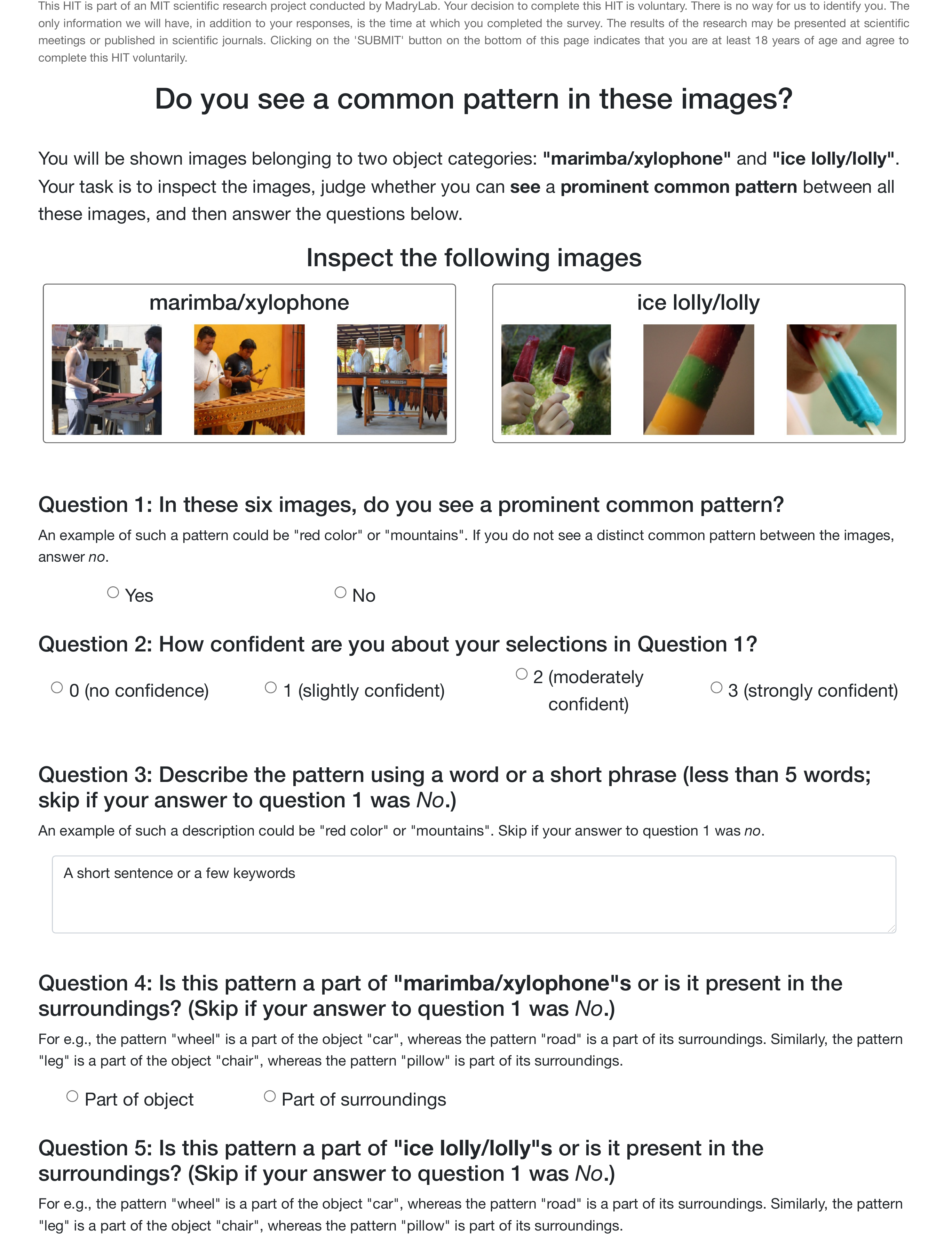}
	\caption{Sample MTurk task to diagnose (spurious) correlations in 
	deep networks via their dense/sparse decision layers.}
	\label{fig:app_task_spurious}
\end{figure}

We then present annotators on MTurk with the six chosen images, grouped 
by class 
along with the label. We ask them: (a) whether the images share a 
common pattern; (b) how confident they are about this selection on a likert 
scale; (c) to provide a short free text description of the pattern; and (d) for 
each class, to determine if the pattern is part of the class object or the 
surrounding. A sample task is shown in Figure~\ref{fig:app_task_spurious}.
Each task was presented to 5 annotators, compensated at \$0.07 per task.

\paragraph{Quality control}
For each task, we aggregated results over all the annotators. While doing so, 
we eliminated individual instances where a particular annotator made no 
selections. We also completely eliminated instances corresponding to 
annotators who consistently (>80\% of the time) left the task blank. Finally, 
while reporting our 
results, we only keep tasks for which we have selections from at least three 
(of 
five) annotators. We determine the final selection based on a majority vote 
over annotators, weighted by their confidence.

\subsubsection{Additional visualizations of spurious correlations}

\begin{figure}[!t]
	\centering
	\includegraphics[width=1\columnwidth]{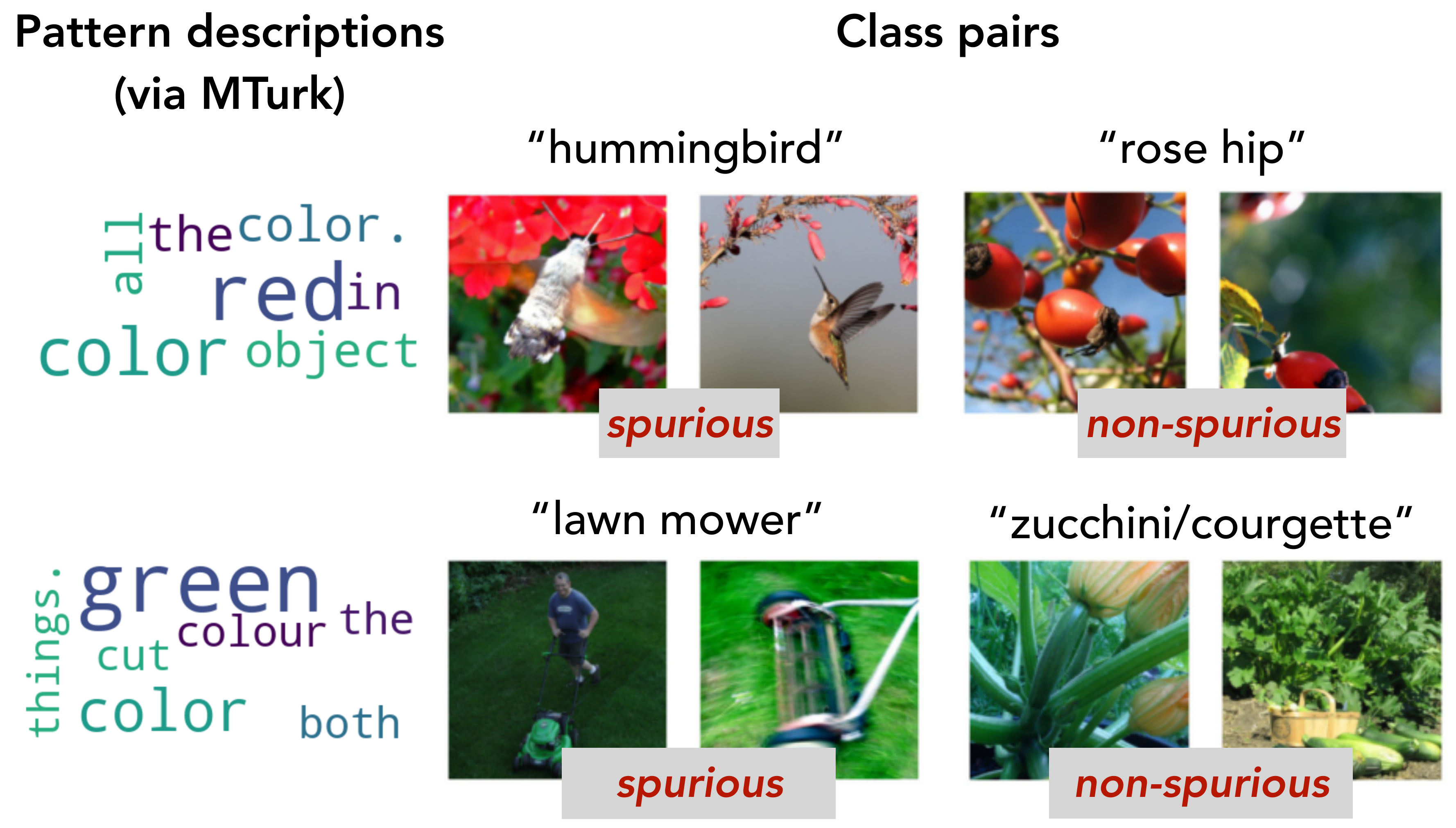}
	 \\ \vspace{0.25cm}
	\includegraphics[width=1\columnwidth]{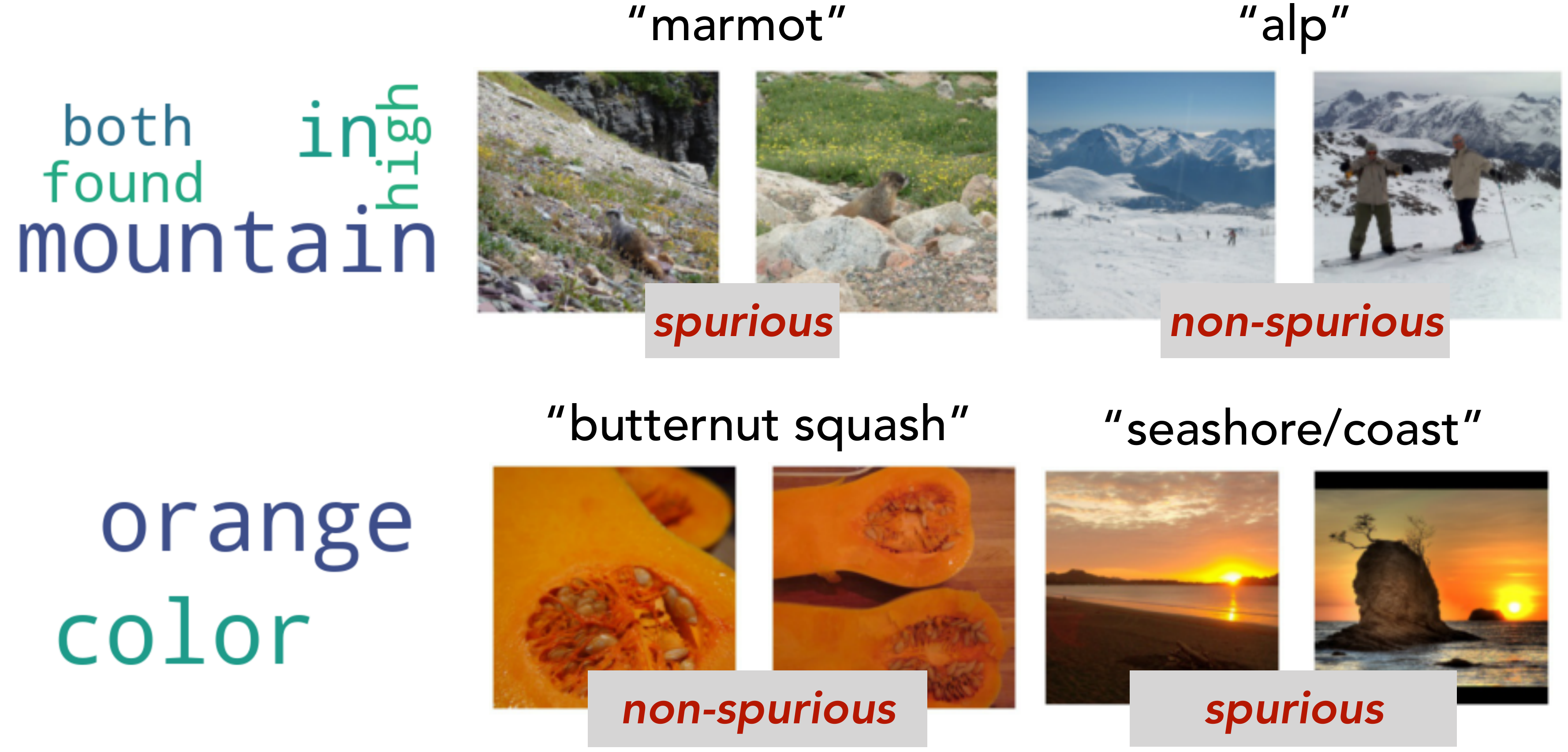}
	\caption{Additional examples of correlations in ImageNet models detected 
		using our MTurk study. 
		Each row contains protypical images from a pair of classes, along 
		with the annotator-provided descriptions for the shared deep feature 
		that 
		these images strongly activate.
		For each class, we also display if annotators marked the feature to be a 
		``spurious correlation''. \\ \\ \\ \\}
	\label{fig:app_spurious_examples}
\end{figure}

\begin{figure}[!t]
	\centering
	\includegraphics[width=1\columnwidth]{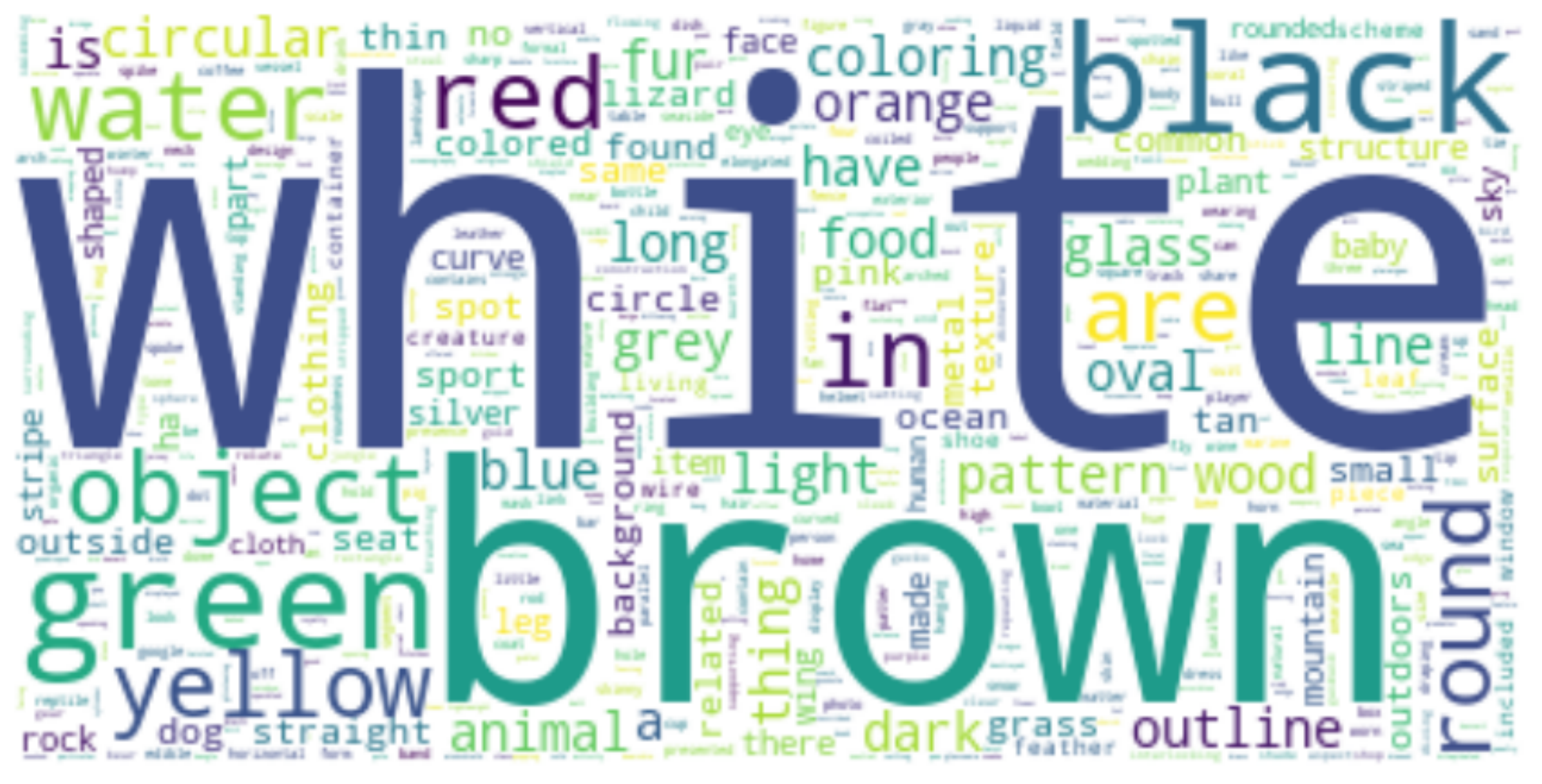}
	\caption{Word cloud visualization of descriptions provided by annotators 
		for patterns learned by ``shared deep features'' in standard 
		ImageNet-trained ResNet-50 classifiers with sparse decision layers.}
	\label{fig:app_feedback}
\end{figure}

In Figure~\ref{fig:app_spurious_examples}, we provide additional examples of 
correlations detected using our MTurk study. Then in 
Figure~\ref{fig:app_feedback}, we summarize annotator-provided 
descriptions for all the patterns identified in ImageNet classifiers with sparse 
decision layers via a word cloud. 
This visualization sheds light into the nature of correlations extracted by 
ImageNet classifiers from their training data---for instance, we see that the 
patterns most frequently identified by annotators relate to object color and 
shape.

%% file: appendix_counterfactuals.tex
\newpage~\newpage
\section{Counterfactual experiments}
\label{app:sentiment_counterfactuals}

\begin{algorithm}[t]
	\caption{Counterfactual generation for a sentence of $n$ words $x = (x_1, \dots x_n)$, 
	a deep encoder $h : \mathbb R^n \rightarrow \mathbb R^m$, and a linear decision layer with coefficients $(w,b)$.}
	\label{alg:nlp_counterfactual}
	\begin{algorithmic}[1]
		\STATE $z = h(x)$ \textit{// calculate deep features}
		\STATE $y = \argmax_y w_yz + b_y$ \textit{// calculate prediction}
		\STATE $Z^+, Z^-$ = $\emptyset, \emptyset$ \textit{// initialize candidate word substitutions}
		\FOR{$i = 1\dots m$}
		\FOR{$j = 1\dots n$}
		\IF{$x_j \in \wordcloud^+(z_i) \wedge w_{yi} > 0$}
		\STATE $Z^+ = Z^+ \cup \{(x_j, z_i)\}$ \textit{// candidate word substitution with positive weight and positive activation}
		\ELSIF{$x_j \in \wordcloud^-(z_i) \wedge w_{yi} < 0$}
		\STATE $Z^- = Z^- \cup \{(x_j,z_i)\}$ \textit{// candidate word substitution with negative weight and negative activation}
		\ENDIF
		\ENDFOR
		\ENDFOR
		\IF{$|Z^+ \cup Z^-|=0$}
		\STATE \Return -1 \textit{// No overlapping words found for counterfactual generation}
		\ENDIF
		\STATE Randomly select $(x_j,z_i) \in Z^+ \cup Z^-$ \textit{// select a random word to substitute and its corresponding feature}
		\IF{$(x_j,z_i)\in Z^+$}
		\STATE Randomly select $\hat x_j \in \wordcloud^-(z_i)$ \textit{// if positive, select a random negative word}
		\ELSIF{$(x_j,z_i)\in Z^-$}
		\STATE Randomly select $\hat x_j \in \wordcloud^+(z_i)$ \textit{// if negative, select a random positive word}
		\ENDIF
		\STATE $\hat x = (x_1, \dots, x_{j-1}, \hat x_j, x_{j+1}, \dots, x_n)$ \textit{// perform word substitution}
		\STATE \Return $\hat x$ \textit{// return generated counterfactual}
	\end{algorithmic}
\end{algorithm}

\subsection{Language counterfactuals}
We describe in detail how to generate counterfactuals from the 
word cloud interpretations and the linear decision layer. The complete algorithm can be found in Algorithm \ref{alg:nlp_counterfactual}, which we describe next. 

Let $x = (x_1, \dots, x_n)$ be a sentence with $n$ words, $z = f(s) \in \mathbb R^m$ be the deep encoding of $x$, and $y = \argmax_y w_yz + b_y \in [k]$ be the model's prediction of $x$ for a given decision layer with coefficients $(w,b)$. 
Our goal is to generate a counterfactual that can flip the model's prediction $y$ to some other class. 
Furthermore, let $\wordcloud^+(z_i)$ and $\wordcloud^{-}(z_i)$ be the 
LIME-based word clouds representing the positive and negative activations 
of $i$th deep feature, $z_i$. Then, counterfactual generation in the language 
setting involves the following steps:  

\begin{enumerate}
	\item Find all deep features which use words in $x$ as evidence for the predicted label $y$ (according to the word clouds). Specifically, calculate $Z = Z^- \cup Z^+$ where
	\begin{align}
	Z^+ =& \{(x_j,z_i) : \exists j \text{ s.t. } x_j \in \wordcloud^+(z_i) \wedge  w_{yi} > 0 \}\\
	Z^- =& \{(x_j,z_i) : \exists j \text{ s.t. } x_j \in \wordcloud^-(z_i) \wedge  w_{yi} < 0 \}
	\end{align}
	\item Randomly select a deep feature (and its word) $(x_j,z_i) \in Z$
	\item If $z_i \in Z^+$, randomly select a word $\hat x \in \wordcloud^-(z_i)$. Otherwise, if $z_i \in Z^-$, randomly select a word $\hat x_j \in \wordcloud^+(z_i)$. 
	\item Perform the word substitution $x_j \rightarrow \hat x_j$ to get the counterfactual sentence, $\hat x = (x_0, \dots, x_{j-1}, \hat x_j, x_{j+1}, \dots, x_n)$. 
\end{enumerate}

Note that it is possible for there to be no features that use words in a given sentence as evidence for its prediction, which results in no candidate word substitutions (i.e. $\|Z\| = 0$). Consequently, it is possible for a sentence to have a counterfactual generated from the dense decision layer but not in the sparse decision layer (or vice versa). For our sentiment counterfactual experiments, we restrict our analysis to sentences which have counterfactuals in both the sparse and dense decision layers. However, we found that 
similar results hold if one considers all possible counterfactuals for each individual model instead. 

\begin{figure}[!b]
	\centering
	\includegraphics[width=0.7\columnwidth]{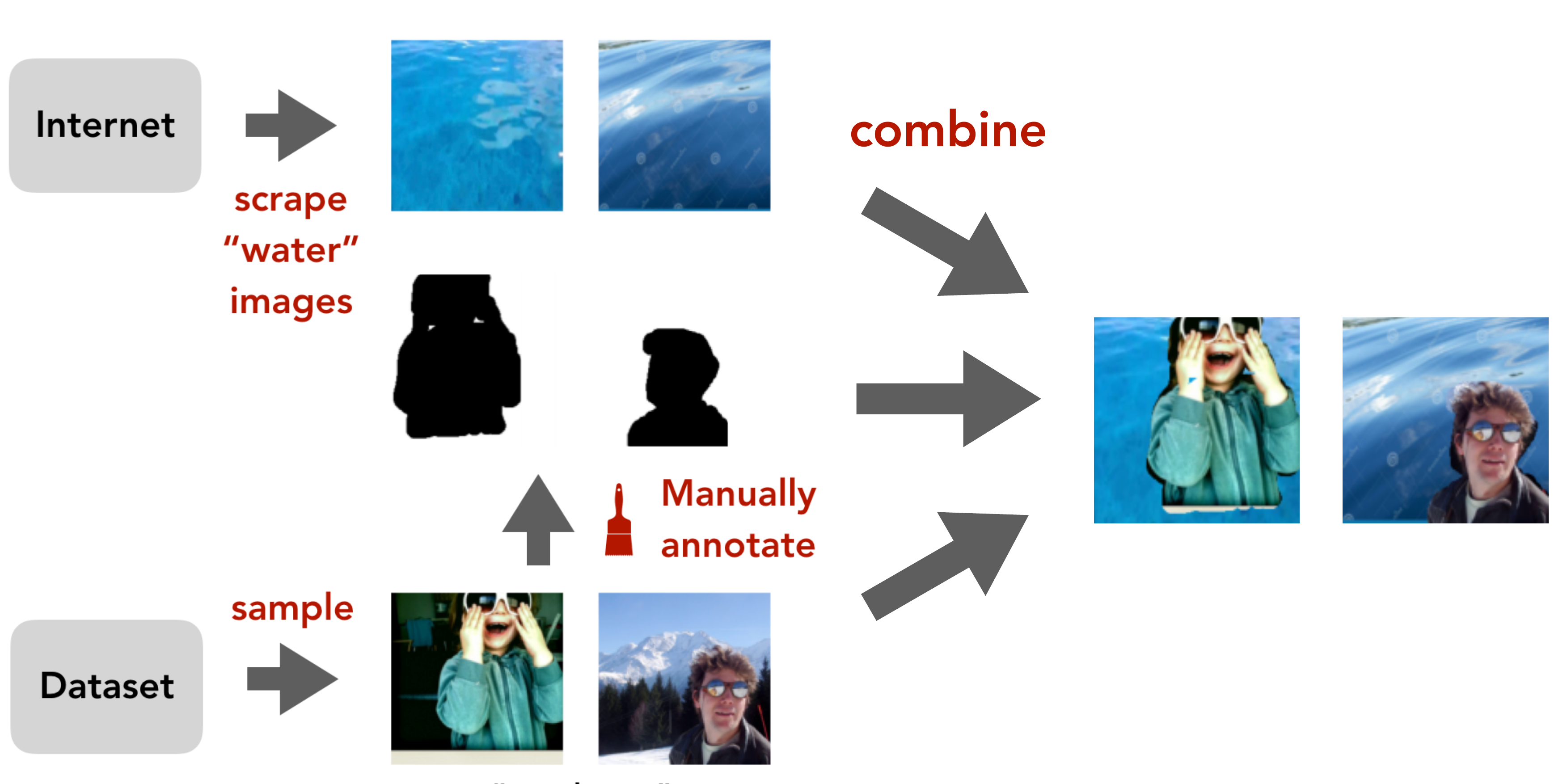}
	\caption{Image counterfactual generation process. We start with a 
		correlation identified during our MTurk study in 
		Section~\ref{sec:biases}---for example, the model associates ``water'' 
		with the class ``snorkel''. To generate the counterfactuals shown in 
		Figure~\ref{fig:counterfactuals_img}, we first select images from other 
		ImageNet classes. We then manually annotate regions in these images 
		to 
		replaces with ``water'' bacgrounds obtained via automated image 
		search 
		on the Internet. Finally, we additively combine the ``water'' backgrounds 
		and the original images, weighted by the mask, to obtain the resulting 
		counterfactual inputs.
	}
	\label{fig:app_vision_counterfactuals}
\end{figure}

\subsection{ImageNet counterfactuals}
In Figure~\ref{fig:app_vision_counterfactuals}, we illustrate our pipeline for 
counterfactual image generation. Our starting point is a particular spurious 
correlation (between a data pattern and a target class) identified via the 
MTurk study in Section~\ref{sec:biases}. We then select images from other 
ImageNet classes to add the spurious pattern to, and annotate the relevant 
region where it should be added. We obtain the spurious patterns by 
automatically scraping search engines. Finally, we combine 
the original images with the retrieved spurious pattern, using the mask as the 
weighting, to obtain the desired counterfactual images. These images are 
then supplied to the model, to test whether the addition of the spurious 
input pattern indeed fools the model into perceiving the counterfactuals as 
belonging to the target class.

%% file: appendix_errors.tex
\section{Validating ImageNet misclassifications}
\label{app:errors}

\subsection{Human study}
\label{app:mturk_mis}

We now detail the setup of our MTurk study from Section~\ref{sec:human}.  
For our analysis, we use a ResNet-50 that has been adversarially-trained 
($\eps=3$) on the ImageNet dataset. 
To obtain a sparse decision layer, we then train a sequence of GLMs via 
elastic net (cf. Section~\ref{sec:glm_explain}) on the deep representation of 
this network. Based on a validation set, we choose a single sparse 
decision layer---with 57.65\% test accuracy and 39.18 deep 
features/class on average. 

\paragraph{Task setup.}
In this task, our goal is to understand if annotators can identify data 
patterns that are responsible for misclassifications.
To this end, we start by identifying deep features that are strongly activated 
for  misclassified inputs.

For any misclassified input $x$ with ground truth label $l$ 
and predicted class $p$, we can compute for every deep feature $f_i(x)$:
\begin{equation}
\gamma_i = W[p,i] \cdot f_i(x) - W[l,i] \cdot f_i(x)
\end{equation}
where $W$ is the weight matrix of the decision layer. Intuitively, this score 
measures the extent to which a deep feature contributes to the predicted 
class, relative to its contribution to the ground truth class.
Then, sorting deep features based on decreasing/increasing values of this 
score, gives 
us a measure of how important each of them are for the predicted/ground 
truth label.
Let us denote $f_p$ as the deep feature with the highest score $\gamma_i$ 
and $f_l$ as the one with the lowest.

We find that for the robust ResNet-50 model with a \sparsemod{}, the single 
top deep feature based on this score ($f_p$) alone is responsible for ~26\% 
of the 
misclassifications (5673 examples in all).
That is, for each of these examples, simply turning $f_p=0$ flips the model's 
prediction from $p$ to $l$.
We henceforth refer to these deep features (one per misclassified input) as 
``problematic'' features.

\begin{figure}[!t]
	\centering
	
	\includegraphics[width=0.9\columnwidth]{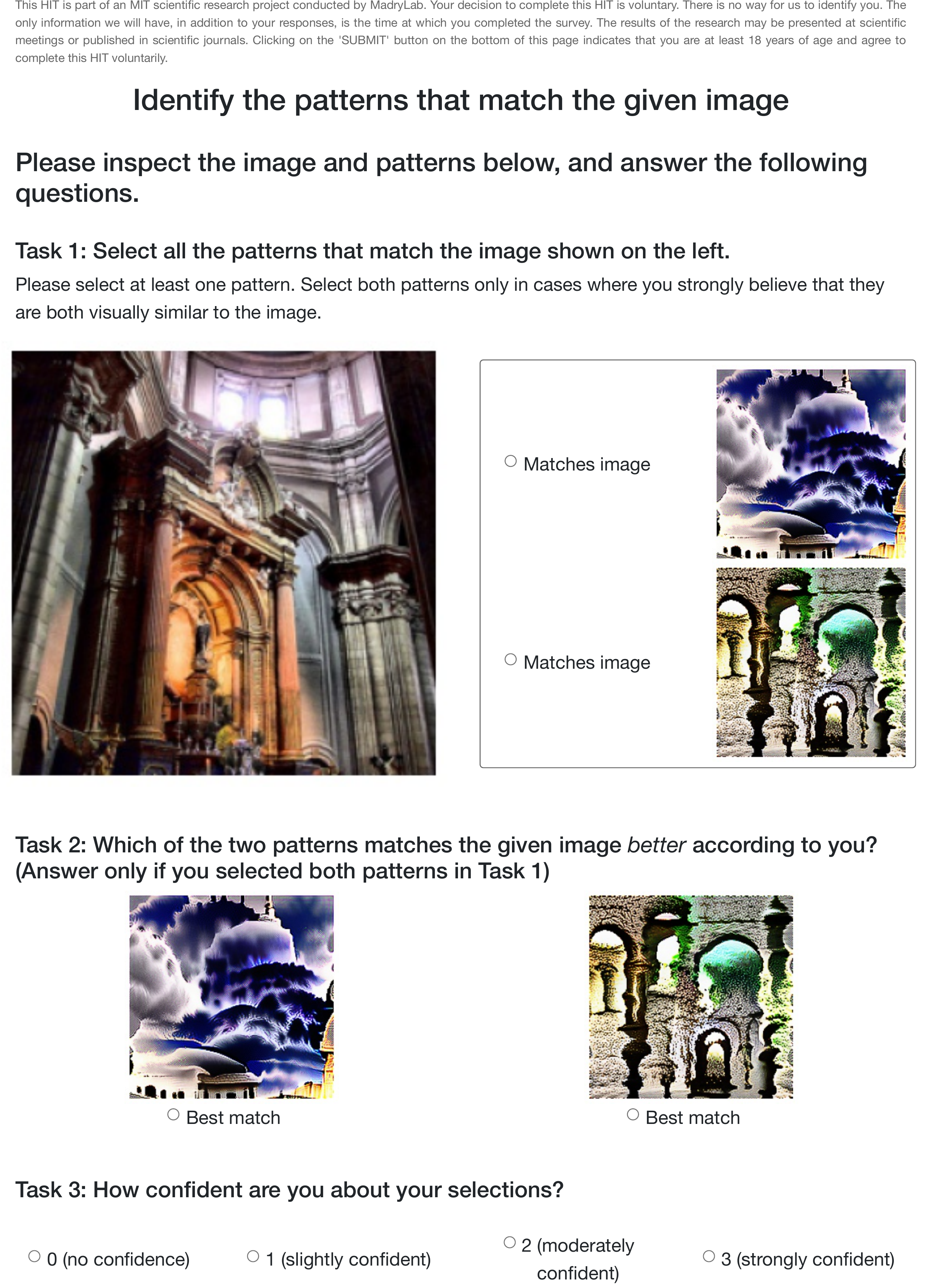}
	\caption{Sample MTurk task to identify input patterns responsible for the 
		misclassifications in deep networks with the help of their (sparse) 
		decision 
		layers.}
	\label{fig:app_task_mis}
\end{figure}
For our task, we randomly subsample 1330 of the aforementioned 5673 
misclassified inputs. We then construct MTurk tasks, wherein annotators are 
presented with one such input (without any information about the ground 
truth or predicted labels), along with the feature visualizations for two deep 
features. These two features are either (with equal probability):
\begin{itemize}
	\item ($f_l$, $f_p$): The deep features which (relatively) contribute most 
	to the ground truth and predicted class respectively.
	\item ($f_l$, $f_r$): The deep feature which (relatively) contributes most 
	to the ground truth class, along with a randomly-chosen one (out of the 
	2048 possible deep features). This is meant to serve as a control.
\end{itemize}

Annotators are then asked: (a) to select all the patterns (i.e., feature 
visualization of a deep feature) that match the image; (b) to select the one 
 that best matches the image (if they selected both in (a)); (c) to mark 
their confidence on a likert scale.
A sample task is shown in Figure~\ref{fig:app_task_mis}.
Each task was presented to 5 annotators, compensated at \$0.03 per task. 

Note that, in the case where the ground truth label for each image is actually 
pertinent to it and that model relies on semantically-meaningful deep features 
for every class, we would expect annotators to select $f_l$ to match the 
image 100\% of the time.
On the other hand, we would expect that annotators rarely select $f_r$ to 
match the image.

\paragraph{Quality control}
For each task, we aggregated results over all the annotators. While doing so, 
we eliminated individual instances where a particular annotator made no 
selections. We also completely eliminated instances corresponding to 
annotators who consistently (>80\% of the times) left the task blank. Finally, 
while reporting our 
results, we only keep tasks for which we have selections from at least two 
(of 
five) annotators. We determine the final selection based on a majority vote 
over annotators, weighted by their confidence.

\subsection{Additional error visualizations}
\label{app:app_validated_errors}

\begin{figure}[h]
	\centering
\includegraphics[width=0.56\columnwidth]{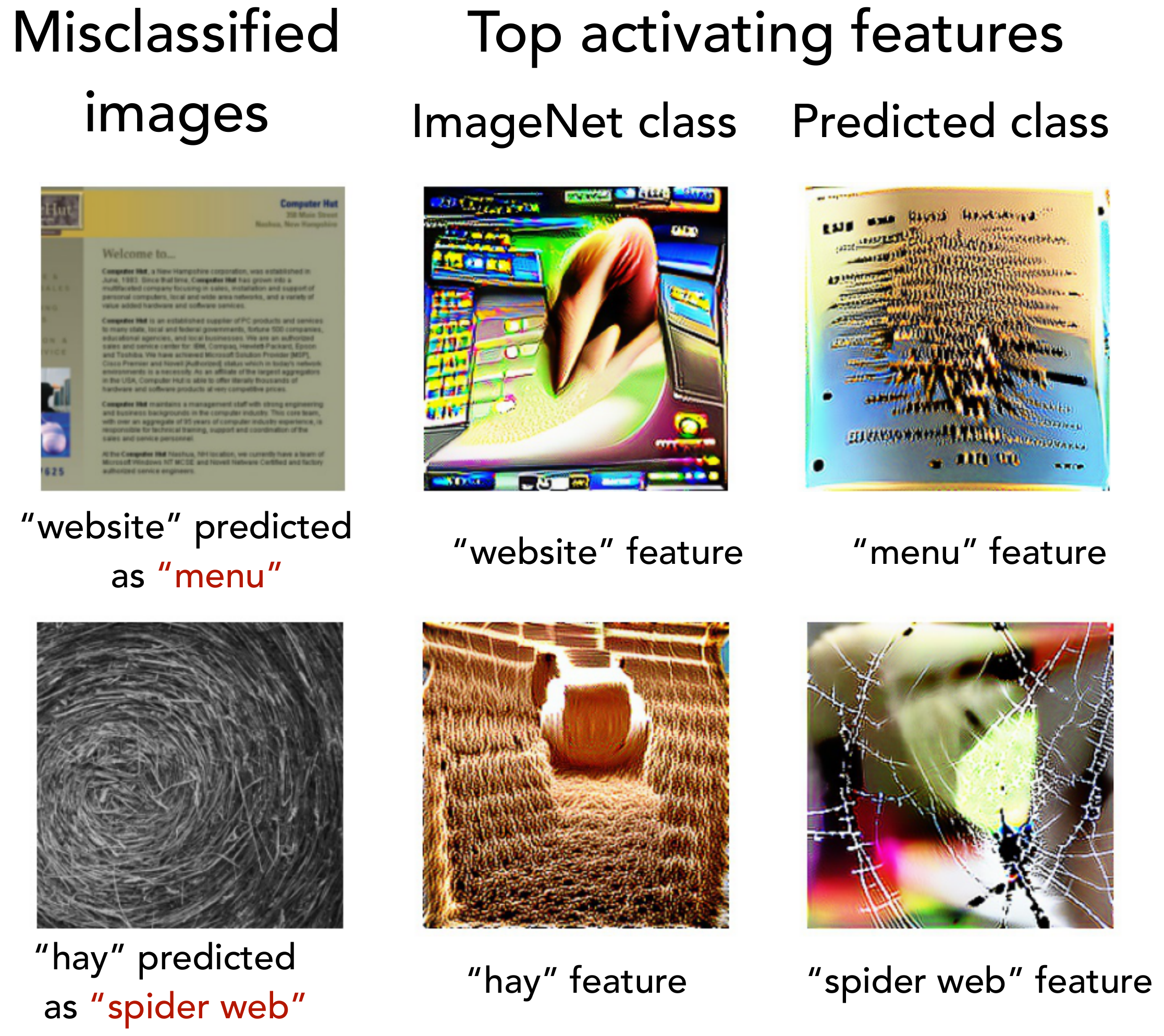}
\\ \vspace{0.25cm}
\includegraphics[width=0.54\columnwidth]{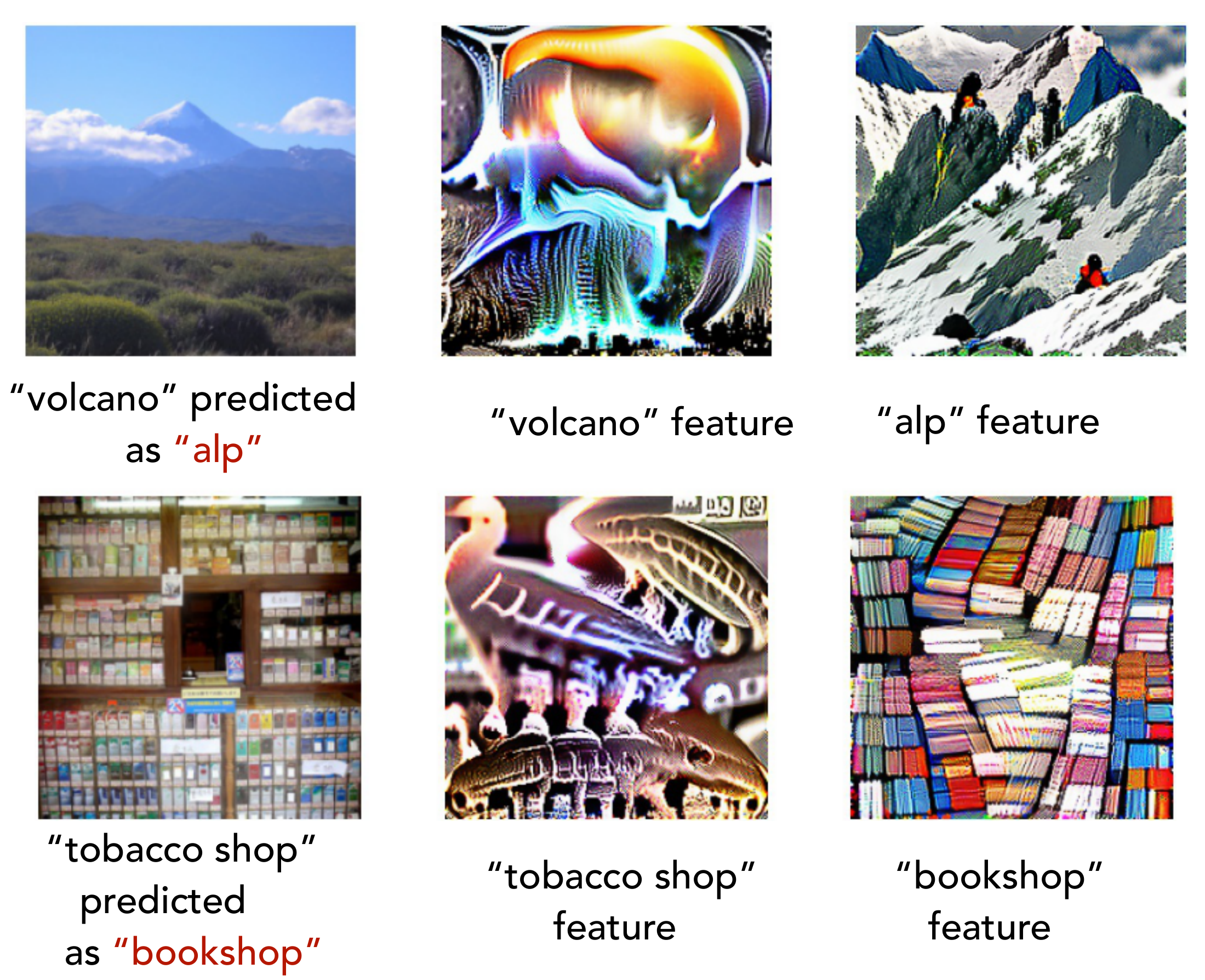}
	\caption{Additional examples of misclassified ImageNet images for which 
	annotators 
		deem the top activated feature for the predicted class 
		(\emph{rightmost}) 
		as a better match than the top activated feature 
		for the ground truth class (\emph{middle}).}
	\label{fig:app_misclassification_examples}
\end{figure}

In Figure~\ref{fig:app_misclassification_examples}, we present additional 
examples of misclassifications for which annotators the top deep feature 
used by the sparse decision layer to detect the predicted class to be a better 
match for the image than the corresponding top feature for the ground truth 
class.

\clearpage
\subsection{Model confusion}
\label{app:confusion}

\begin{figure}[!h]
\centering
\includegraphics[width=0.7\columnwidth]{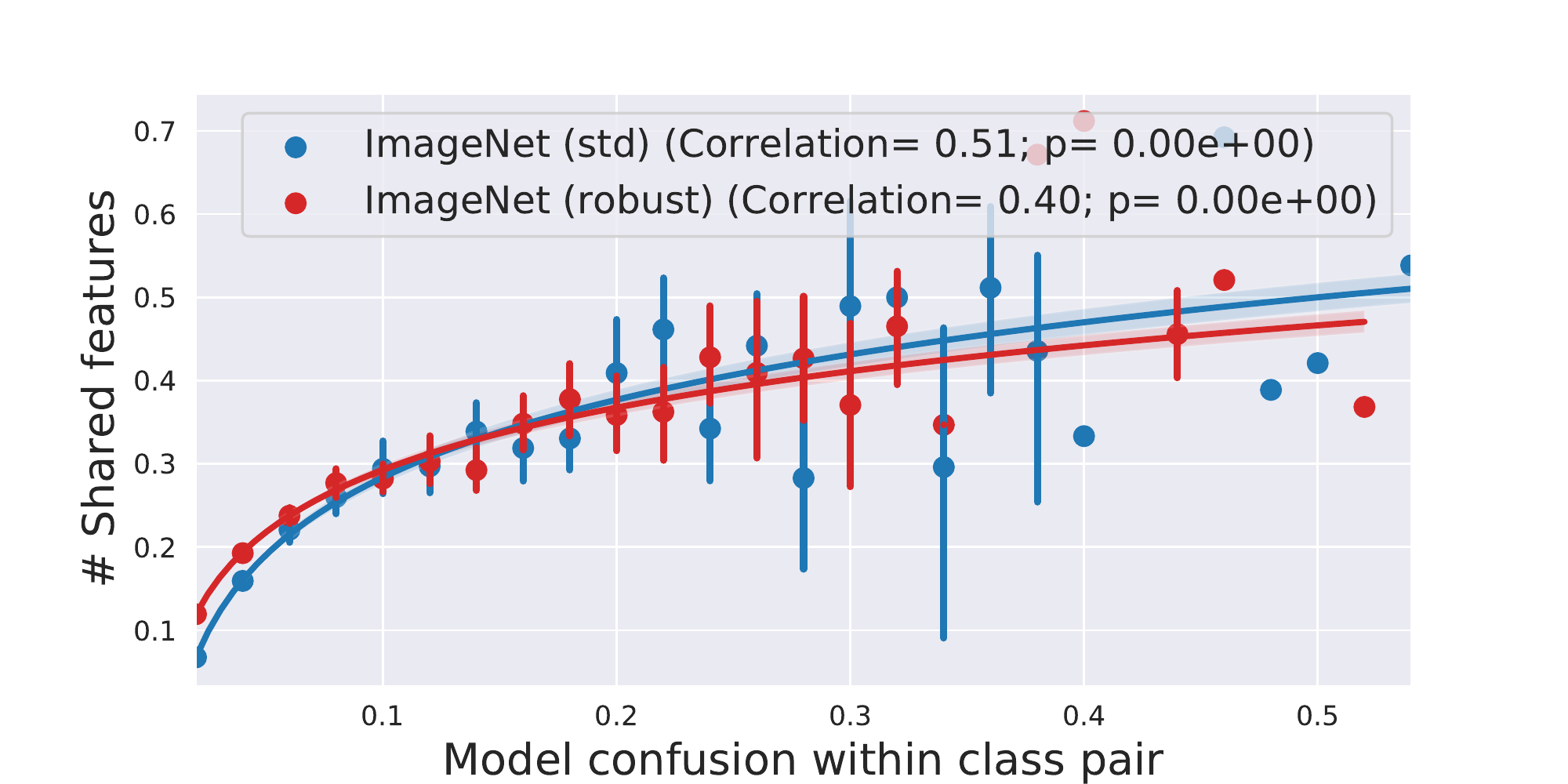}
\caption{Correlation between the number of features shared in the sparse 
decision layer of a model for two classes, and model confusion between 
them. Model confusion within a class pair is measured as the maximum of the 
corresponding entries ($C_{(i,j)}$, $C_{(j,i)}$) of the overall confusion matrix. 
}
\label{fig:app_confusion_correlation}
\end{figure}

\begin{figure}[!h]
	
	\centering
	\includegraphics[width=1\columnwidth]{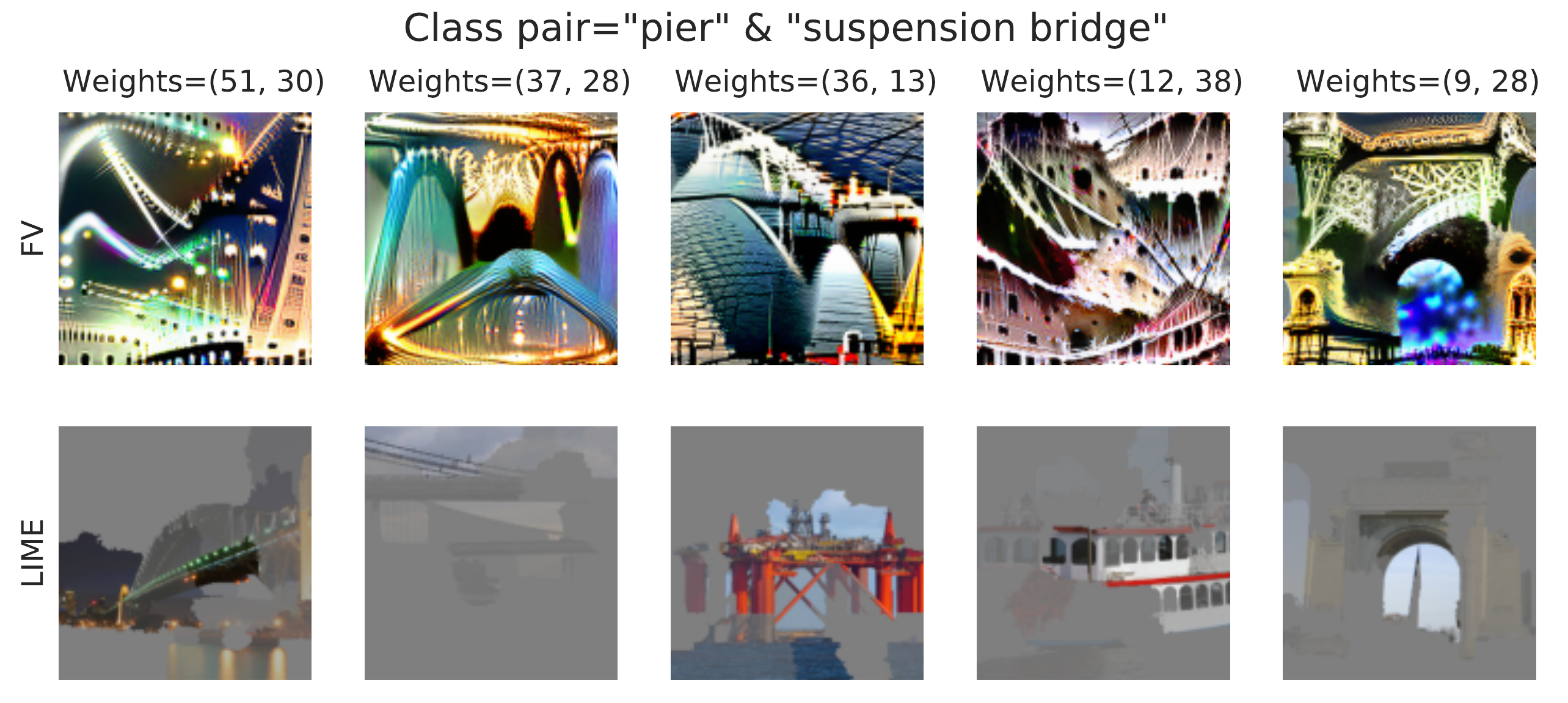}
	\caption{Sample visualization of confusing features: Five of the deep 
	features used by a robust ($\eps=3$) ImageNet-trained ResNet-50 with 
	\sparsemod{} to identify objects of classes ``pier'' and ``suspension 
	bridge'' which are frequently confused by the model ($C_{i,j}$ and 
	$C_{j,i}$ are 16\% and 24\% respectively).
	Each of these deep features is interpreted using 
		feature 
		visualizations (FV) and LIME superpixels; shown alongside their  linear 
		coefficients (W). }
	\label{fig:confusion_example}
\end{figure}

In Figure~\ref{fig:app_confusion_correlation}, we visualize the correlation 
between model confusion within a pair of classes, and the number of shared 
features between them in the sparse decision layer.
Model confusion within a class pair $(i,j)$ is measured as $\max(C_{(i,j)}, 
C_{(j,i)})$, where $C$ is the overall confusion matrix. We find that for 
models with sparse decision layers, the feature overlap between two classes, 
is significantly correlated with model errors within that class pair.
One can thus inspect the corresponding shared features---cf. 
Figure~\ref{fig:confusion_example} for an example---to better 
understand the underlying causes for inter-class model confusion.

